\pdfoutput=1
\documentclass{CVM}

\usepackage{times}
\usepackage{epsfig}
\usepackage{graphicx}
\usepackage{amsmath}
\usepackage{amssymb}
\usepackage{booktabs}
\usepackage{microtype}
\usepackage{caption}
\usepackage{enumitem}
\usepackage{float}

\usepackage{cuted}
\usepackage{capt-of}

\DeclareMathOperator{\E}{\mathbb{E}}
\DeclareMathOperator*{\argmax}{arg\,max}
\newcommand{\datasetname}{\emph{ArtSem}}

\CVMsetup{
type      = {Research Article},
doi       = {s41095-0xx-xxxx-x},
title     = {Controllable Multi-domain Semantic Artwork Synthesis},
author    = {Yuantian Huang$^{1}$\cor{}, Satoshi Iizuka$^{1}$, Edgar Simo-Serra$^{2}$, and Kazuhiro Fukui$^{1}$
},
runauthor = {Y. Huang, S. Iizuka, E. Simo-Serra and K. Fukui},
abstract  = {
    \begin{figure}[b]
    \small
        \begin{tabular}{p{80.5mm}} \toprule\\ \end{tabular}
        \vskip -3mm \noindent \setlength{\tabcolsep}{1pt}
        \begin{tabular}{p{3.5mm}p{80mm}}
    $1\quad $ & University of Tsukuba, Tsukuba, 305-8577, Japan. E-mail: Y. Huang, huang\_yuantian@cvlab.cs.tsukuba.ac.jp; S. Iizuka, iizuka@cs.tsukuba.ac.jp; K. Fukui, kfukui@cs.tsukuba.ac.jp.\\
    $2\quad $ & Waseda University, Tokyo, 169-8050, Japan. E-mail: ess@waseda.jp.
    \end{tabular} 
    \end{figure}
    
We present a novel framework for the multi-domain synthesis of artworks from
semantic layouts. One of the main limitations of this challenging task is 
the lack of publicly available segmentation datasets for art synthesis. To address
this problem, we propose a dataset called \datasetname~that contains
40,000 images of artwork from four different domains, with their corresponding
semantic label maps. We first extracted semantic maps
from landscape photography and used a conditional generative adversarial
network (GAN)-based approach for generating high-quality artwork from semantic
maps without requiring paired training data.
Furthermore, we propose
an artwork-synthesis model using domain-dependent variational encoders for
high-quality multi-domain synthesis. Subsequently, the model was improved and complemented
with a simple but effective normalization method based
on jointly normalizing semantics and style, which we call spatially
style-adaptive normalization (SSTAN). Compared to the previous methods, which only
take semantic layout as the input, our model jointly learns
style and semantic information representation, improving the generation quality
of artistic images. These results indicate that our model learned to
separate the domains in the latent space. Thus, we can perform
fine-grained control of the synthesized artwork by identifying hyperplanes
that separate the different domains. Moreover, by combining the proposed
dataset and approach, we generated user-controllable artworks of higher quality
than that of existing approaches, as corroborated by
quantitative metrics and a user study.

},
keywords  = {Semantic Artwork synthesis, Generative Adversarial Networks, Datasets, Non-photorealistic Rendering.},
copyright = {The Author(s) 2023\\
\\
\includegraphics[width=1\linewidth]{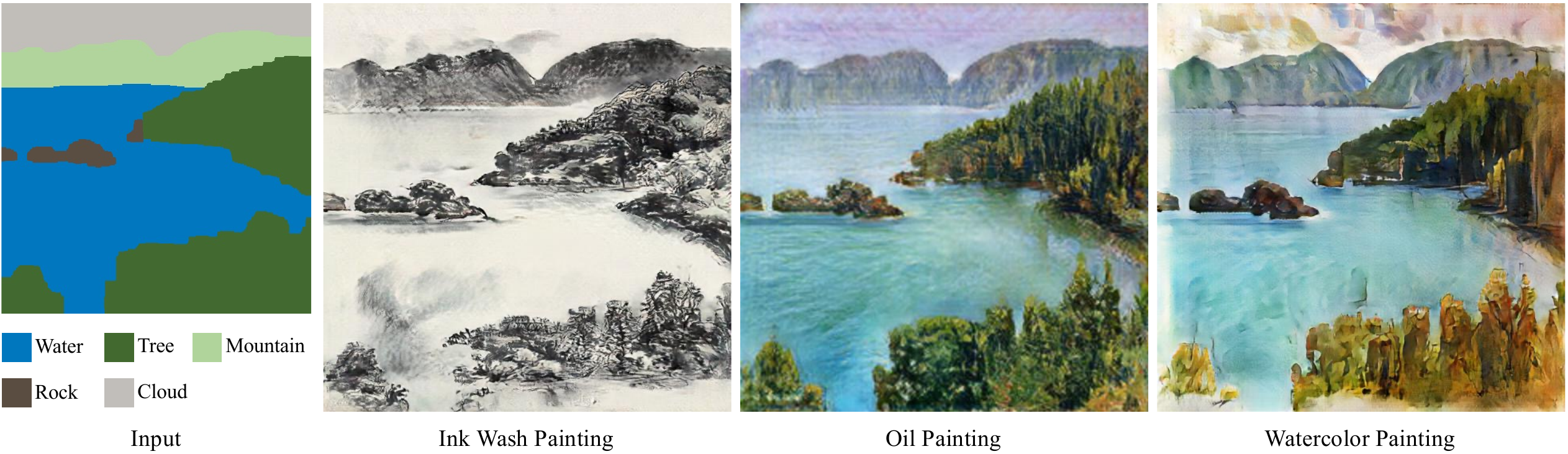}
    \captionof{figure}{\textbf{Controllable artwork synthesis.} Our approach can synthesize
    artwork from different domains using user-provided semantic label maps. The
    user can freely change the semantic layout and target domain to
    intuitively and interactively create new artworks.}
    \label{Top}
    },
}

\begin{document}

\maketitle

\begin{table*}[t]
\caption{\textbf{Comparison of interactive artwork synthesis approaches.}}

\centering
\begin{tabular}{rccccc}
 \toprule
         & Multiple domains  & Generation in     & Cross-domain   & Real-time           & Uses\\
 Methods & in a single model & a specific domain & style morphing & interactive editing & reference \\
 \midrule
 Two-step NST &   &  &  &  & \textbf{\checkmark}\\
 Co-ModGAN           & \textbf{\checkmark} &  &  & \textbf{\checkmark} & \\
 OASIS               & \textbf{\checkmark} &  &  & \textbf{\checkmark} & \\
 SMIS          & \textbf{\checkmark} &  &  & \textbf{\checkmark} & \checkmark\\
 SEAN          & \textbf{\checkmark} & \textbf{\checkmark} (fixed) &  & \textbf{\checkmark} & \checkmark\\
CMSAS~(Ours)          & \textbf{\checkmark} & \textbf{\checkmark} (diverse) & \textbf{\checkmark} & \textbf{\checkmark} & \textbf{\checkmark} \\
 \bottomrule
\end{tabular}
\label{Approaches_Comparison}
\end{table*}

\section{Introduction}
Image synthesis consists of generating new images using existing
data, whereas artwork synthesis focuses on generating images in the art domain. While
there is no consensus on whether computer-synthesized images are
art~\cite{hertzmann2018_art}, it has applications such as being used as an art
teaching tool, inspiring other artists, and providing different perspectives
when understanding the artwork. Furthermore, conditional artwork synthesis 
has emerged as an important tool for artists to generate new art with a notable
examples such as image stylization~\cite{styletransfer} which can
transfer artistic styles to photography while preserving content.

Existing approaches to artwork generation have focused on unconditional
generation~\cite{artgan,elgammal2017can}, image-to-image
translation~\cite{zhu2017unpaired}, or generation from
sketches~\cite{pix2pix,liu2020sketch}. However, these approaches yielded 
limited or no output control, making it difficult to determine whether, for example, 
a sketch is of a mountain or mountain-shaped rock. We
proposed using semantic maps that provide high control over the
generated image content. Some existing approaches can only partially achieve 
the goals~\cite{men2018common,champandard2016semantic} and require
additional user inputs to generate the images.
For more general image generation, Park \etal~\cite{park2019gaugan} 
proposed spatially-adaptive denormalization (SPADE) for semantic image synthesis
allowing users to control the semantics and styles when synthesizing a
photorealistic image. However, it cannot exploit multi-domain data and
require a paired training dataset. 

This study proposes solving the controllability and data problems by
introducing a new semantic map and artwork paired dataset, named
\datasetname, and a multi-domain high-quality artwork synthesis
model. The dataset is created from landscape photographs by
first computing semantic maps using an off-the-shelf semantic segmentation
model with graph cut-based post-processing to create human-like maps.
Subsequently, we train landscape photography with an artwork generation model using
unpaired training data to create high-quality multi-domain paired training
data. Finally, the refinement process ensured high-quality data
suitable for training artwork synthesis models.

We propose a controllable multi-domain semantic artwork synthesis ~(CMSAS) model
consisting of multiple domain-specific variational encoders that 
convert the artwork images into latent vectors and a generator that converts
semantic label maps with an encoded latent vector in artwork images. 
Motivated by the observation that the images generated by SPADE
are inclined to be less artistic,
we based our generator on a novel spatially style-adaptive normalization (SSTAN)
modules. While the SPADE modules are only aware of semantic information, SSTAN
modules inject latent codes that represent styles into the normalization
layers, and perform feature map modulation using semantics and style
information. Accordingly, the learned modulation parameters depend on
the input semantic layout and latent code, effectively propagating 
information throughout the network and significantly improving the results.
Furthermore, because multiple encoders can learn
domain-specific latent vectors, we propose to separate the domain latent vectors
with a hyperplane for fine-grained control of the output artwork domain. 
Finally, the developed CMSAS model
generates artwork from a semantic map for interactive applications.

We evaluated our approach quantitatively using automatic metrics and 
a perceptual user study in addition to the qualitative results. Results show that
our approach outperformed the existing approaches in all metrics.

This study presents the following contributions:
\begin{itemize}[noitemsep,nolistsep,leftmargin=*]
\item A single-model semantic artwork synthesis approach that generates high-quality artworks from easy-to-manipulate semantic layout inputs in multiple domains.
\item A high-quality pixel-aligned semantic artwork dataset that contains artistic images with paired segmentation masks.
\item An effective normalization method that significantly improves the artwork generation quality.
\item Highly controllable generation with domain and style control via latent space manipulation
\item In-depth evaluation of our method based on qualitative and quantitative comparisons with existing approaches.
\end{itemize}


\section{Related Work}
This section discusses related works and compares approaches similar to our
proposed method. A high-level comparison of similar approaches is presented in
Table~\ref{Approaches_Comparison}.
Specifically, two-step neural style transfer~(NST) approaches
combining semantic natural image synthesis~\cite{park2019gaugan}
and style transfer~\cite{styletransfer} are computationally expensive
for real-time editing. Moreover, their practical use is limited
by the restricted set of pre-defined styles.
Image-conditioning approaches such as Co-ModGAN~\cite{COMOD} cannot
generate artworks in a specific domain or use a reference style image.
The same limitations are shared by OASIS~\cite{OASIS} since it relies on
random noise as the style input.
SMIS~\cite{zhu2020SMIS} uses a single variational auto-encoder
(VAE) to transform input images into latent vectors for reference-style images.
However, these latent vectors are inseparable and cannot be constrained within
a specific domain, as discussed in Section~\ref{DomainControl}.
SEAN~\cite{zhu2020sean} can generate artwork in a specific domain
using a fixed pre-computed mean style code, but the approach lacks diversity
and cannot interpolate between different domains.
Our results verify that the proposed approach is the most flexible of the existing approaches.

\subsection{Artwork Synthesis}
The concept of artwork synthesis goes back to image
analogies~\cite{image_analogies}, where filters
related to a painting style are automatically learned from training data
based on a simple multi-scale
autoregression. It is a generative
art~\cite{dehlinger2007fine, phon2012controlling} that attempts to generate
artworks algorithmically based on heuristics.
Recently, neural-style transfer~\cite{styletransfer} methods that solve
the problem of applying artistic styles derived from reference images to
photographs while preserving their content through optimization have
become popular. The neural-style transfer is further discussed in 
next section.
With the advent of generative adversarial networks (GAN)~\cite{goodfellow2014},
data-driven generation approaches have become dominant.
For GAN-based artwork generation, artGAN~\cite{artgan, artgan_improved} and
CAN~\cite{elgammal2017can} extend and apply GANs to
generate artworks. However, the low-resolution and limited quality of
generated artworks restrict their applicability. Although some StyleGAN-based
methods~\cite{xue2021end,dobler2022art} and text-conditioned diffusion models
~\cite{ramesh2022hierarchical, rombach2022high} generated higher-quality artworks, the
results are less controllable, and diffusion models incur significantly higher computational costs
than GAN-based approaches, significantly limiting their application.
Pix2pixGAN~\cite{pix2pix} has demonstrated its ability to generate 
high-quality artwork from hand-drawn sketches. However, these approaches are
difficult to control because only the edge information is used.
Specifically, a sketch cannot clearly indicate the differences between 
different objects having the same contour (e.g., mountains and rocks). Unlike
previous studies, our novel semantics-based model allows users to control the
shape, contour, and the semantic information of the input label map
to create new artwork with higher quality and fidelity.

\subsection{Neural Style Transfer}
Gatys \etal~\cite{styletransfer} first used a convolutional neural network
(CNN)~\cite{CNN} to extract style and content representations from the images
and optimize the image content to match different art styles. Their
approach creates new images by optimizing style and content loss,
calculated by matching the Gram matrix statistics of pre-trained CNN
features. Neural style transfer has many interesting applications in
fields such as art, graphics, images, and video processing. There are various
extensions and improvements to the original neural-style transfer, focusing on
different aspects, such as multiple style transfer~\cite{styimprove1},
content-aware style transfer~\cite{styimprove2, styimprove3}, and other
aspects~\cite{styimprove5, STROTSS, EMD, AdaIN}.
This study adopted the concept of style loss to improve the
generation quality of non-photorealistic style images in our dataset. 
The effectiveness of the method is verified by 
comparing it with state-of-the-art methods.

\subsection{Image-to-Image Translation}
Image-to-image translation aims to convert an image input into the desired image
output and is dominated by CNN approaches.
Supervised approaches~\cite{pix2pix, highresolutionpix2pix} employ paired training
data and obtain impressive results. However, for the art domain, obtaining
paired training data is challenging.
Many approaches have started to focus on
the unsupervised image-to-image translation~\cite{zhu2017unpaired,
yi2017dualgan, hoffman2018cycada, zhu2017toward,
huang2018multimodal, lee2018diverse, lee2020drit++, liu2019few, chen2018gated,
chang2020domain, SwappingAutoencoder}. A notable example is
CycleGAN~\cite{zhu2017unpaired}, which uses cycle consistency with
adversarial losses to overcome the lack of paired images. Although
limited in resolution, they demonstrated the possibility of producing realistic images
for various datasets, including several Western painting styles. 
Based on CycleGAN, ChipGAN~\cite{ChipGAN} introduced edge loss
that enforces brushstroke constraints and successfully generates impressive
results for ink-wash painting. However, these methods work only with
photorealistic images as inputs and cannot handle abstract inputs such
as semantic label maps.
Therefore, we designed a
two-stage image-transformation framework to produce a high-quality
paired dataset of semantic labels and artwork images.
Replacing photographs with semantic label maps as inputs
gives users more control over the final result while preserving
high performance using the existing techniques.

\begin{figure*}[ht]
\centering
\includegraphics[width=1\linewidth]{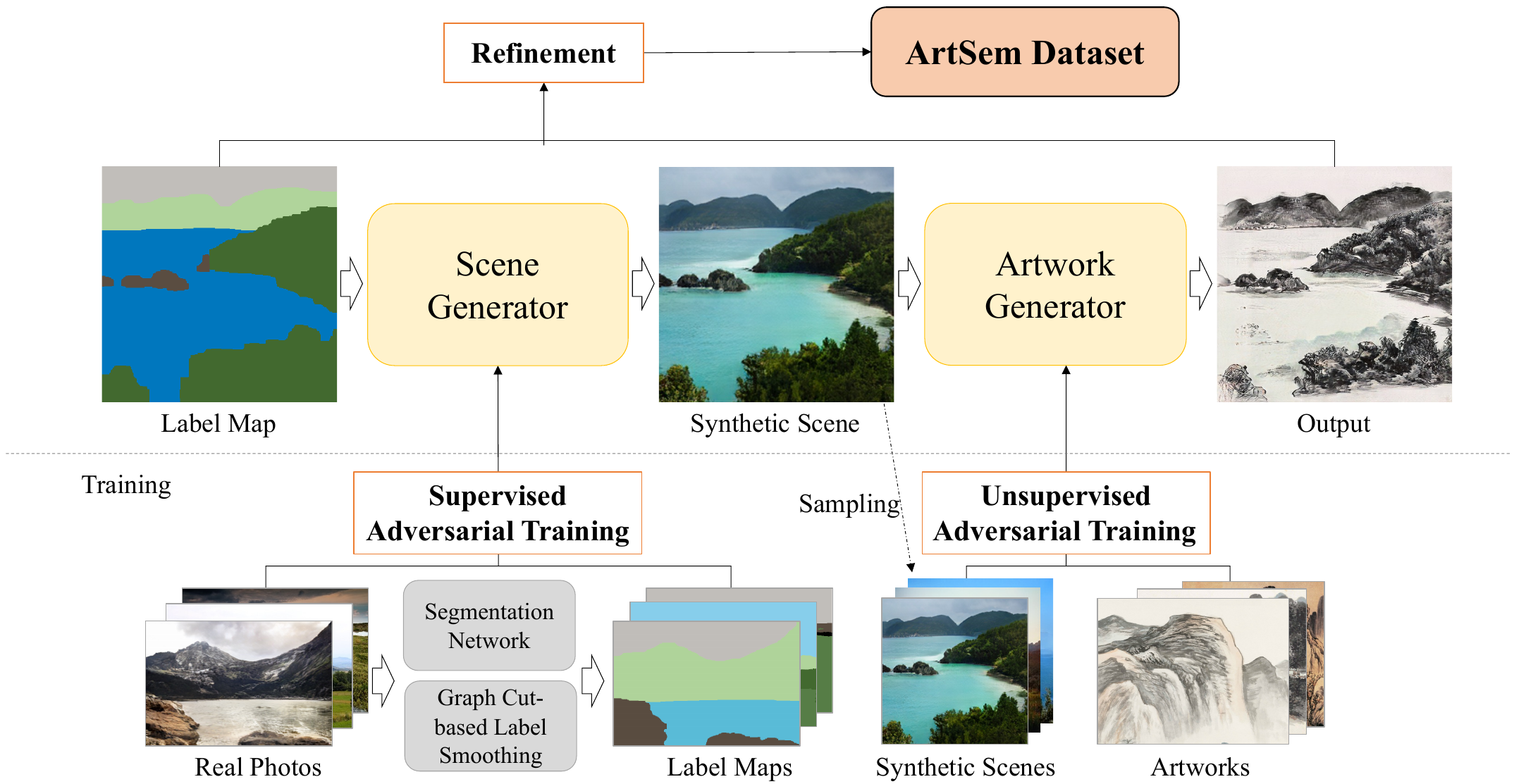}
    \caption{\textbf{Our proposed dataset construction approach.} We use a
    segmentation network~\cite{deeplabv2} and graph cut-based smoothing to obtain
    semantic label maps for training the landscape scene generator. The synthesized
    scene images and real artworks are then used as a supervised training set for
    synthesizing art images. Finally, we use a refinement approach that
    mimics human judgment to obtain the final training dataset. After training
    with the dataset, our framework can generate high-quality artworks conditioned
    on input semantic label maps.}
    \label{fig:overview}
\end{figure*}

\subsection{Semantic Image Synthesis}
As a subfield of image-to-image translation, semantic image synthesis focuses
on generating new images conditioned on semantic image maps, and are mainly
dominated by GAN~\cite{goodfellow2014}. 
For example, pix2pixGAN~\cite{pix2pix} firstly established a common framework
for mapping paired data, successfully generating realistic images
and has been extended in further research~\cite{v2v, choi2018stargan, highresolutionpix2pix,
qi2018semiparametric, wang2019example, zhang2019shadowgan, zhou2021jittor}.
Recently, Park \etal~\cite{park2019gaugan} proposed SPADE,
which outperformed the former approaches in terms of generating
photorealistic images conditioned by semantic layout.
Based on the SPADE architecture,
SMIS~\cite{zhu2020SMIS} successfully produced semantically multimodal images by 
replacing all regular convolution layers in the generator with group convolutions.
OASIS~\cite{OASIS} surpassed SPADE in terms of diversity while
maintaining similar quality by redesigning the discriminator.
SEAN~\cite{zhu2020sean} used style input images to create spatially 
varying normalization parameters per semantic region.
Other improvements~\cite{wang2023towards} have also been made in different aspects.
Another notable approach is the Co-Mod GAN~\cite{COMOD} 
which achieves semantic image synthesis via the co-modulation of 
conditional and stochastic style representations 
based on unconditional generative architectures, such as StyleGAN2~\cite{styleGANv2}.

Although these methods generate good results in photorealistic image
generation, the results were less convincing when generating
artwork. Furthermore, our approach focused on providing in-depth controllability, while
most existing approaches have limited or zero controllability.


\section{ArtSem Dataset}
Ideally, we would prefer using high-quality manual semantic segmentation
annotations of diverse artworks for supervised semantic training of semantic
artwork synthesis. However, given the nature of the art, it is challenging to
create high-quality annotations because of the need for domain-specific knowledge
and the inherent ambiguity of the artwork. 
Recent research~\cite{SegmentationArt} has attempted to address this problem; 
however, they have limitations in terms of class and precision.
Instead of manual annotation, we proposed a method for the large-scale
automatic generation of paired semantic maps and diverse artwork.
Our approach is as follows: First, we use a semantic segmentation model to obtain
segmentation maps from real-scene photographs. Next, we train an
image-to-image translation model to convert semantic maps into different
art styles. We base our approach on conditional unsupervised adversarial
training, which can learn the mapping from semantic maps to artworks.
Finally, we perform refinement based on the mimicked human judgment to generate a
final dataset. An overview of the dataset construction approach is presented in
Fig.~\ref{fig:overview}. This study focuses on various landscape
paintings, such as canyons and lake shores.

\subsection{Data Collection}
First, we collected 50,000 landscape photographs of various outdoor scenes
from Flickr. We then removed 15,000 samples based on generated label
maps if they had irrelevant labels, such as people or animals. 
Of the remaining 35,000 images, 1,000 were used as the validation set.
We obtained images from four different domains: ink-wash, Monet oil, Van Gogh oil, and watercolor
paintings. The images were obtained as follows:
\begin{itemize}[noitemsep,nolistsep,leftmargin=*]
\item \emph{Ink-wash paintings:} We collect them from search engines and
manually remove images that are unrecognizable or contain too much irrelevant
content. We also homogenize the style and tone by manually removing images that
are different. This is repeated until 1,000 paintings are obtained.
\item \emph{Monet oil paintings:} These are taken from Zhu
\etal~\cite{zhu2017unpaired} and were originally downloaded from
\url{wikiart.org}. There are a total of 1,072 Monet oil paintings.
\item \emph{Van Gogh oil paintings:} These were taken from the same dataset as
the Monet oil paintings. There is a total of 400 Van Gogh oil paintings.
\item \emph{Watercolor paintings:} Similar to the ink-wash paintings, we
collect 1,000 manually curated images using search engines.
\end{itemize}

\newcommand{\ssfig}[1]{\includegraphics[width=0.32\linewidth]{figures/comp_smoothing/#1}}
\begin{figure}[t]
    \centering
    \setlength{\tabcolsep}{1pt}
    \begin{tabular}{ccc}
    \ssfig{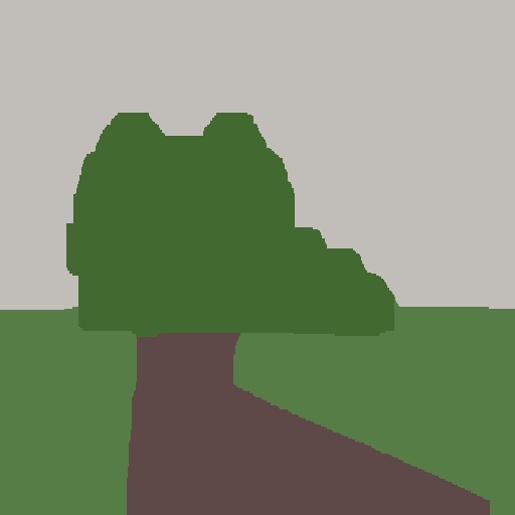} & \ssfig{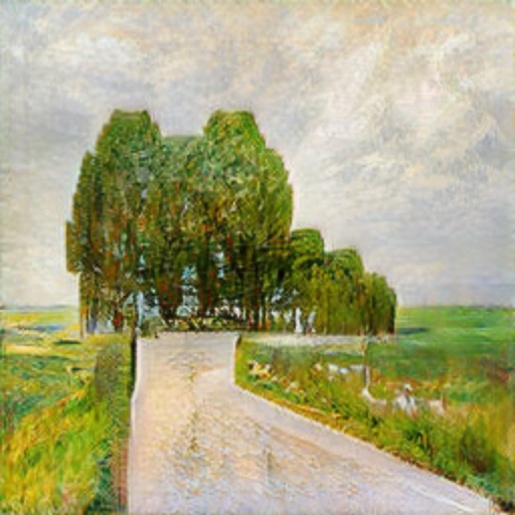} & \ssfig{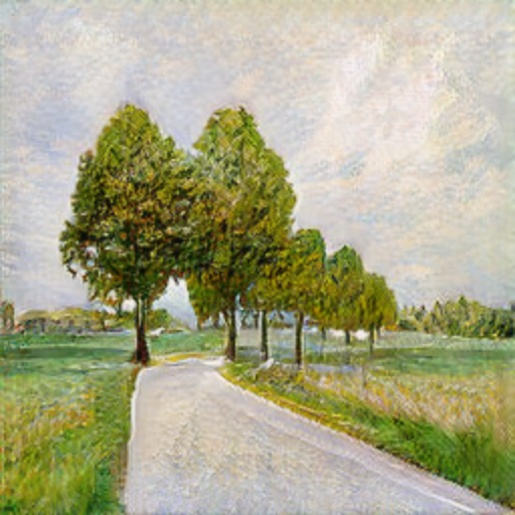} \\
    \ssfig{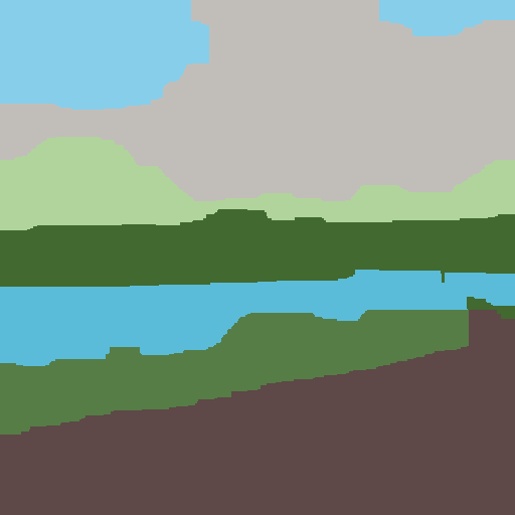} & \ssfig{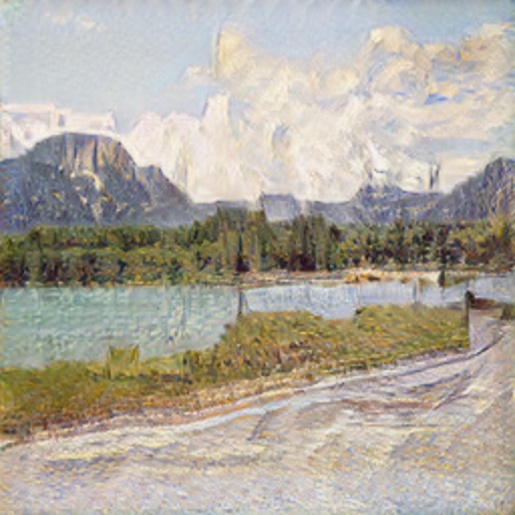} & \ssfig{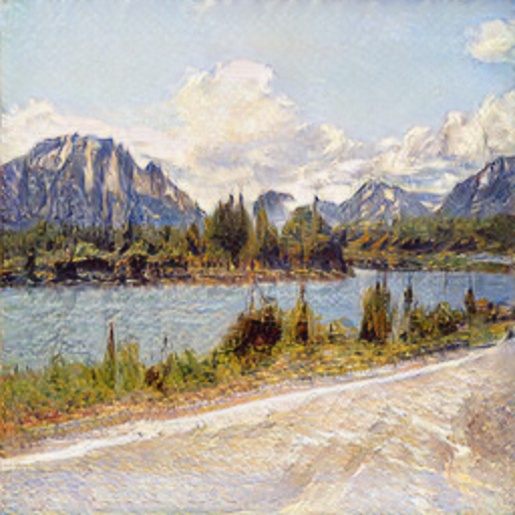} \\
    Input & w/o smoothing & with smoothing \\
    \end{tabular}
    \setlength{\tabcolsep}{6pt}
    \caption{\textbf{Effect of label smoothing.}
    We compare the results from our model trained on the
    datasets with and without graph-cut-based label smoothing, which
    eliminates small irregularities allowing for more natural synthesized
    images. The effects are especially noticeable around the edges and outline, where
    the smoothing is the strongest.}
    \label{fig:compare_smoothing}
\end{figure}

\subsection{Semantic Scene Generation}

In this section, we aim to learn to map from an input segmentation mask to a
photorealistic image. We employ a state-of-the-art network
architecture~\cite{park2019gaugan} for semantic image synthesis based on
spatially-adaptive normalization. This model requires a dataset containing 
several landscape photos and uses paired semantic layouts for training.
Although some datasets such as MSCOCO~\cite{coco} and
ADE20K~\cite{zhou2019semantic} are widely used in semantic
segmentation, most images in the dataset are not related to the landscape, which is
the focus of this study. We generate training data from 35,000 landscape
images using a pre-trained semantic segmentation network~\cite{deeplabv2} to
obtain paired training data. The model was trained on the MSCOCO
dataset~\cite{coco}, which outputs 182 different label classes. We removed
irrelevant labels and combined similar labels, such as moss and grass, to
simplify the semantic map generation. Finally, we use 16 classes that
commonly appear in the wild landscape. Please refer to the supplementary
materials for more details.

The generated semantic maps exhibited irregularities and small inconsistencies that
rendered them unsuitable for training high-quality generation models. We adopt a
graph-cut-based label smoothing approach~\cite{LabelSmoothingviaGraphCuts} to
improve the quality of the results, given that we only require a low-frequency
semantic map. An example of the smoothing effect is presented in
Fig.~\ref{fig:compare_smoothing}. 

Subsequently, we trained a SPADE model~\cite{park2019gaugan} to learn the mapping
from semantic maps to photorealistic landscape images using the generated data.
The model was trained with $256\times 256$ pixel images, and bicubic
up-sampling was used to increase the resolution to $512\times 512$ pixels.

\newcommand{\cefig}[1]{\includegraphics[width=0.32\linewidth]{figures/comp_edgeloss/#1}}
\begin{figure}[t]
   \centering
   \setlength{\tabcolsep}{1pt}
   \begin{tabular}{ccc}
   \cefig{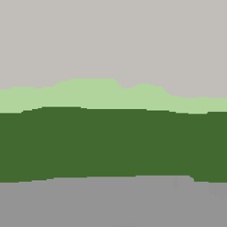} & \cefig{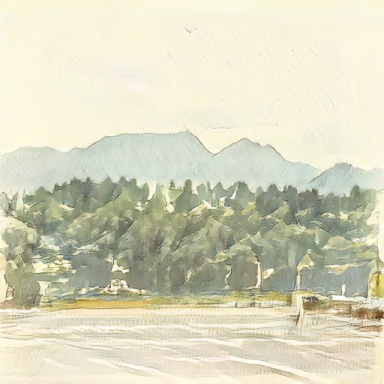} & \cefig{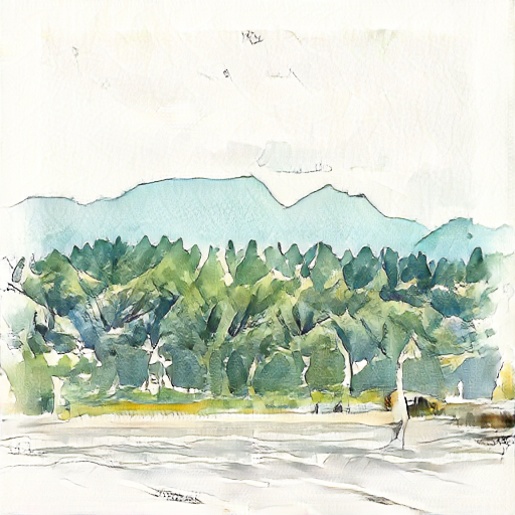} \\
   \cefig{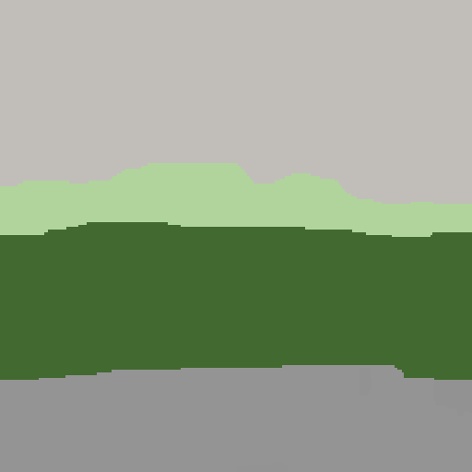} & \cefig{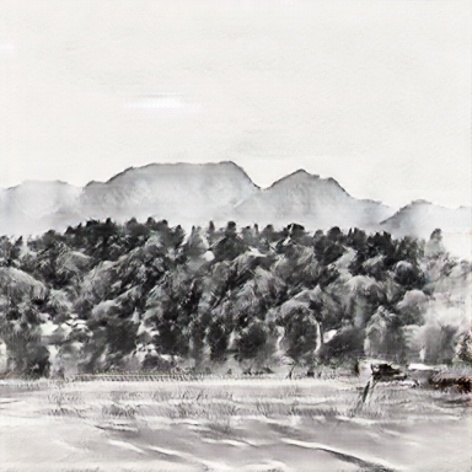} & \cefig{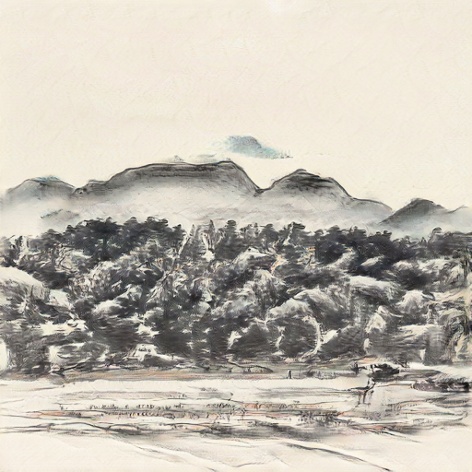} \\
   Input & w/o style and edge & with style and edge \\
   \end{tabular}
   \setlength{\tabcolsep}{6pt}
   \caption{\textbf{Effect of style and edge losses.}
   The model can generate
   clear outlines and realistic artwork that matches the target style by employing style and edge losses.}
   \label{fig:losses}
\end{figure}

\subsection{Weakly Supervised Artwork Generation}
Our objective at this stage is to learn a mapping function between the
synthesized landscape images and real artwork without any explicit paired
training data. We randomly selected 5,000 synthesized images from the previous
stage as landscape images and used the collected artwork images to train a
generation model. The as-synthesized images, when used for
inference, provided improved results than when using the synthesized images for training.
Please refer to the supplementary materials for a detailed discussion on this topic.

Our generation model is inspired by CycleGAN~\cite{zhu2017unpaired} and trains
jointly a model that generates artworks from synthetic landscapes and a model
that generates synthetic landscapes from the artwork. Using only 
adversarial and cycle-consistency losses led to poor results, given that the
synthesized images have damaged parts and the generated images have high
abstraction levels. To improve the results, we modified the training approach to
include the style and edge loss terms.
Separate models are trained for each modality.
Fig.~\ref{fig:losses} demonstrate that the proposed model successfully outperformed the base
model with two auxiliary losses.

\newcommand{\figsbad}[1]{\includegraphics[width=0.24\linewidth]{results/samples_dataset/bad_samples/#1}}
\begin{figure}[t]
\begin{center}
\setlength{\tabcolsep}{1pt}
\begin{tabular}{cccc}

\figsbad{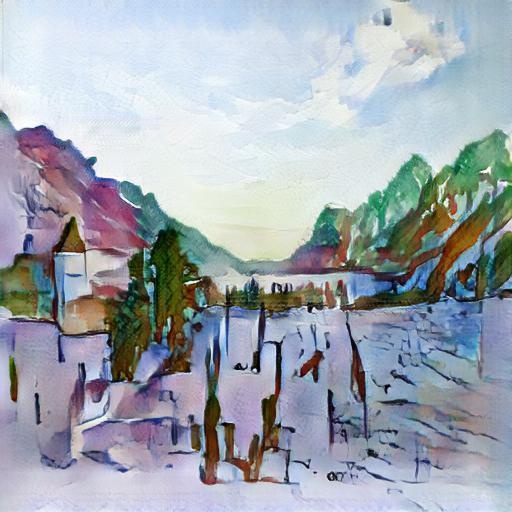} &
\figsbad{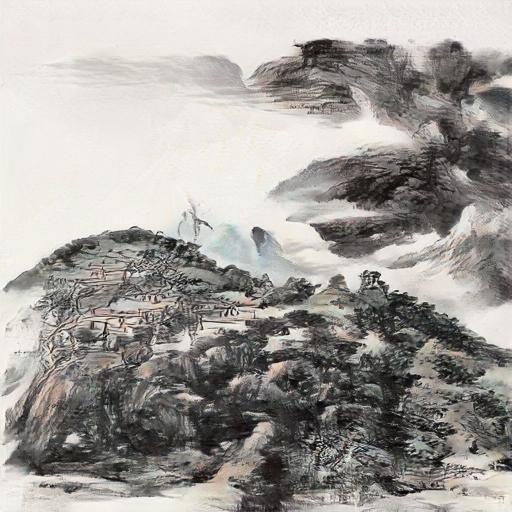} &
\figsbad{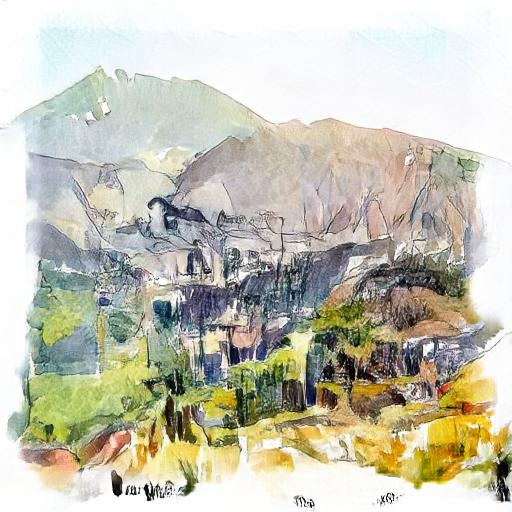} &
\figsbad{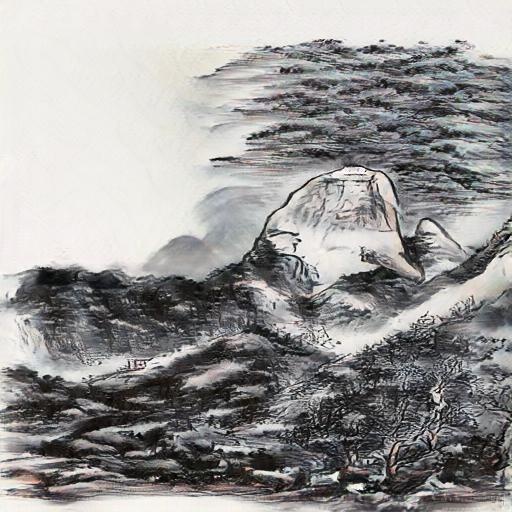} \\

\figsbad{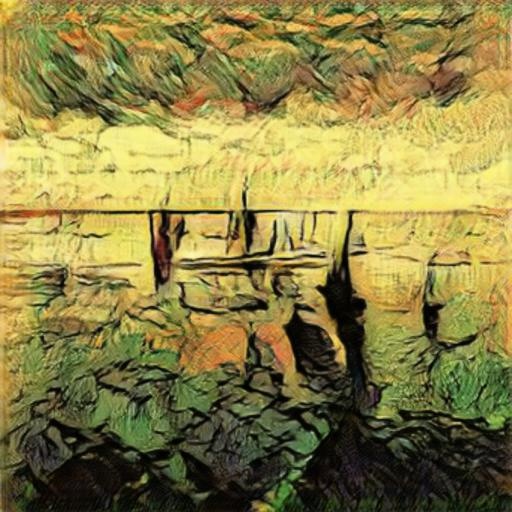} &
\figsbad{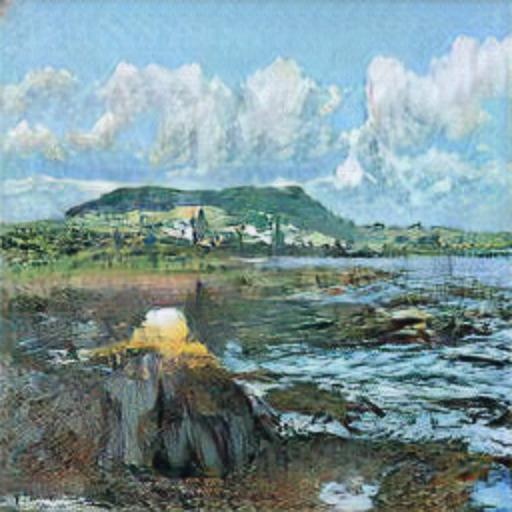} &
\figsbad{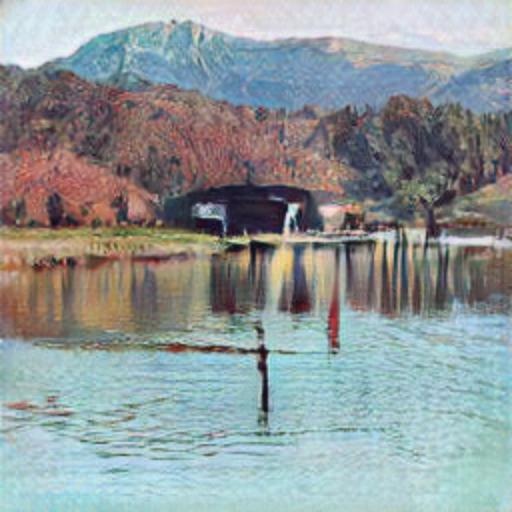} &
\figsbad{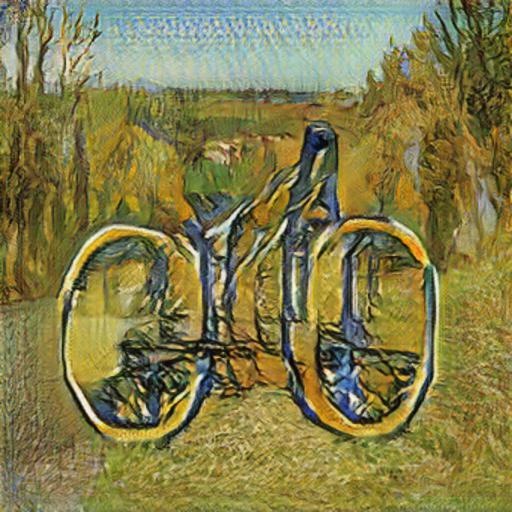} \\

\end{tabular}
\setlength{\tabcolsep}{6pt}
\end{center}
\caption{\textbf{Examples of automatically removed poorly generated images.} It
is difficult to identify what is in the scene if confusing
or partially drawn objects are present. Our dataset
refinement approach removes such features when generating the training data.}
\label{bad_samples}

\end{figure}

\subsubsection{Objective Function}
We employ adversarial and cycle-consistency losses in our model, which
simultaneously learns two mappings $G_{x\to y} \colon X \to Y$ and $G_{y\to x} \colon Y \to X$,
where $X$ and $Y$ are the source and target domains, respectively. For a mapping function
$G$ and its discriminator $D_Y$, the adversarial loss is given as
\begin{align}
    L_{adv} &= \E_{y\sim{X}}\left[\log{D_Y(y)}\right] + \E_{x\sim{Y}}\left[\log{(1-D_Y(G(x)))}\right]
    \label{ad_loss}
\end{align}
Cycle-consistency loss is defined as follows:
\begin{align}
    L_{cycle} &= \E_{x\sim{X}}\left[\| G_{y\to x}\left(G_{x\to y}(x)\right)-x \|_1\right] \nonumber \\
         &+ \E_{y\sim{Y}}\left[\| G_{x\to y}\left(G_{y\to x}(y)\right)-y \|_1\right]
    \label{cycle_loss}
\end{align}

We introduced style loss~\cite{styletransfer} to improve
the style similarity. In particular, we extracted style representations from 
subsets of VGG-19~\cite{VGG} layers: \textit{'conv1\_1', 'conv2\_1', 'conv3\_1,
' conv4\_1' and 'conv5\_1'}. Given the computed features $F(I)^l$ from 
layer $l$ for a synthesized or real image $I$, we compute the Gram matrix using
$M(I)^l_{ij} = \sum_k F(I)_{ik}^l F(I)_{jk}^l$. This allows style loss to
be computed as the weighted sum of the difference of the computed Gram matrices
as:
\begin{align}
& L_{style} = \nonumber \\
& \E_{x\sim{X},y\sim{Y}} \sum_{l=0}^L \frac{\omega_l}{4N^2_l M^2_l} \sum_{i,j}\left( M(G_{x\to y}(x))^l_{ij} - M(y)^l_{ij} \right)^2
    \label{style_loss_onelayer}
\end{align}
\noindent where $\omega_l$ is a weighting term for layer $l$, $N_l$ is the
number of feature maps of layer $l$, and $M_l$ is the number of pixels in each
feature map of layer $l$.

We further introduced edge loss~\cite{ChipGAN} to emphasize the contours and lines in the
generated images. Although the initial purpose of this loss was to imitate 
brush strokes of ink-wash paintings, it was also effective in
synthesizing other artwork types.
In particular, we used a holistic nested edge detector~\cite{HED} $E$ to obtain
edge maps of the synthesized landscape images $E(x)$ and edge maps of the synthesized
artwork $E(G(x))$. The edge loss is computed as follows:
\begin{align}
    L_{edge} &= \E_{x\sim{X}} \Big(-\frac{1}{N}\sum_{i=1}^N{\mu E(x)_i\log E(G(x))_i} \nonumber \\
         &+ (1-\mu)(1-E(x)_i)\log\left( 1-E(G(x))_i \right) \Big)
    \label{Edge_loss}
\end{align}
\noindent where $N$ is the total number of pixels in the edge map, and $\mu$ is
a balancing weight. The sums of the probabilities for non-edges and edges of every
pixel in $E(x)$ can be computed using $\mu = N_- /N $ and $1-\mu = N_+/N$, $N_-$
and $N_+$, respectively.
 
The objective of the second stage is as follows:
\begin{equation}
L_{total} = L_{adv} + \alpha L_{cycle} + \beta L_{style} + \lambda L_{edge}
\label{equation_losses}
\end{equation}
\noindent where 
$\alpha=10$, $\beta=0.1$,
and $\gamma=10$  control the relative
importance of the individual objectives in all the experiments.

\newcommand{\figsX}[1]{\includegraphics[width=0.18\linewidth]{results/samples_dataset/#1}}
\begin{figure}[t]
\begin{center}
\setlength{\tabcolsep}{1pt}
\begin{tabular}{ccccccc}
 & Label  & Generated &Label  & Generated & Real \\
 & Map & Artwork & Map & Artwork & Artwork \\

\scriptsize
\raisebox{0mm}{ \rotatebox{90}{Ink-wash} } &
\figsX{picture11.jpg} &
\figsX{picture1.jpg} &
\figsX{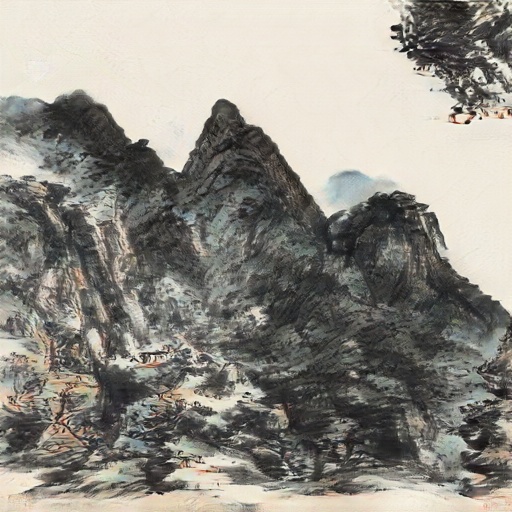} &
\figsX{21727506069.jpg} &
\figsX{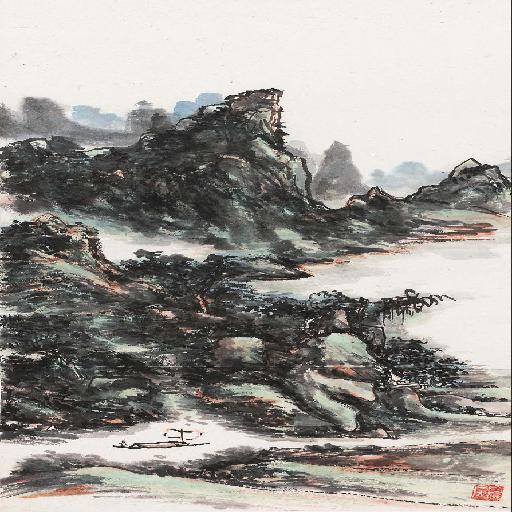} \\

\scriptsize
\raisebox{4mm}{ \rotatebox{90}{Monet} } &
\figsX{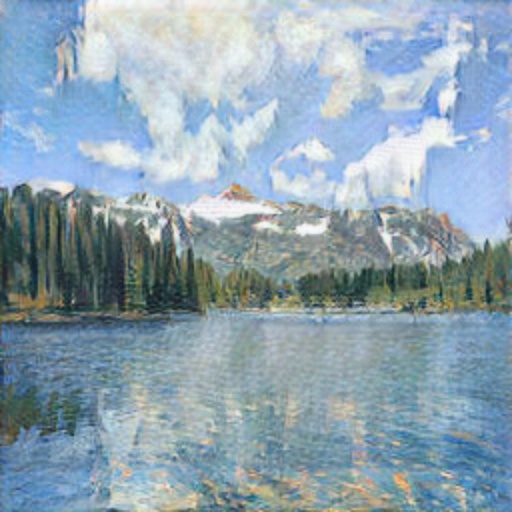} &
\figsX{472845617.jpg} &
\figsX{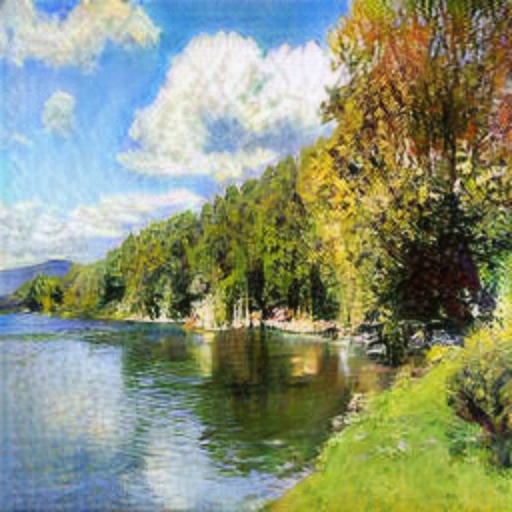} &
\figsX{26163623789.jpg} &
\figsX{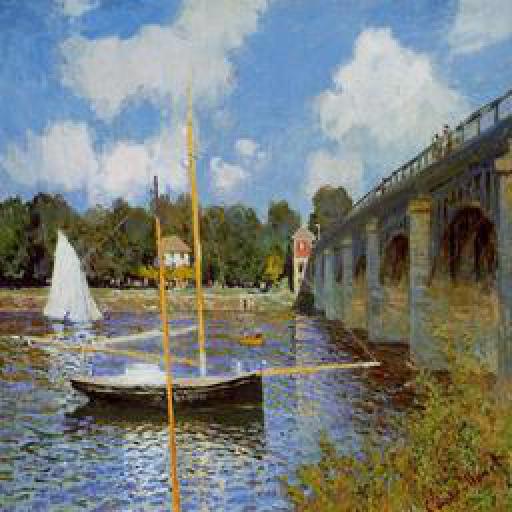} \\

\scriptsize
\raisebox{2mm}{ \rotatebox{90}{Van Gogh} } &
\figsX{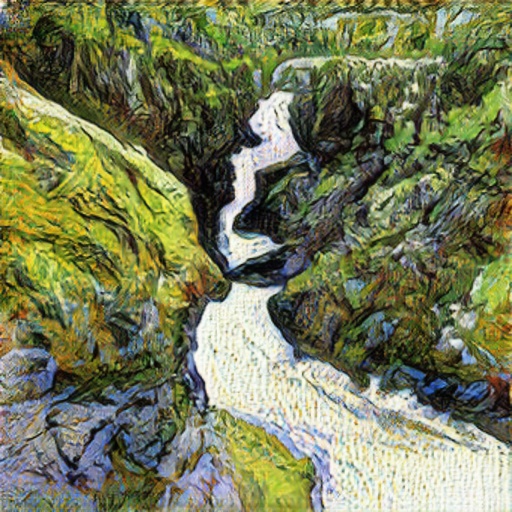} &
\figsX{97330.jpg} &
\figsX{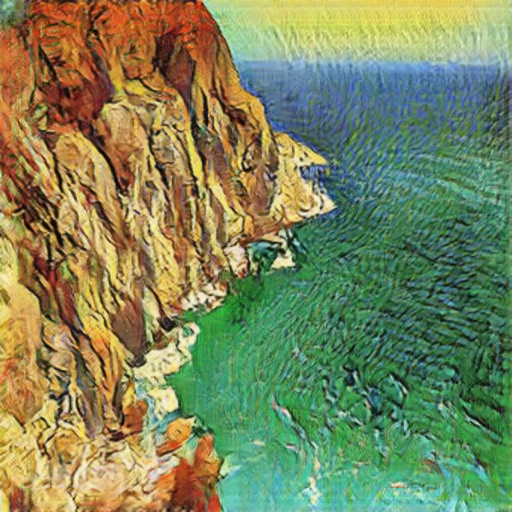} &
\figsX{355068392.jpg} &
\figsX{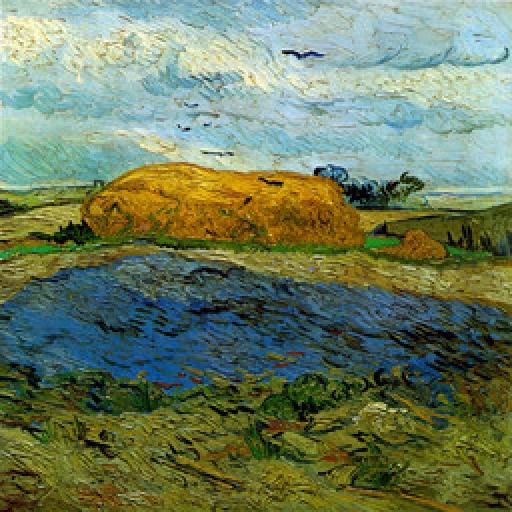} \\

\scriptsize
\raisebox{1mm}{ \rotatebox{90}{Watercolor} } &
\figsX{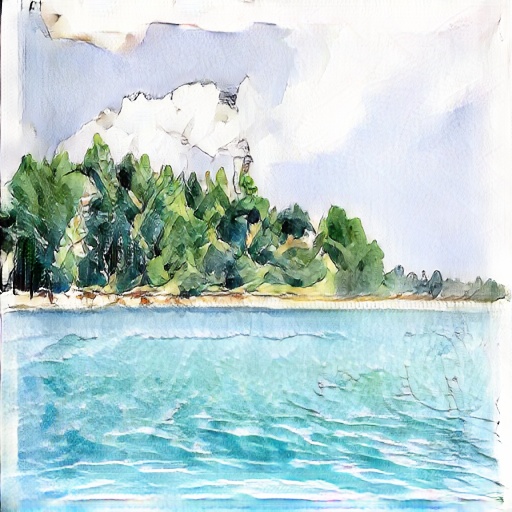} &
\figsX{6306846326.jpg} &
\figsX{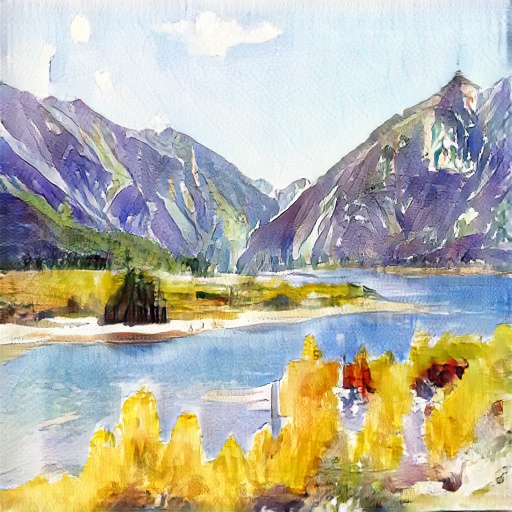} &
\figsX{6176858280.jpg} &
\figsX{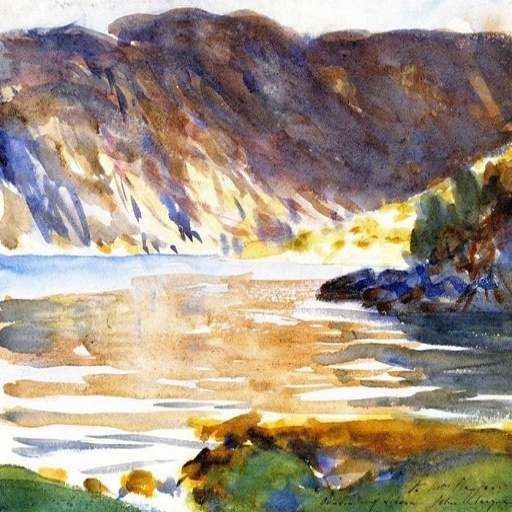} \\

\end{tabular}
\setlength{\tabcolsep}{6pt}
\end{center}
\caption{\textbf{Samples from our \datasetname~dataset and real artworks.}
There is diversity in the supervised label map and artwork images in the four
different artistic styles. }

\label{samples_artwork}
\end{figure}

\begin{figure*}[ht]
    \centering
    \includegraphics[width=\linewidth]{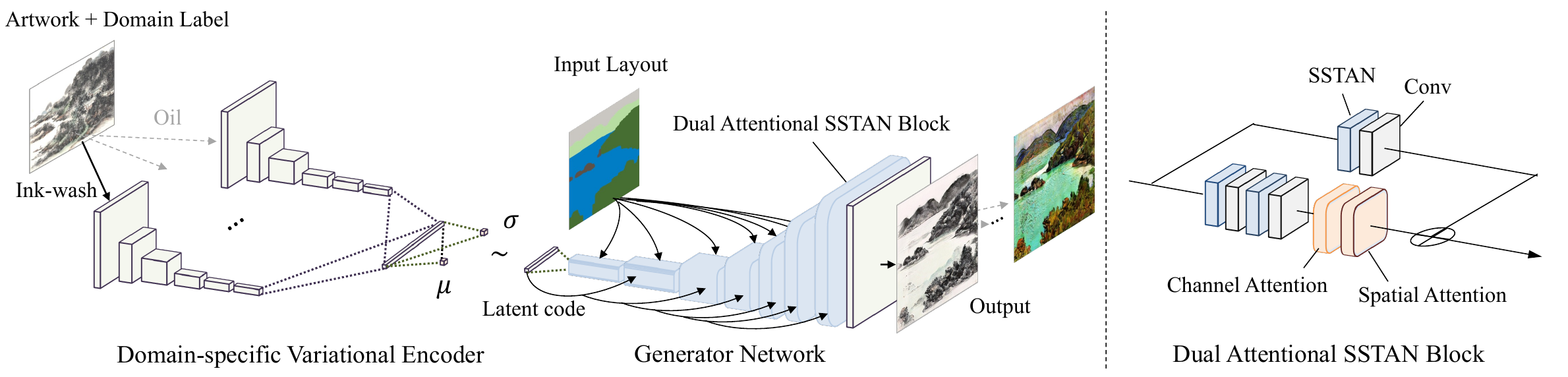}
    \caption{\textbf{Proposed Architecture.} Our model consists of a
    domain-specific variational encoder and a generator network based on SSTAN.
    The encoder extracts the style from an input image as a latent
    code and passes it to the generator to synthesize new artworks with similar
    style appearance while conditioned by an input label map. The
    attention module and SSTAN in the generator
    improve the quality of the generated images.}
    \label{fig:overview_networks}

\end{figure*}

\subsection{Training}
We trained the networks from scratch in all experiments except for the pre-trained
DeepLabV2~\cite{deeplabv2} model for label map generation and 
VGG-19~\cite{VGG} model for style loss computation. The learning rate was 0.0002 in
the first 100 epochs and linearly decayed to zero over the last 50 epochs for
label map generation. For artwork generation, the learning rate was maintained at 0.0002 during the first 100
epochs, and then decayed to zero over the next 100 epochs. All the models were trained on two NVIDIA 1080Ti GPUs with 11GB memory.

\subsection{Dataset Refinement}
\label{refinement}
Following semantic scene generation and weakly supervised artwork
generation, we obtain sets of semantic label maps, synthetic landscape images, and
artwork images in the four domains. However, some synthesized artworks are less
than ideal and can affect the models trained on such data, as shown in
Fig.~\ref{bad_samples}. To reduce data noise, we annotated
2,000 images as either good or bad samples for training, and then trained a VGG
classification model~\cite{VGG} to mimic human annotation. After
training, we ran the model on the dataset and selected the top 10,000 images
using the GMM-QIGA~\cite{gu2020giqa} quality score to obtain four different
sets of paired synthetic landscapes and artworks, one for each domain. Some
examples are presented in Fig.~\ref{samples_artwork}.


\section{Proposed Framework}
We propose CMSAS model
comprising two subcomponents: domain-specific variation encoders
that encodes artwork images into latent vectors and a generator that generates
new artwork based on an input semantic map and encoded latent vector. The
generator was modified to use attention modules in each residual block with
spatially style-adaptive normalization for better generation
quality when synthesizing artistic images. An overview of this model is provided in
Fig.~\ref{fig:overview_networks}. Furthermore, the proposed model could precisely manipulate the synthesized artwork by exploiting the latent space structure.

\subsection{Model Architecture}
We divided the encoder into domain-specific and shared
components. The domain-specific component consists of all the convolutional
layers, while the last two fully connected layers output Gaussian
distribution, characterized by the mean and standard deviation.
In addition, the encoder uses the domain type as input to
activate the domain-specific components.
We use a shared discriminator because our experiments show no 
notable improvements compared to using one per domain.
We also use a shared generator because it is computationally heavy compared with
encoders.
Inspired by~\cite{fu2019dual}, we introduce a dual attention module in each residual block to improve the 
generator performance. Each
dual attention module consists of channel and spatial
attention, computed on the channel and
spatial axes, respectively. The model has seven residual blocks, and the
full layout is given in Fig.~\ref{fig:overview_networks}(right).
For an in-depth overview of the full model, please refer to the supplementary materials.

\begin{figure}[t]
    \centering
    \includegraphics[width=1\linewidth]{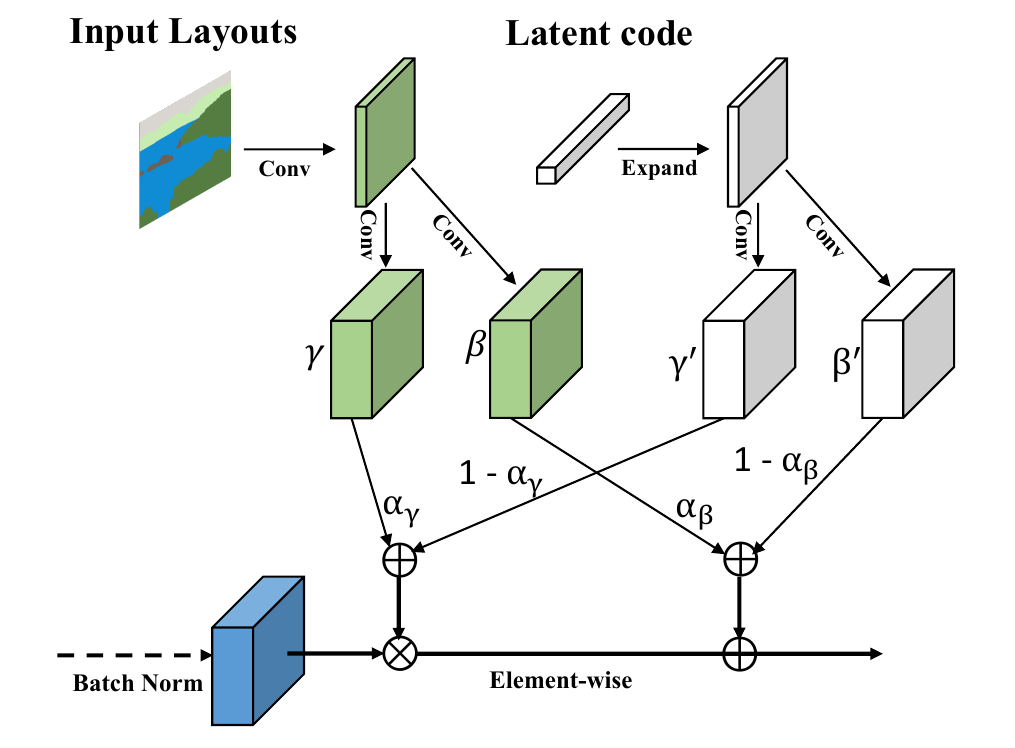}
    \caption{\textbf{Proposed SSTAN module.} The semantic layout is
    projected onto an embedding space and then convolved to produce the
    modulation values $\gamma$ and $\beta$. On the right side, the extracted
    input latent code is first extended to a 2-dimensional space and then
    convolved to produce the modulation values $\gamma^{\prime}$ and
    $\beta^{\prime}$. These are used to modify the features of the neural
    network.}
    \label{fig:Normalization}
\end{figure}

\newcommand{\figsDomain}[1]{\includegraphics[width=0.5\linewidth]{figures/domains/#1}}
\begin{figure}[t]
\begin{center}
\setlength{\tabcolsep}{1pt}
\begin{tabular}{cccc}
\figsDomain{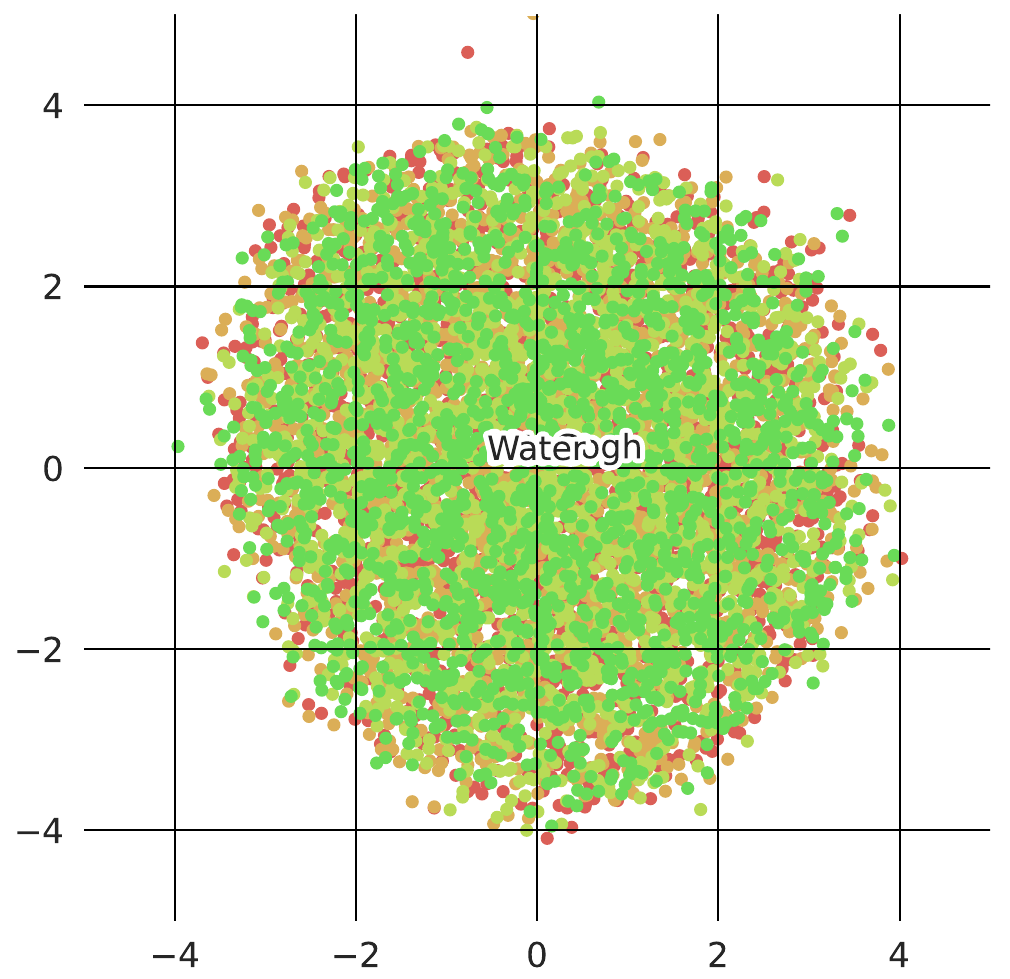} &
\figsDomain{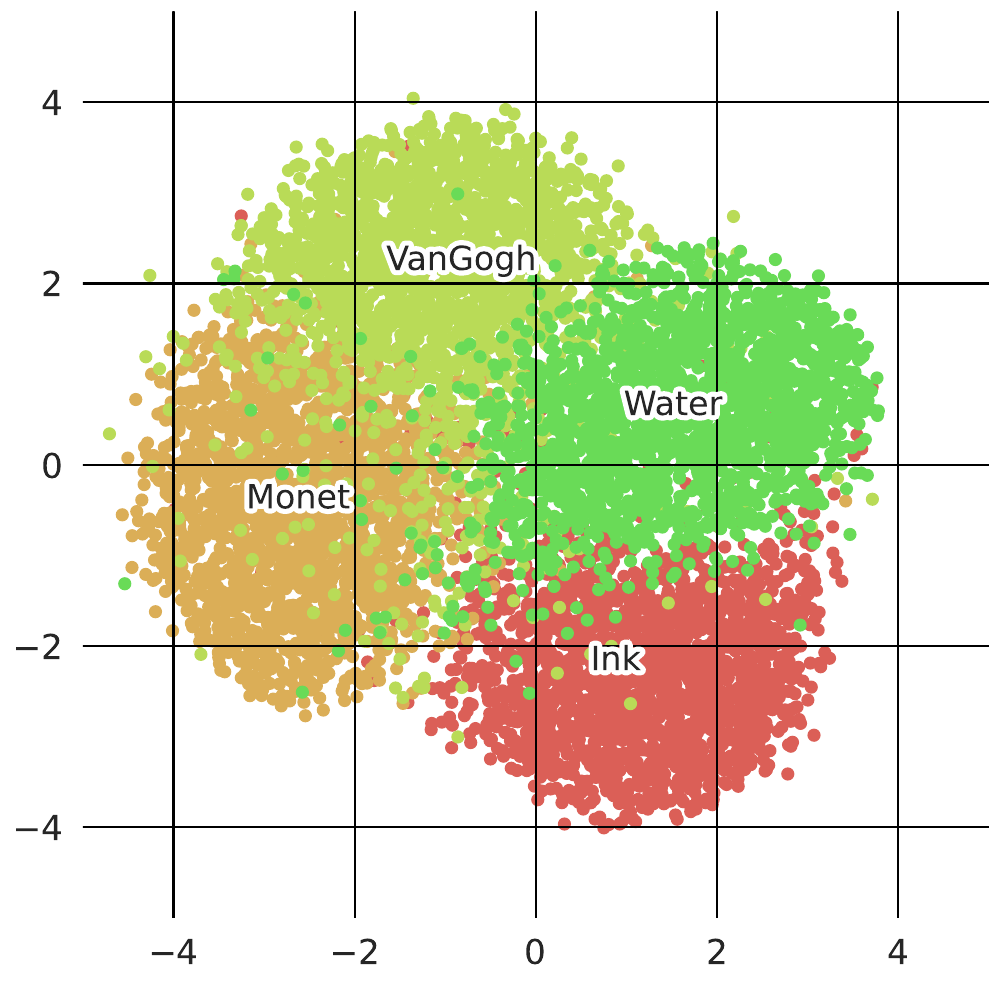} \\
Single Encoder & Domain-Specific Encoders\\
(Four Domains) & (Four Domains) \\

\figsDomain{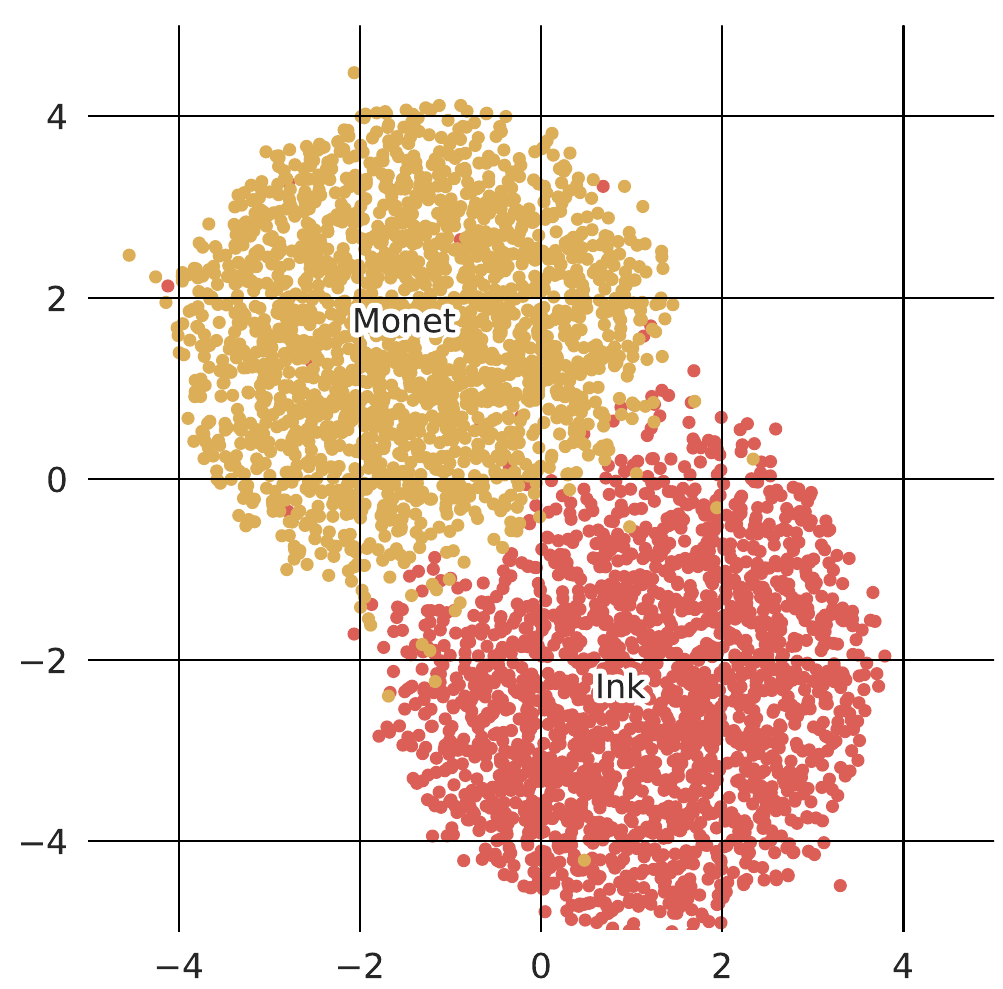} &
\figsDomain{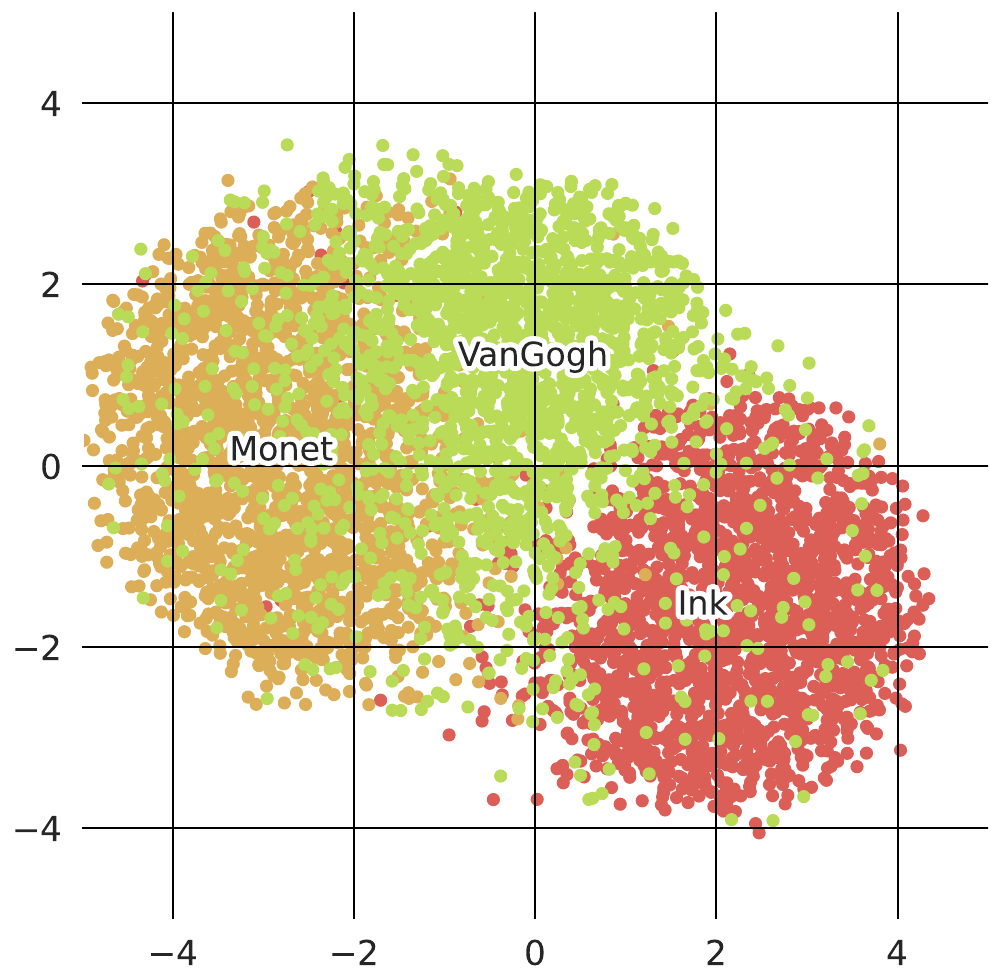} \\
Domain-Specific Encoders & Domain-Specific Encoders\\
(Two Domains) & (Three Domains) \\

\end{tabular}
\setlength{\tabcolsep}{6pt}
\end{center}
\caption{\textbf{Visualization of the latent codes.} By projecting
the latent codes to a two-dimensional space with UMAP, we understand how the latent space is divided into 
different artwork domains using our domain-specific variational encoders. In
contrast, the single encoder fails to separate the different domains.
Comparisons of visualizations trained with different numbers of domains 
show that the separability remains high in all three patterns.}
\label{fig:visualization}
\end{figure}

\begin{figure*}[!ht]
    \centering
    \includegraphics[width=\linewidth]{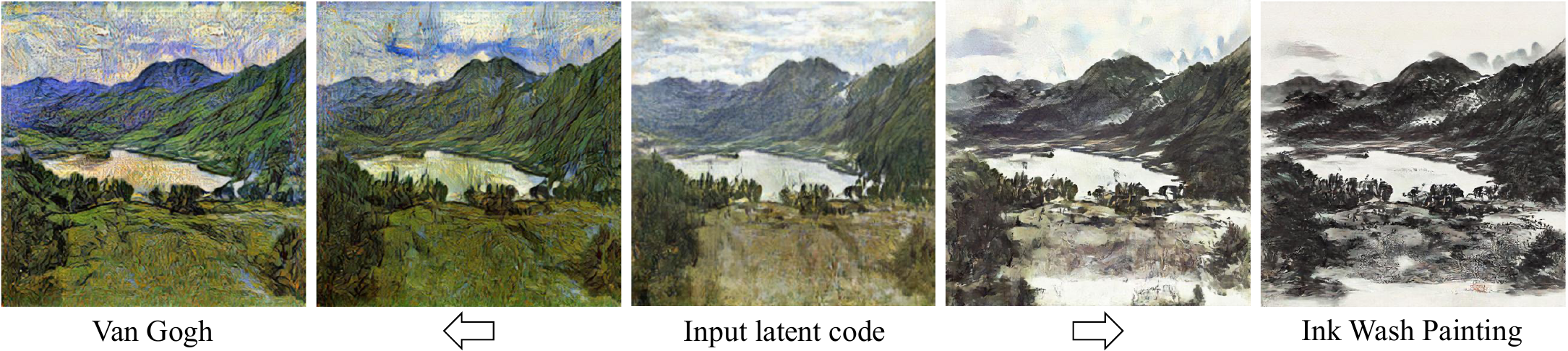}
    \caption{\textbf{Latent space manipulation.}
    For the same semantic map, we interpolate between latent codes of
    different artwork styles to obtain diverse results.}
    \label{fig:Latent_code_manipulation}
\end{figure*}

\subsection{Spatially Style-Adaptive Normalization}
Based on the observation that the generated images from models using the SPADE module are less artistic, we strengthen the influence of the
style information by passing it to the normalization layers of each residual block
instead of just once at the generator input. Therefore, we propose a novel conditional
normalization module called SSTAN to
jointly learn semantic and style representations for higher-quality artwork
generation. The modulation parameters of SSTAN are
tensors with spatial dimensions, making the feature map modulations spatially
adaptive. The latent code was expanded to the same size as the
input semantic layout. Fig.~\ref{fig:Normalization} illustrates the 
SSTAN module structure. The semantic layout and latent code that represent the style of
the input artwork were processed by different convolutional neural networks to
learn modulation parameters. 

The input of each SSTAN block is the latent code $l$ and segmentation mask $M$.
Let $h$ denote the network input activation for a batch of $N$ samples;
let $C$, $H$, and $W$ be the number of channels, height, and width of the
activation map, respectively. The modulated activation value at location $(n \in [1,N], c \in [1,C], y \in [1,H], x \in [1,W])$ becomes
\begin{equation}
\text{SSTAN}(n,c,y,x) = \gamma_{c,y,x}(l,M)\frac{h_{n,c,y,x}-\mu_{c}}{\sigma_{c}} + \beta_{c,y,x}(l,M)
\end{equation}
\noindent where $h_{n,c,y,x}$ is the activation of the previous layer before
normalization; $\mu_{c}$ and $\sigma_{c}$ denote the mean and standard
deviation of the activation in channel $c$, respectively.
\begin{align}
\mu_{c} &= \frac{1}{NHW}\sum_{n,y,x}{h_{n,c,y,x}} \\
\sigma_{c} &=\sqrt{\frac{1}{NHW}\sum_{n,y,x}\left({(h_{n,c,y,x})^2 - (\mu_{c})^2}\right)}
\end{align}
The weighted sums of $\mu_{c}$ and $\sigma_{c}$ are used to modulate the
activations of the generator. Two learnable parameters, $\alpha_\gamma$ and
$\alpha_\beta$, which leverages the weight of each element, are
trained directly to minimize the loss with backpropagation.
The final modulation parameters $\gamma_{c,y,x}$ and $\beta{c,y,x}$ are defined
as follows:
\begin{equation}
\gamma_{c,y,x}(l,M) = \alpha_\gamma\gamma_{c,y,x}{(l)} + (1-\alpha_\gamma)\gamma^{\prime}_{c,y,x}{(M)}
\end{equation}
\begin{equation}
\beta_{c,y,x}(l,M) = \alpha_\beta\beta_{c,y,x}{(l)} + (1-\alpha_\beta)\beta^{\prime}_{c,y,x}{(M)}
\end{equation}

\begin{table}[t]
\centering
\caption{\textbf{Quantitative effect of the number of domains.}
Although performance slightly decreases with the number of domains, joint
training allows for interpolation and more flexible interactivity.}
\begin{tabular}{rcccc}
 \toprule
 Domains & Per & Two & Three & Four\\
 \midrule
 Ink-wash & \textbf{45.95} & 46.14 & 46.17 & 49.33 \\
 Monet & \textbf{63.91} & 64.18 & 65.20 & 65.97 \\
 Van Gogh & 123.90 & $/$ & 125.87 & \textbf{115.3} \\
 Watercolor & \textbf{68.04} & $/$ & $/$ & 68.96 \\
 \bottomrule
\end{tabular}
\label{tbl:domains_FID}
\end{table}

\subsection{Domain and Style Control}
\label{DomainControl}
To provide more control over the output artwork images, we assumed that the latent vectors of the different domains are separable with a
hyperplanes~\cite{interfaceGAN}. 
We used a uniform manifold approximation and projection (UMAP~\cite{UMAP}), a
dimensionality reduction tool for projecting latent vectors in low-dimensional space
for intuitive visualization. 
As shown in Fig.~\ref{fig:visualization}, unlike a single encoder,
which learns entangled representations that are less separable in their latent space,
our domain-specific encoders learn disentangled spaces that can be 
better separated into domains.
Consequently, this allows our proposed model to achieve latent space manipulation to specify the domain of the generated artwork and perform
cross-domain style morphing between different domains.
Table~\ref{tbl:domains_FID} shows that the FID score of the generated artwork decreased
only slightly as the number of domains increased. 

For each latent code $z \in
\mathcal{Z}$, our generator can be seen as a mapping $f_g\colon \mathcal{Z}
\rightarrow \mathcal{A}$, where $\mathcal{A}$ is the manifold of artworks. For
each domain, we defined a scoring function $f_s \colon \mathcal{A} \rightarrow
[0,1]$ that predicts whether an artwork corresponds to a particular
domain. We then formulate the problem of finding a normal $n$
of a hyperplane that separates the latent codes based on their codes as
follows:
\begin{equation}
\argmax_{n} \max\left( 0, t\left(n^\intercal z \right) \right) \quad\text{subject to}\quad \|n\|=1
\end{equation}
\noindent where
\begin{equation}
t = \begin{cases}
    1 & f_s\left(f_g(z)\right) > 0.5 \\
   -1 & otherwise
\end{cases}
\end{equation}

We can then control the output domain of the artwork by moving 
towards a particular domain. This can be achieved by adjusting a value $\alpha$ along the normal
direction to obtain a new latent vector: \mbox{$z' = z + \lambda n$}, as shown
in Fig.~\ref{fig:Latent_code_manipulation}. This approach enables us to
generate the artwork without the explicit need for an input style image by setting
$z=0$ and a larger value of $\lambda$.

In particular, we implemented $f_s$ using a VGG classifier~\cite{VGG} trained to
predict a domain from 40,000 synthesized images in the proposed dataset.

\subsection{Objective Function}
Our loss functions include adversarial, feature matching
~\cite{highresolutionpix2pix}, perceptual
~\cite{johnson2016perceptual}, and KL divergence losses.

\noindent\textbf{Adversarial loss.} Let $E$ be the domain-specific variation encoders,
$G$ be the dual attentional SSTAN generator, and
$D_k$ be the $k$-th discriminator at the different scales.
Given an input artwork $A$, a latent code $l$, that represents the style of $A$ and 
are extracted by $E$, can be defined as $l=E(A)$.
Let $M$ be the corresponding segmentation mask for $A$.
The adversarial loss is then defined as
\begin{align}
    L_{adv} &= \E[\max(0,1-D_k(A,M))] \nonumber \\
         &+ \E[\max(0,1-D_k(G(l,M),M))]
    \label{AD_loss}
\end{align}
where $G$ attempts to minimize the objective against an adversarial $D_k$ that attempts to maximize it during the training.

\noindent\textbf{Feature matching loss.} Let $T=2$ be the total number
of layers in discriminator $D_k$, $D_k^{(i)}$ and $N_i$ be the output
feature maps and the number of elements in the $i$-th layer of
$D_k$, respectively. We formulate the feature matching loss $L_{FM}$ as
\begin{align}
	& L_{FM} = \nonumber \\
 &\E\sum_{i=1}^T -\frac{1}{N_i}[\|D_k^{(i)}(A, M) - D_k^{(i)}(G(l, M), M)\|_1]
    \label{FM_loss}
\end{align}
where $G$ attempts to minimize this objective against an adversarial $D_k$ that attempts to maximize it during the training.

\noindent\textbf{Perceptual loss.}
Let $N$ be the total number of layers used to calculate the perceptual loss, 
$F^{(i)}$ be the output feature maps of the $i$-th layer of the VGG~\cite{VGG} network, and
$M_i$ be the number of elements of $F^{(i)}$. We define the perceptual loss $L_{P}$ as:
\begin{align}
	L_{P} = \E\sum_{i=1}^N -\frac{1}{M_i}[\|F^{(i)}(A) - F^{(i)}(G(l, M))\|_1]
    \label{perceptual_loss}
\end{align}

\noindent\textbf{KL Divergence loss.}
Let $p(z)$ be a standard Gaussian distribution. The variational distribution $q$
is fully determined by a mean vector $\mu$ and a variance vector$\sigma$,
which are the output of our encoder as shown in Fig.~\ref{fig:overview_networks}.
We use the reparameterization trick~\cite{kldloss} for backpropagating
the gradient from the generator to the encoder. 

\begin{align}
	L_{KLD} &= D_{KL}(q(z|x))\|p(z))
    \label{KLD_loss}
\end{align}

\noindent\textbf{Full Objective.}
Our full objective is as follows:
\begin{align}
	L_{full} = L_{adv} & + \lambda_1L_{FM} + \lambda_2L_{P}  + \lambda_3L_{KLD}
    \label{Full_loss}
\end{align}

\noindent where 
$\lambda_1=10$, $\lambda_2=10$,
and $\lambda_3=0.01$ control the relative
importance of individual objectives in all the experiments.


\begin{table}[t]
\caption{\textbf{Quantitative evaluation.} We compare the proposed method with existing approaches
for conditional image generation using the FID metric. The baseline models
SMIS, OASIS, and Co-Mod can not specify the domain of generated artworks; therefore
we only evaluate their results in a mix of 4 domains. The best results are
highlighted in bold.}

\centering
\begin{tabular}{>{\raggedleft\arraybackslash}p{1.2cm}	
>{\centering\arraybackslash}p{0.85cm}
>{\centering\arraybackslash}p{0.85cm}
>{\centering\arraybackslash}p{1.15cm}
>{\centering\arraybackslash}p{0.85cm}
>{\centering\arraybackslash}p{0.85cm}
}
 \toprule
 Domains & SMIS & OASIS & Co-Mod & SEAN & CMSAS \\
 & & & GAN & & (Ours)\\
 \midrule
 Ink-wash & - & - & - & 86.47 & \textbf{49.33} \\
 Monet & - & - & - & 84.47 & \textbf{65.97} \\
 Van Gogh & - & - & - & 145.74 & \textbf{115.3} \\
 Watercolor & - & - & - & 106.37 & \textbf{68.96} \\
 Mixed & 97.38 & 78.04 & 67.38 & 89.52 & \textbf{55.03} \\
 \bottomrule
\end{tabular}
\label{FID_results}
\end{table}

\begin{table}
\caption{\textbf{Quantitative evaluation with style transfer algorithms.} We
compare the proposed method with existing approaches for neural style transfer using the FID
metric. The best results are highlighted in bold.}

\centering
\begin{tabular}{>{\raggedleft\arraybackslash}p{1.2cm}	
>{\centering\arraybackslash}p{0.85cm}
>{\centering\arraybackslash}p{0.85cm}
>{\centering\arraybackslash}p{1.15cm}
>{\centering\arraybackslash}p{0.85cm}
>{\centering\arraybackslash}p{0.85cm}
}
 \toprule
 Domains & NST & AdaIN & STROTSS & SEAN & CMSAS \\
 &  & & &  & (Ours) \\
 \midrule
 Ink-wash & 76.15 & 96.59 & 75.26 & 62.33 & \textbf{45.35} \\
 Monet & 94.02 & 93.25 & 76.23 & 75.74 & \textbf{67.4} \\
 Van Gogh & 155.17 & 161.53 & 128.29 & 123.36 & \textbf{116.2} \\
 Watercolor & 109.13 & 124.53 & 75.25 & 73.66 & \textbf{66.0} \\
 \bottomrule
\end{tabular}
\label{FID_results_with_ref}
\end{table}

\section{Results}
We evaluated our approach and performed qualitative and
quantitative comparisons with other methods.

\subsection{Baselines}
We chose three existing models as baselines for multimodal semantic
image synthesis: SMIS~\cite{zhu2020SMIS}, OASIS~\cite{OASIS},
and Co-Mod GAN models~\cite{COMOD}. For generation within specific domains,
we use the SEAN~\cite{zhu2020sean} model as our baseline model. Alls
baseline models have shown an impressive ability to synthesize photorealistic
images in different styles. For a fair comparison, all models are trained using
our proposed dataset for semantic artwork synthesis. Furthermore, all 
baseline models were trained using the implementations provided by the authors.
We chose three style transfer methods as our baseline models for the
reference-based generation comparison, which are neural style
transfer~\cite{styletransfer}, AdaIN~\cite{AdaIN} and STROTSS~\cite{STROTSS}.

\newcommand{\figsM}[1]{\includegraphics[width=0.102\linewidth]{results/multimodal/#1}}
\begin{figure*}[p]
\begin{center}
\setlength{\tabcolsep}{1pt}
\begin{tabular}{ccccccccc}

Input &  \multicolumn{2}{c}{SMIS} & \multicolumn{2}{c}{OASIS} &  \multicolumn{2}{c}{Co-Mod} & \multicolumn{2}{c}{CMSAS} \\
  &  \multicolumn{2}{c}{} & \multicolumn{2}{c}{} & \multicolumn{2}{c}{GAN} &  \multicolumn{2}{c}{(Ours)}\\

\figsM{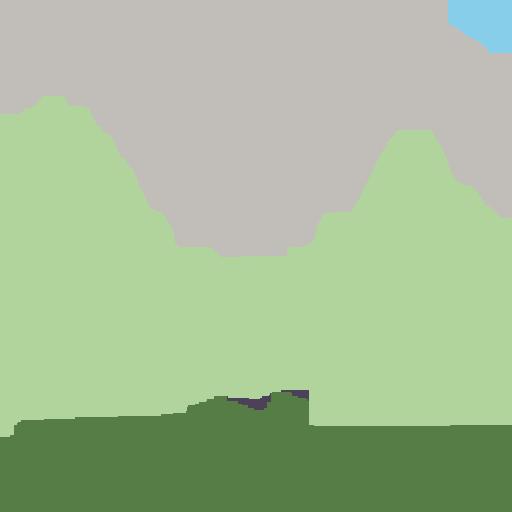} &~
\figsM{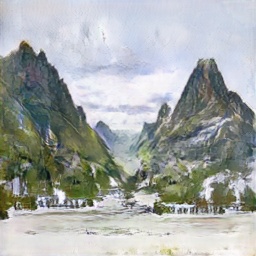} &
\figsM{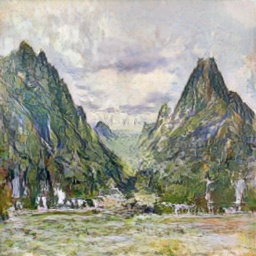} &~
\figsM{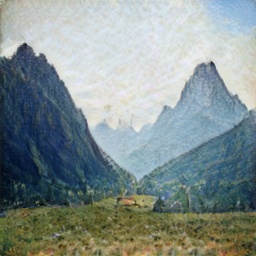} &
\figsM{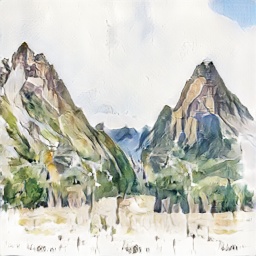} &~
\figsM{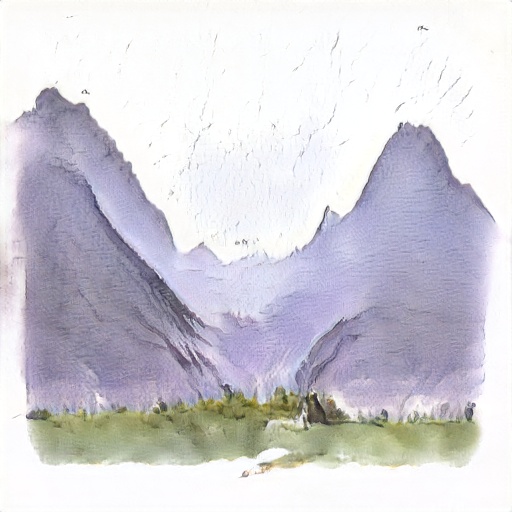} &
\figsM{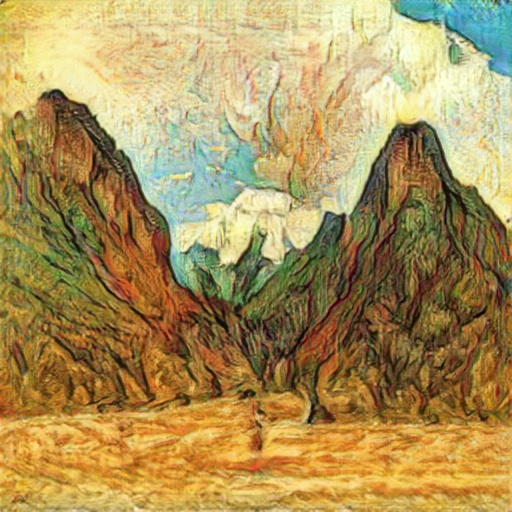} &~
\figsM{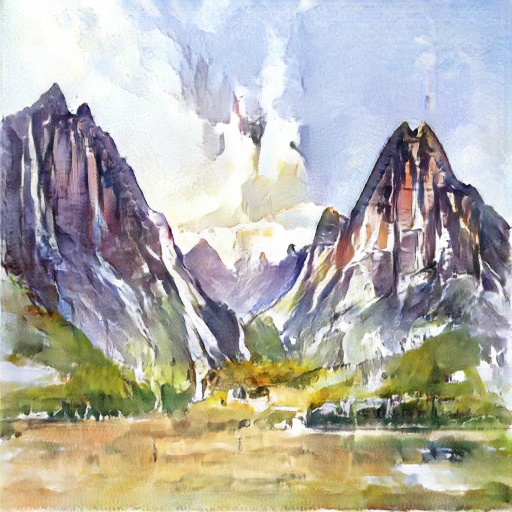} &
\figsM{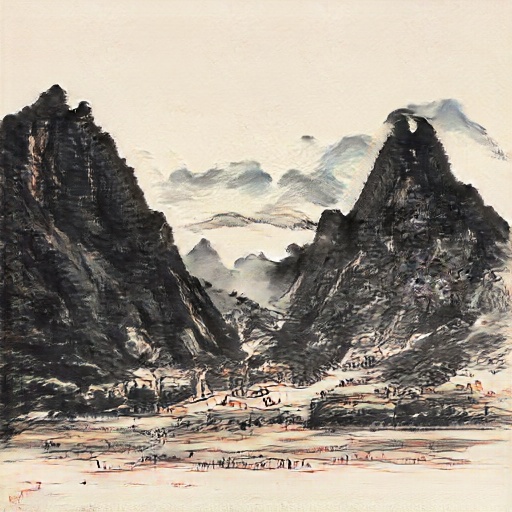} \\
 &
\figsM{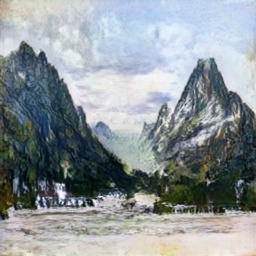} &
\figsM{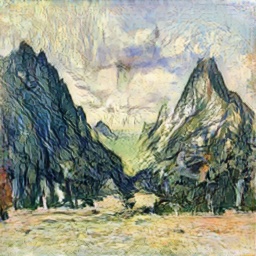} &~
\figsM{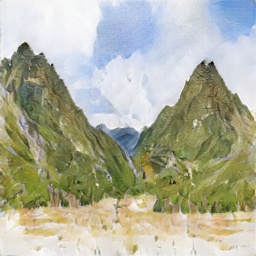} &
\figsM{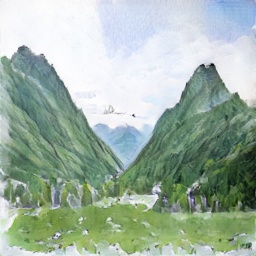} &~
\figsM{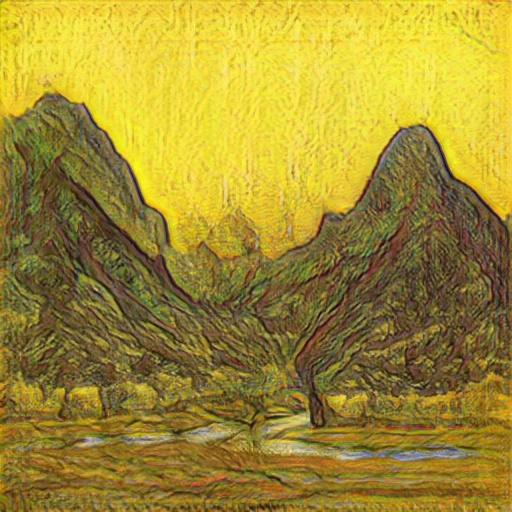} &
\figsM{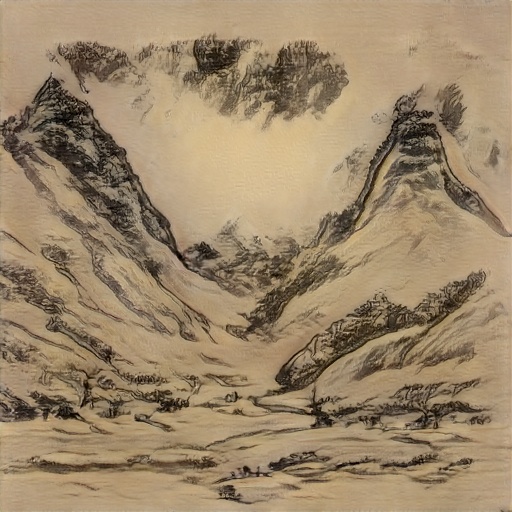} &~
\figsM{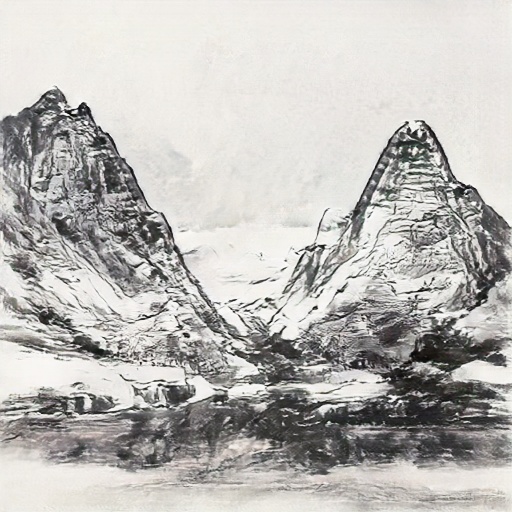} &
\figsM{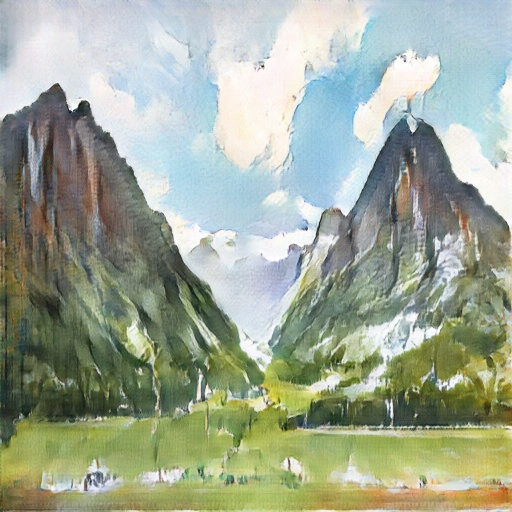} \\
&
\figsM{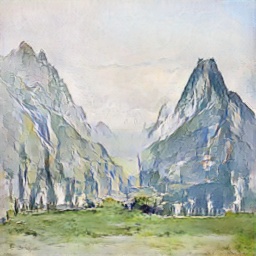} &
\figsM{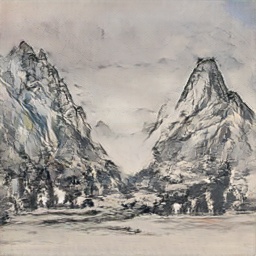} &~
\figsM{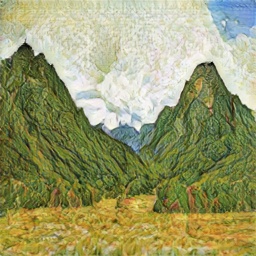} &
\figsM{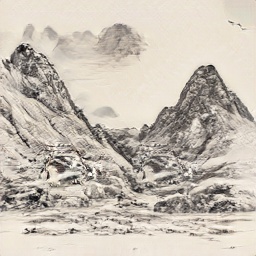} &~
\figsM{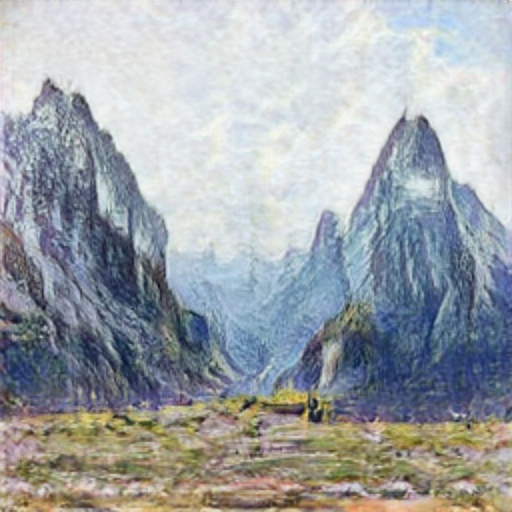} &
\figsM{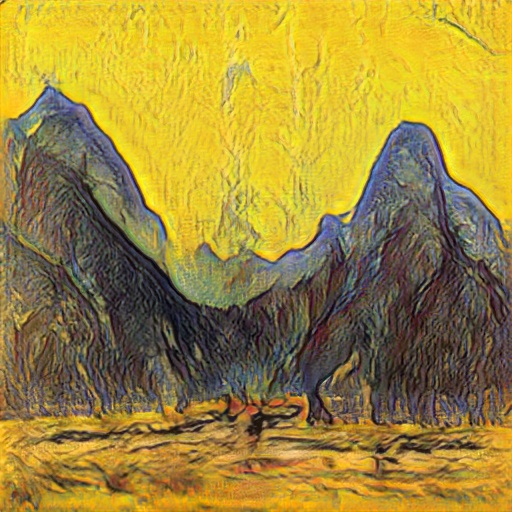} &~
\figsM{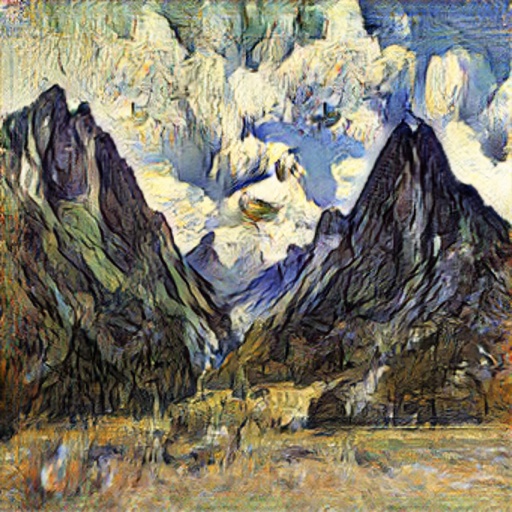} &
\figsM{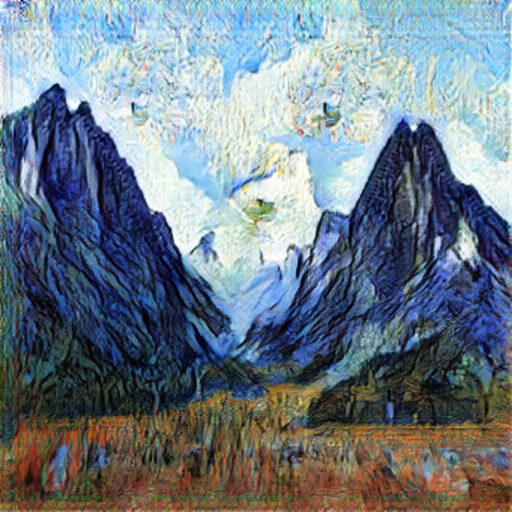} \\

\figsM{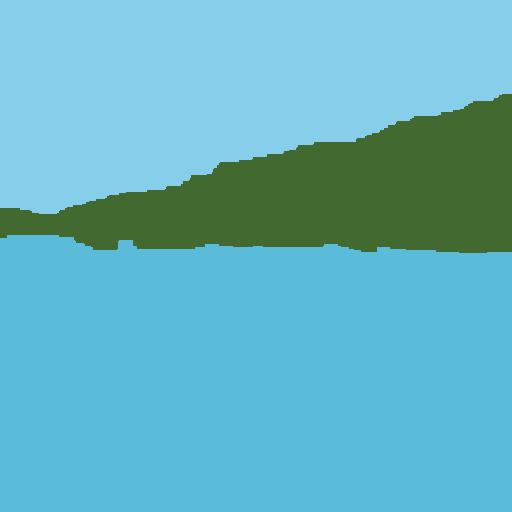} &~
\figsM{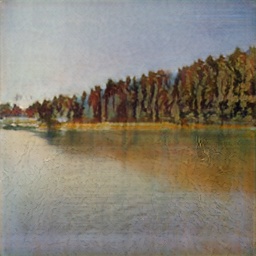} &
\figsM{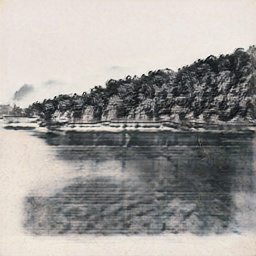} &~
\figsM{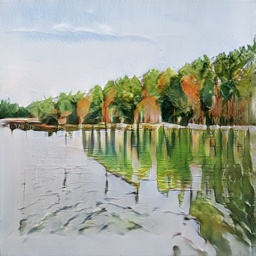} &
\figsM{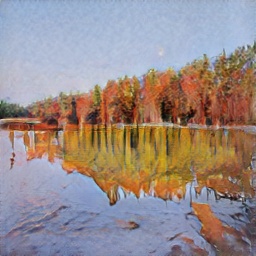} &~
\figsM{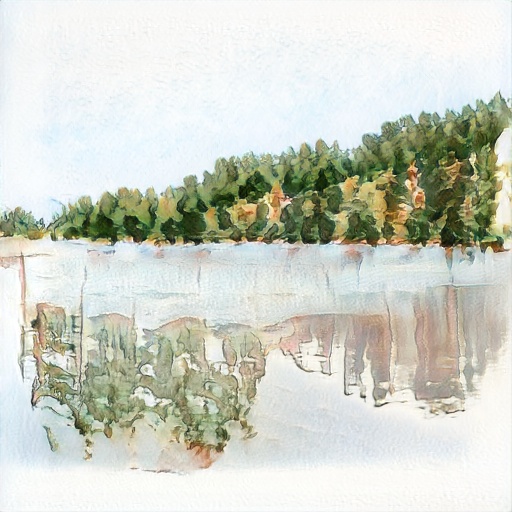} &
\figsM{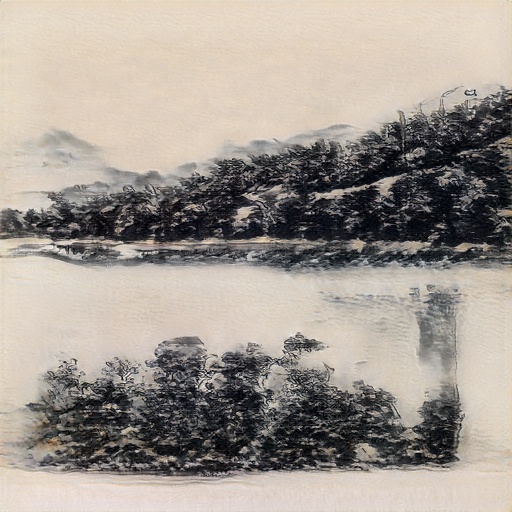} &~
\figsM{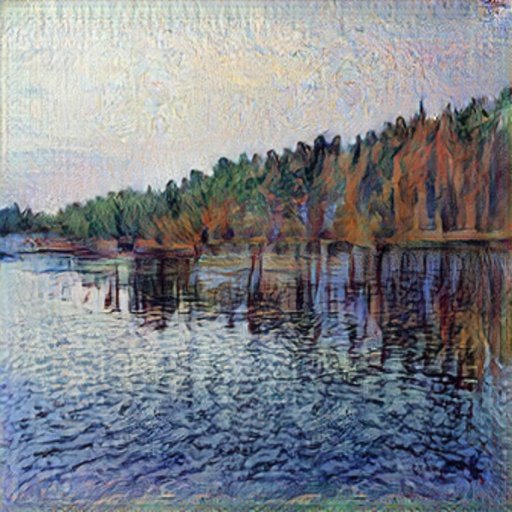} &
\figsM{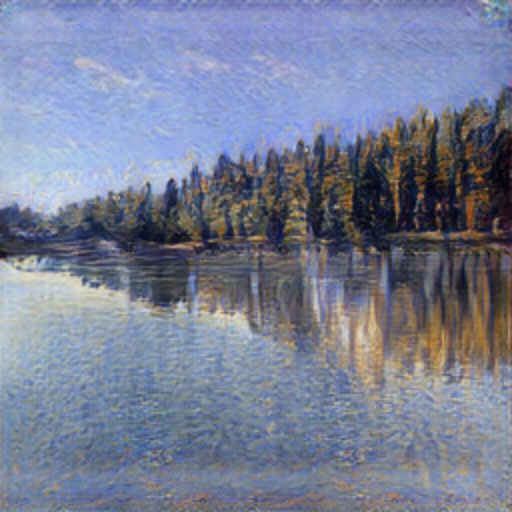} \\
 &
\figsM{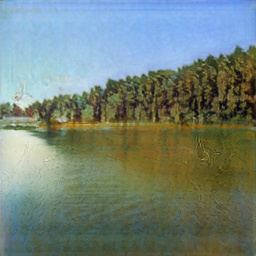} &
\figsM{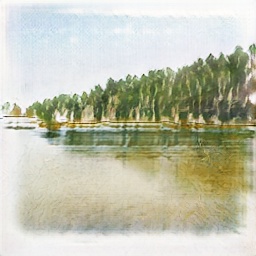} &~
\figsM{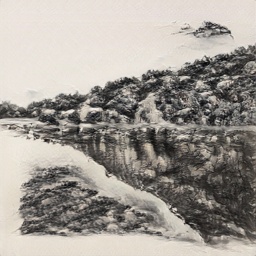} &
\figsM{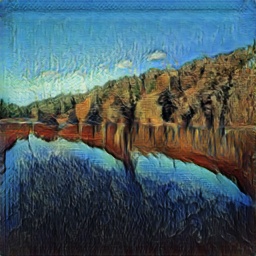} &~
\figsM{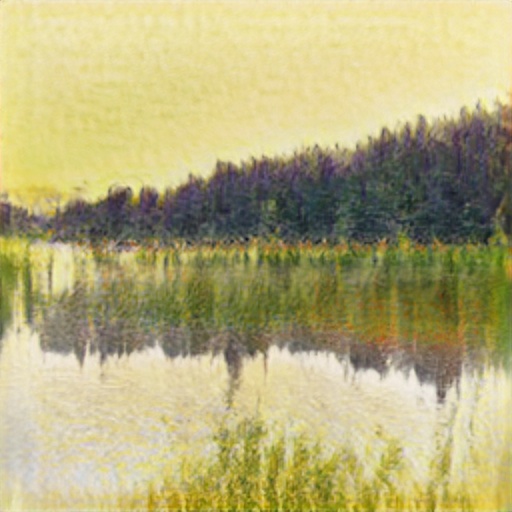} &
\figsM{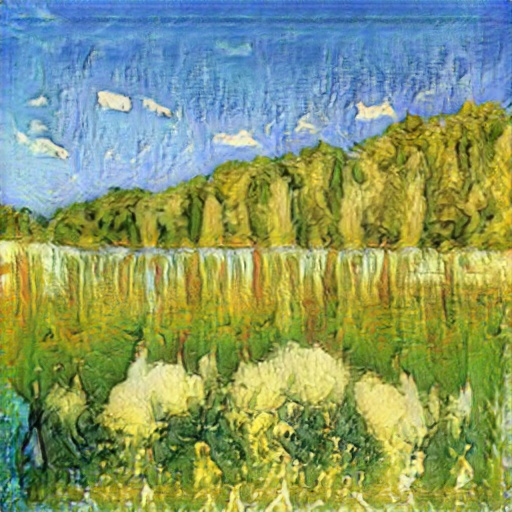} &~
\figsM{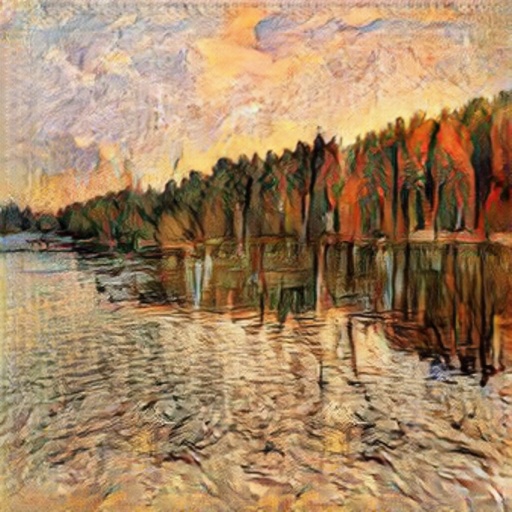} &
\figsM{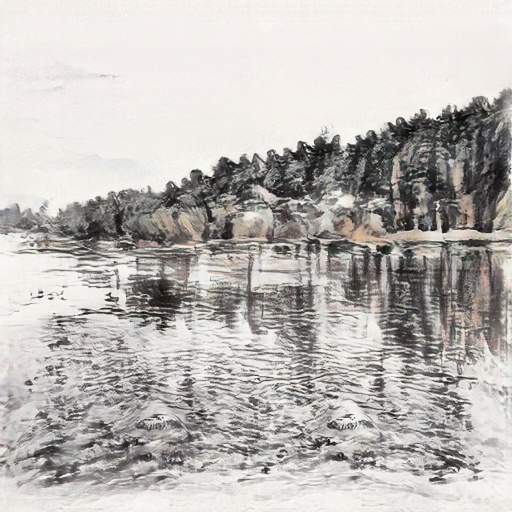} \\
&
\figsM{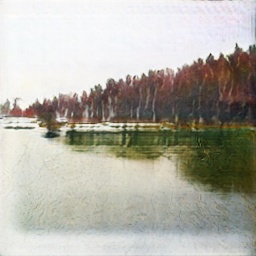} &
\figsM{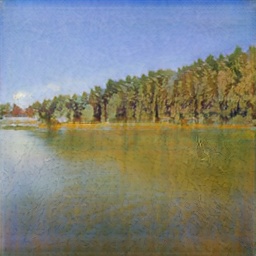} &~
\figsM{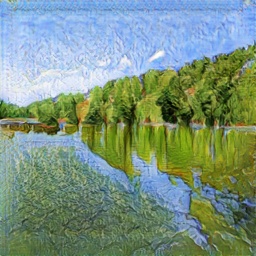} &
\figsM{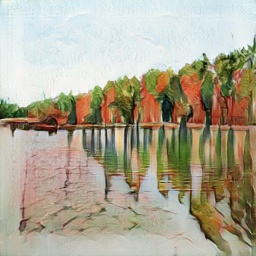} &~
\figsM{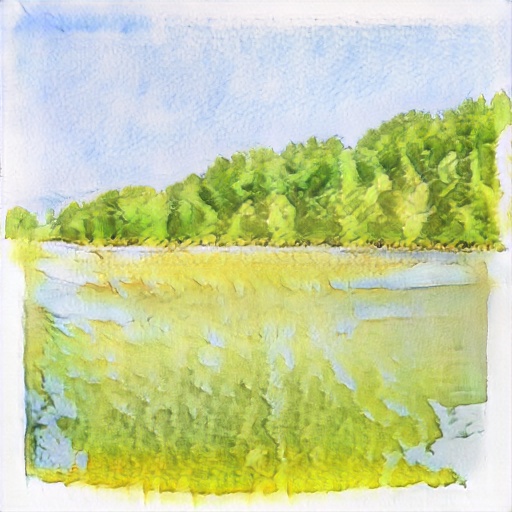} &
\figsM{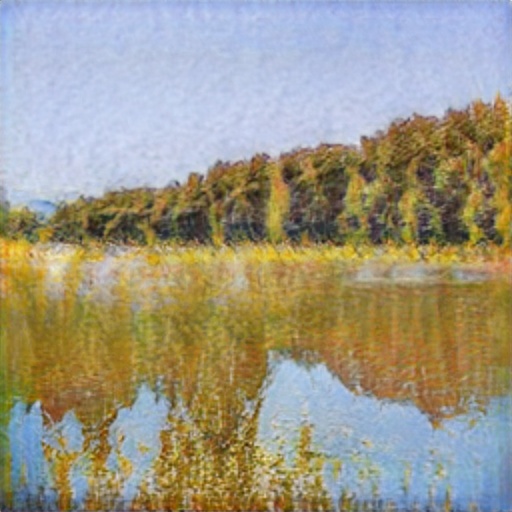} &~
\figsM{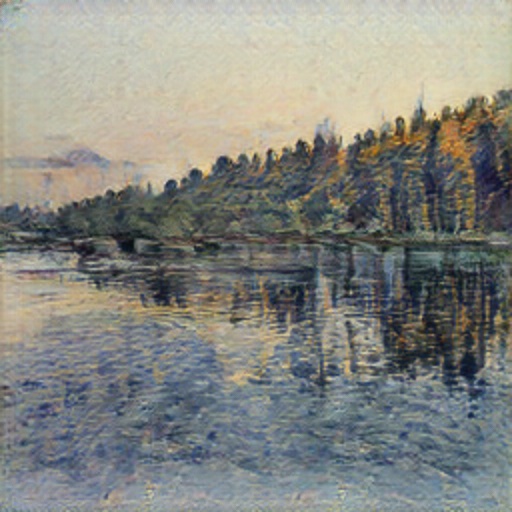} &
\figsM{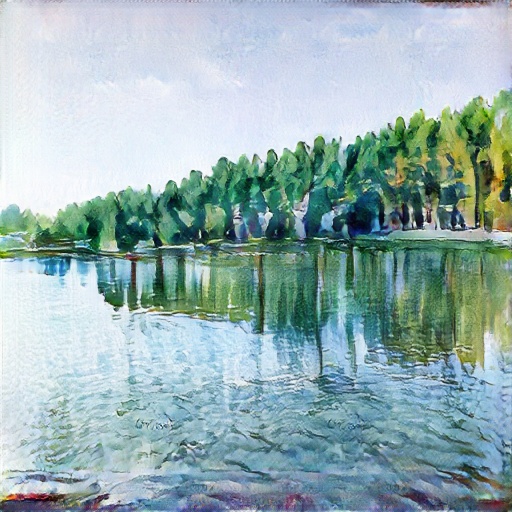} \\

\figsM{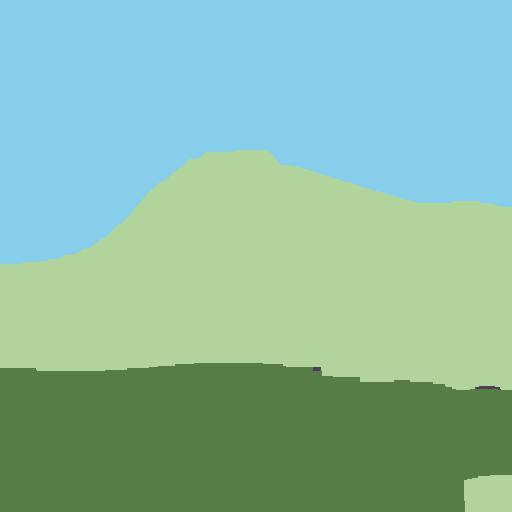} &~
\figsM{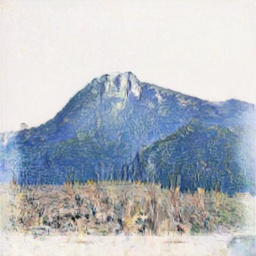} &
\figsM{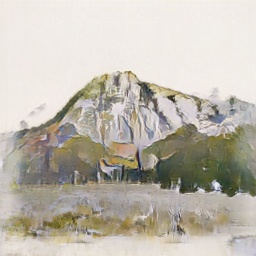} &~
\figsM{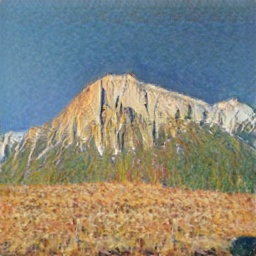} &
\figsM{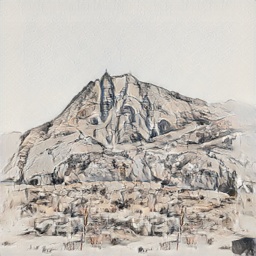} &~
\figsM{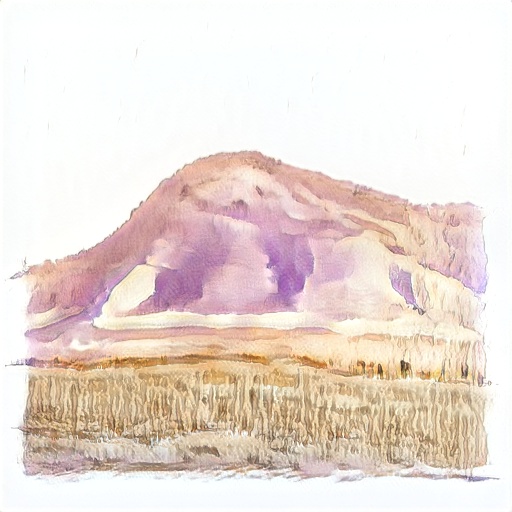} &
\figsM{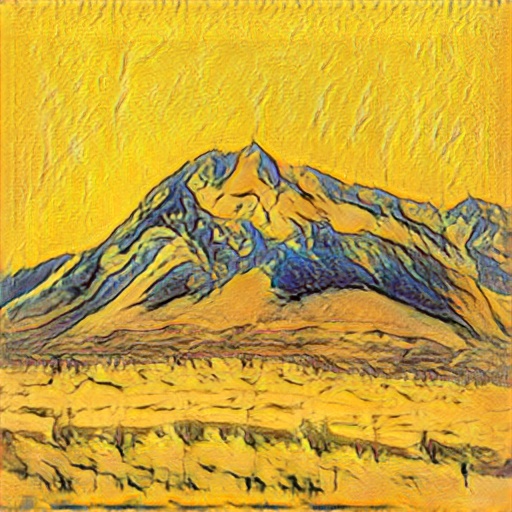} &~
\figsM{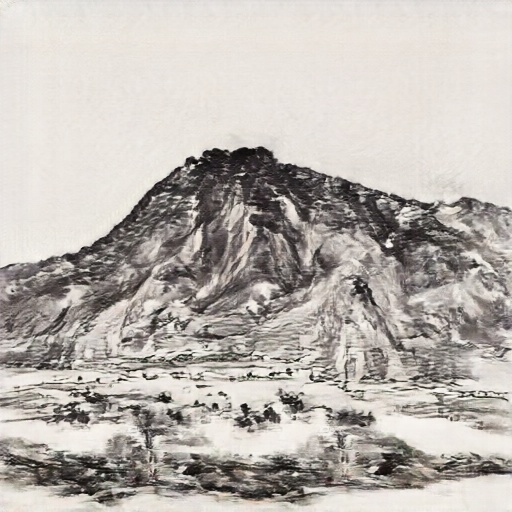} &
\figsM{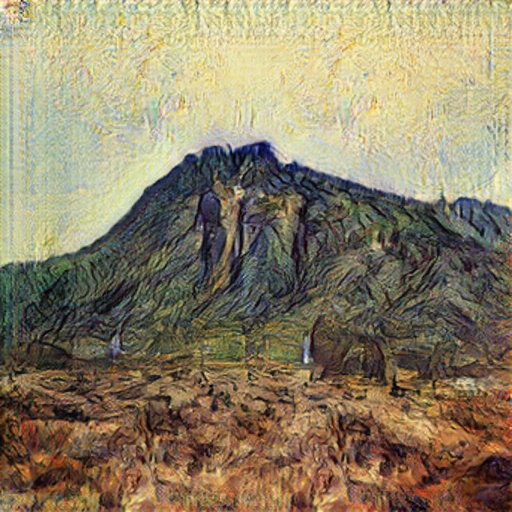} \\
 &
\figsM{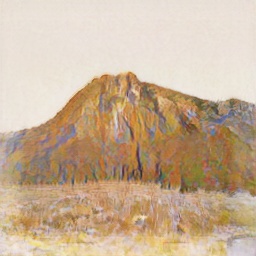} &
\figsM{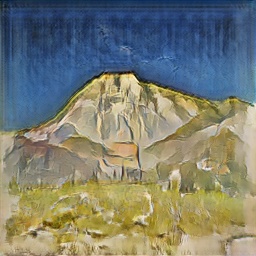} &~
\figsM{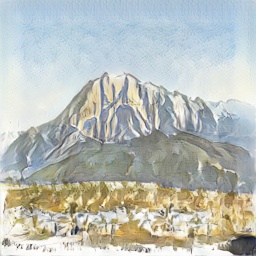} &
\figsM{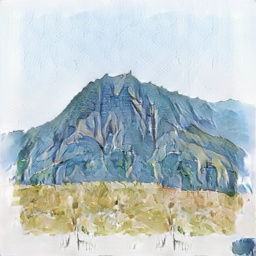} &~
\figsM{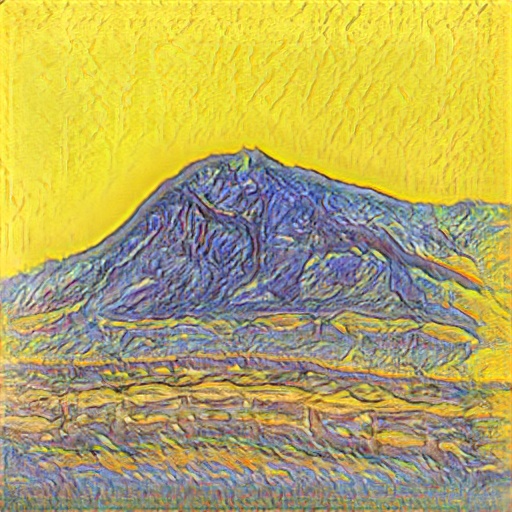} &
\figsM{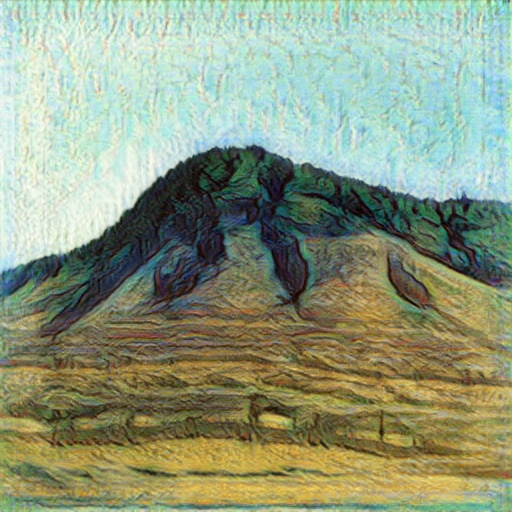} &~
\figsM{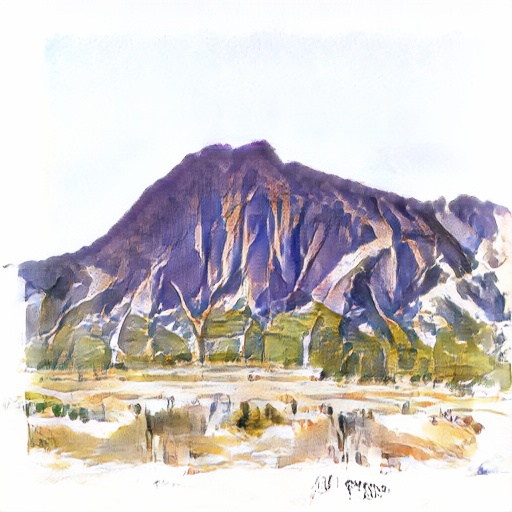} &
\figsM{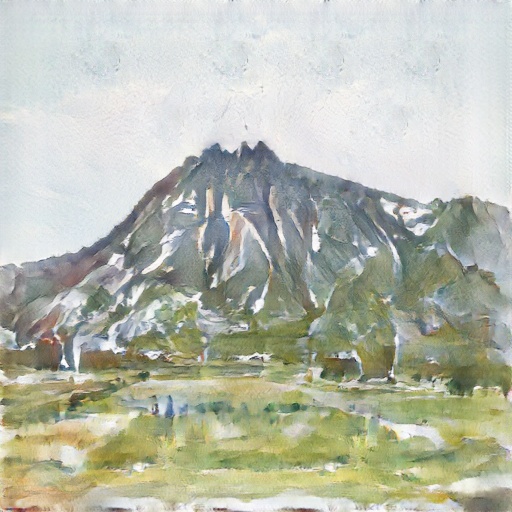} \\
&
\figsM{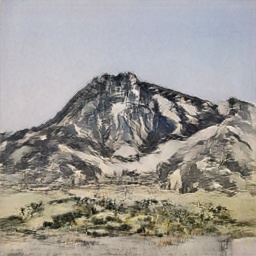} &
\figsM{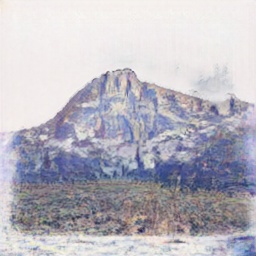} &~
\figsM{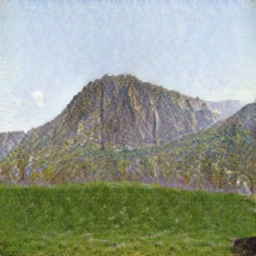} &
\figsM{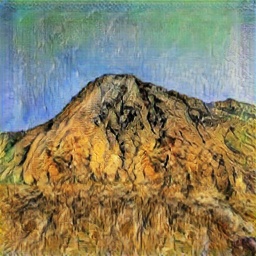} &~
\figsM{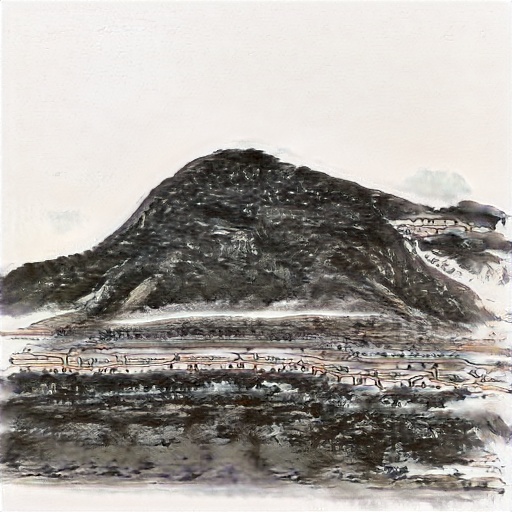} &
\figsM{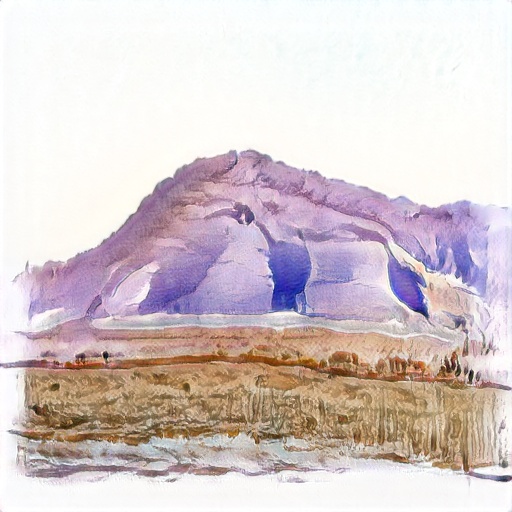} &~
\figsM{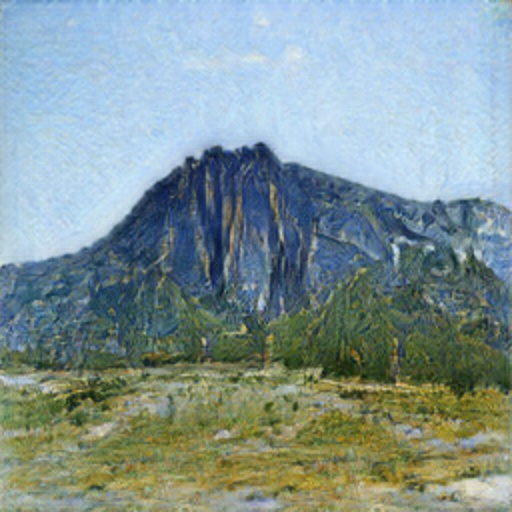} &
\figsM{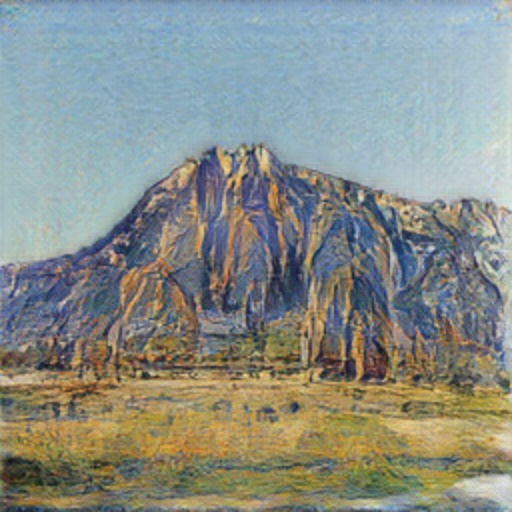} \\

\end{tabular}
\setlength{\tabcolsep}{6pt}
\end{center}
\caption{\textbf{Qualitative comparison of multimodal generations.} 
Our proposed model generates a higher quality of texture details and style
intensity compared to the baseline models. The SIMS, OASIS,
and Co-ModGAN models cannot generate artwork in a
pre-specified domain due to using a random input vector. The domain-specific
encoders used in our proposed method provide improved controllability via latent
space manipulation, as discussed in Sec.~\ref{DomainControl}.}

\label{fig:multimodal}
\end{figure*}

\newcommand{\figsB}[1]{\includegraphics[width=0.158\linewidth]{results/domain_specific/#1}}
\begin{figure*}[ht]
\begin{center}
\setlength{\tabcolsep}{1pt}
\begin{tabular}{cccccccc}
& Input & SEAN & CMSAS~(Ours) & Input & SEAN & CMSAS~(Ours) \\

\raisebox{1mm}{ \rotatebox{90}{~~Ink-wash} } &
\figsB{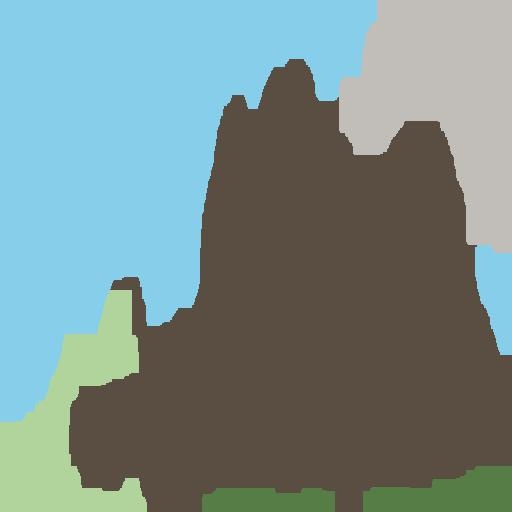} &
\figsB{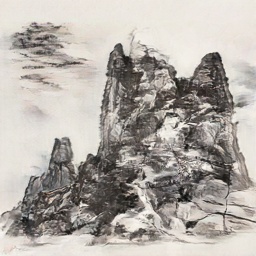} &
\figsB{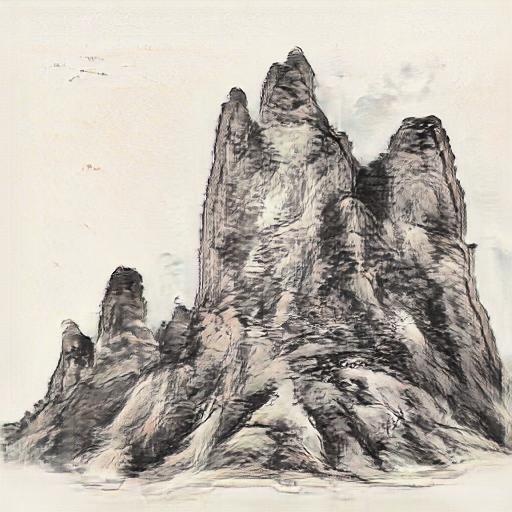} &
\figsB{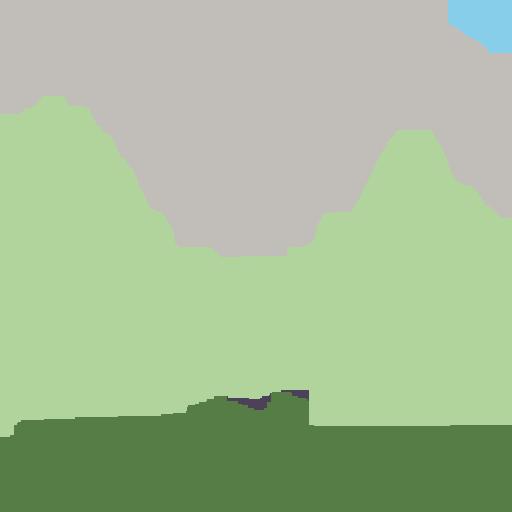} &
\figsB{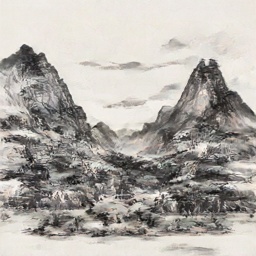} &
\figsB{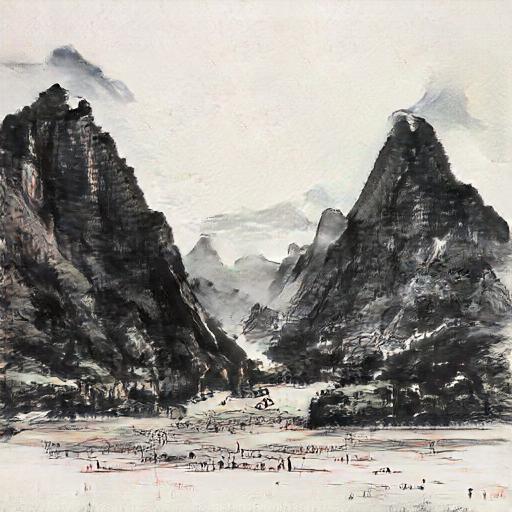} \\
\raisebox{1mm}{ \rotatebox{90}{~~~~~Monet} } &
\figsB{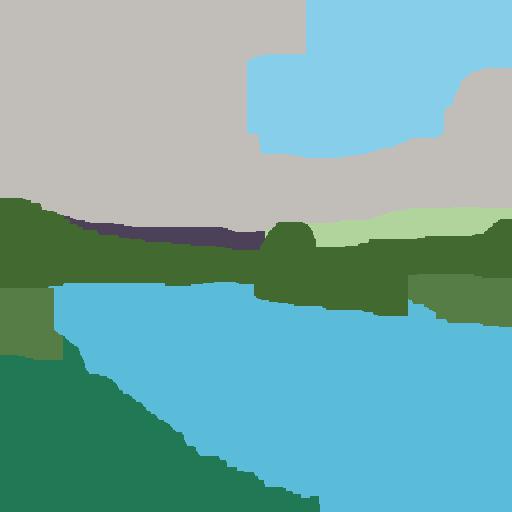} &
\figsB{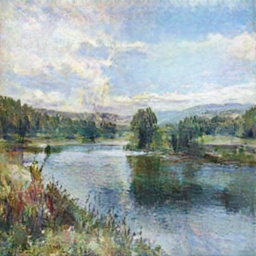} &
\figsB{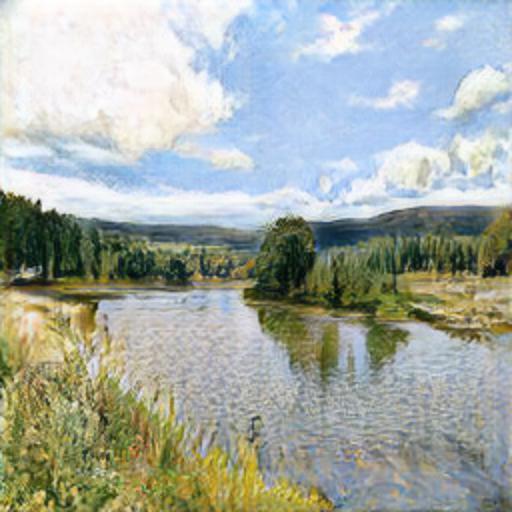} &
\figsB{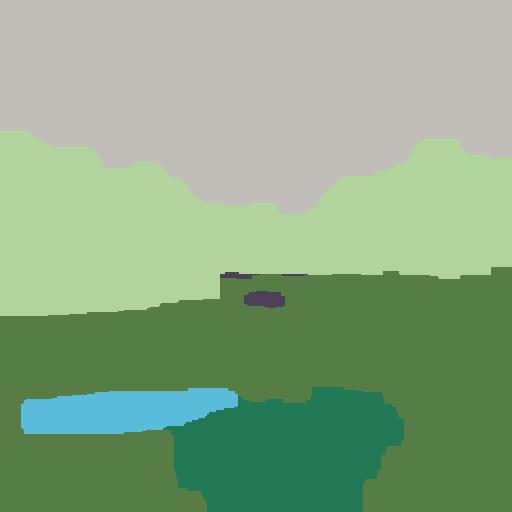} &
\figsB{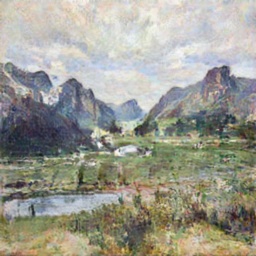} &
\figsB{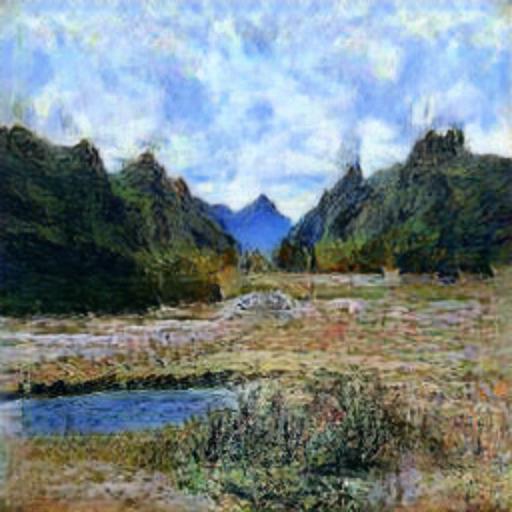} \\

{ \rotatebox{90}{~~~Van Gogh} } &
\figsB{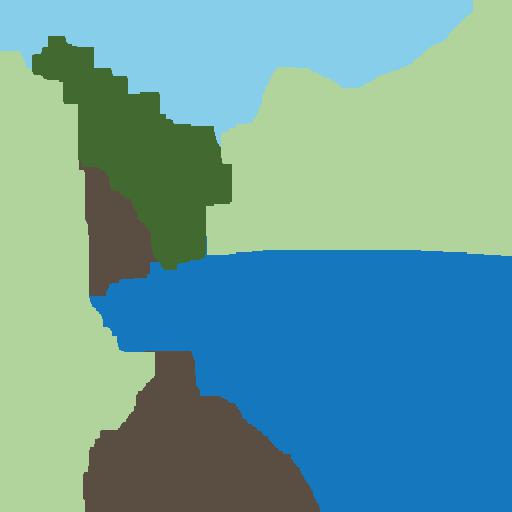} &
\figsB{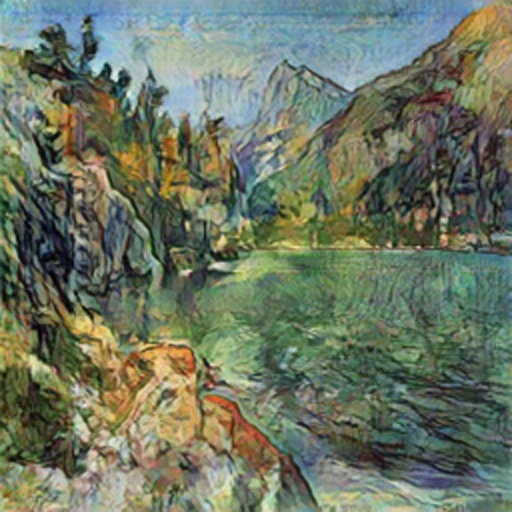} &
\figsB{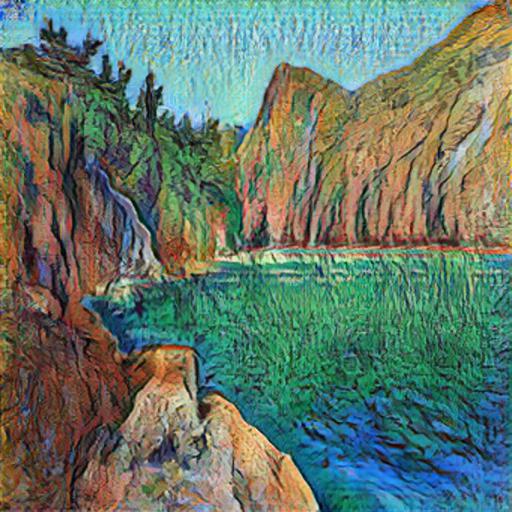} &
\figsB{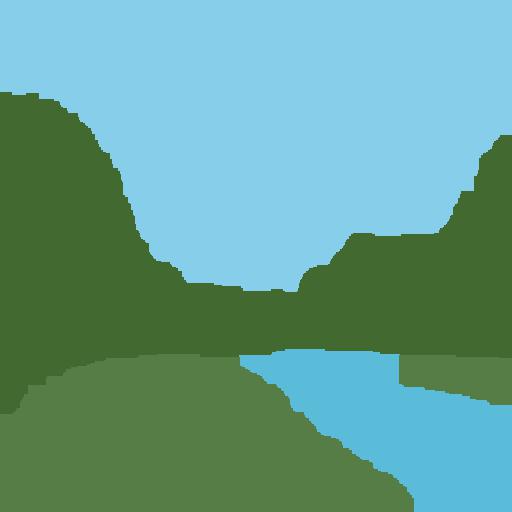} &
\figsB{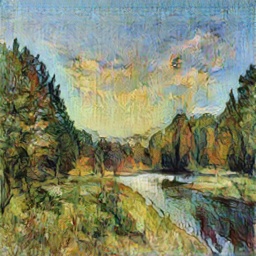} &
\figsB{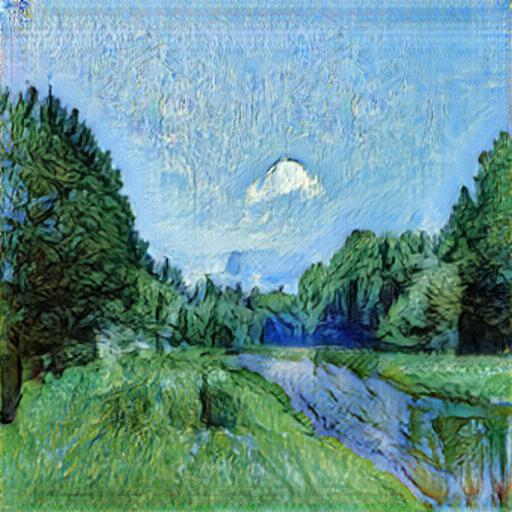} \\
{ \rotatebox{90}{~~Watercolor} } &
\figsB{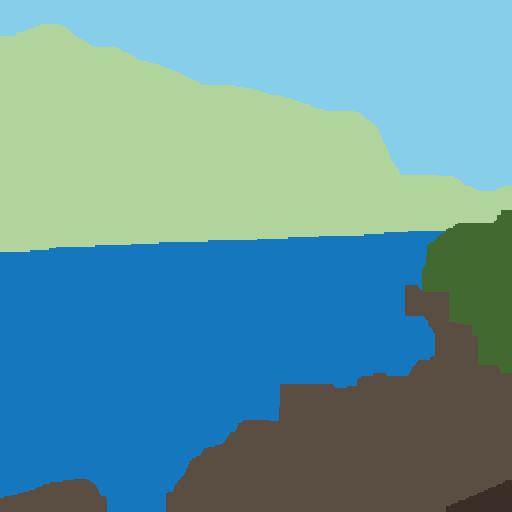} &
\figsB{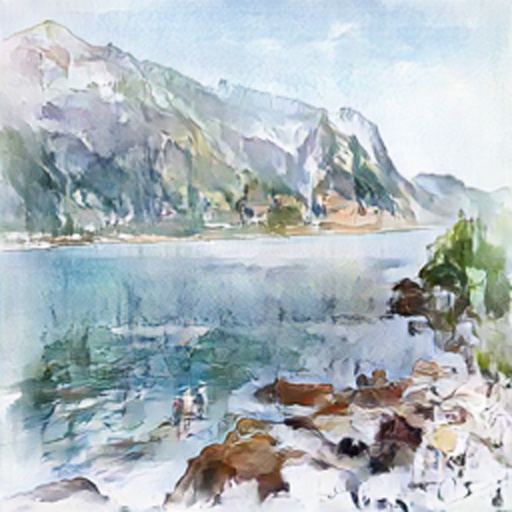} &
\figsB{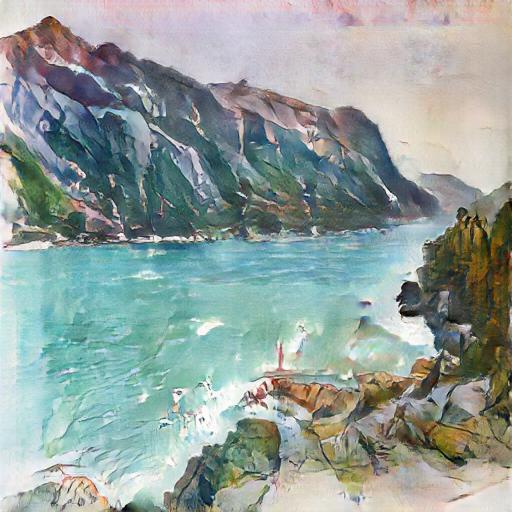}&
\figsB{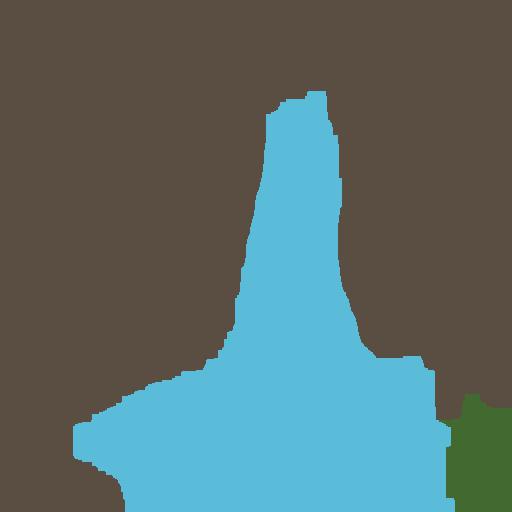} &
\figsB{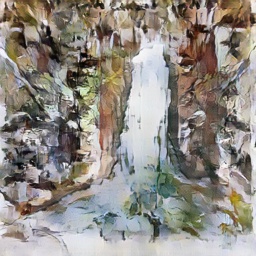} &
\figsB{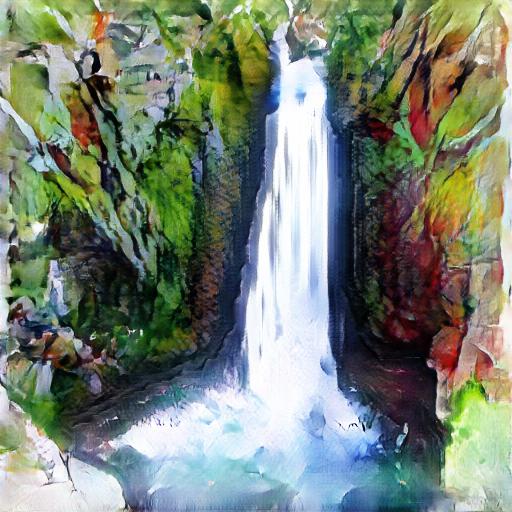} \\

\end{tabular}
\setlength{\tabcolsep}{6pt}
\end{center}
\caption{\textbf{Qualitative comparison for domain-specific
generations.} We compare our approach against existing approaches for artwork
generation in each of the specific domains.}
\label{fig:final_results_1}
\end{figure*}

\subsection{Implementation Details}
We trained all the networks from scratch using our dataset. 
The image size was set to $512\times 512$ pixels in our proposed model
and the Co-Mod GAN~\cite{COMOD}, and $256\times 256$ in other
baseline models owing to GPU memory limitations. For further details, please refer to
the supplementary material. A learning rate of 0.0002 was used for the first
40 epochs before decaying linearly to zero over the following 20 epochs. 
In the inference phase, the SEAN model required the style code as input, which we
pre-computed from the mean style codes in each domain for all the training data.
For our model, we set the input latent code $z$ to a 0 vector. 
and $\alpha = 3$, leading to a positive style appearance
for a particular domain in the quantitative comparison. This setting
results in the stable generation of synthesized artwork in each domain;
therefore, the proposed framework requires only label maps as input.

\newcommand{\figsD}[1]{\includegraphics[width=0.13\linewidth]{results/compare_with_NST_reference/#1}}
\begin{figure*}[p]
\begin{center}
\setlength{\tabcolsep}{1pt}
\begin{tabular}{cccccccccc}
& Input & Reference & SPADE + & SPADE + & SPADE + & SEAN & CMSAS \\
&  & Image & NST & AdaIN & STROTSS &  & (Ours) \\
\raisebox{1mm}{ \rotatebox{90}{~~Ink-wash} } &
\figsD{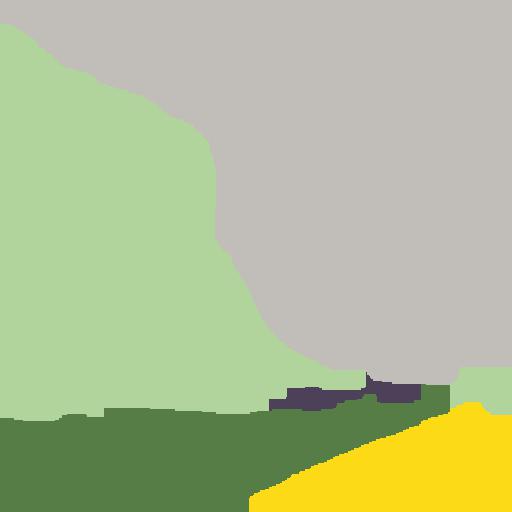} &
\figsD{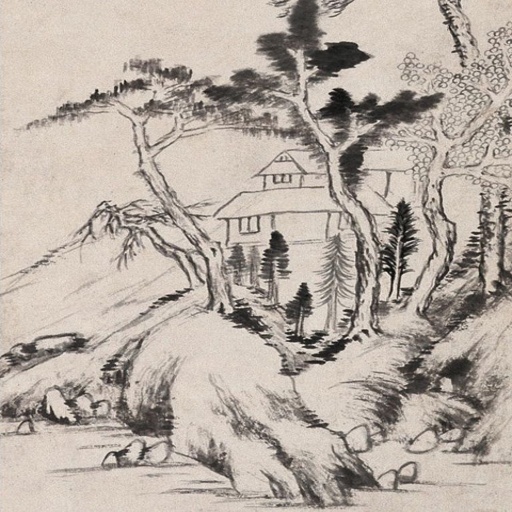} &
\figsD{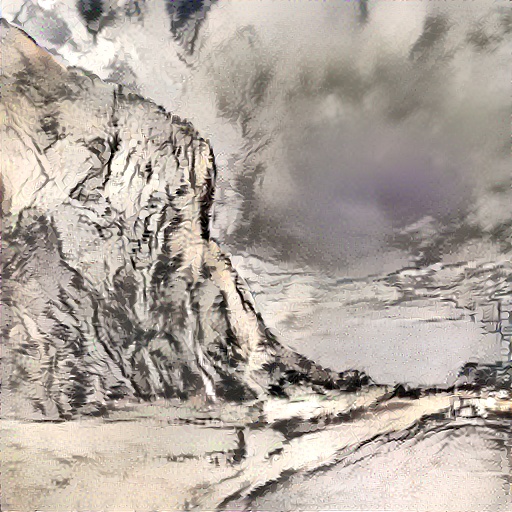} &
\figsD{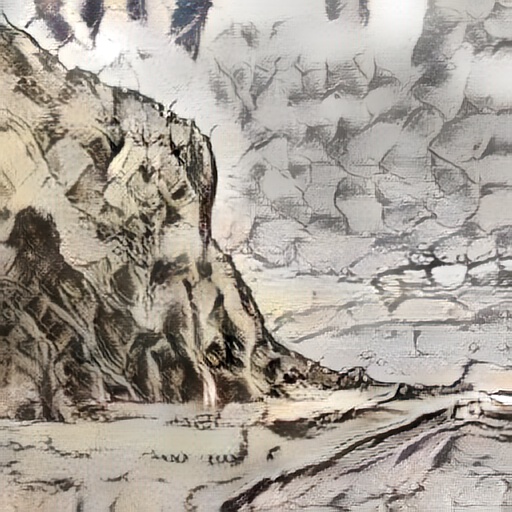} &
\figsD{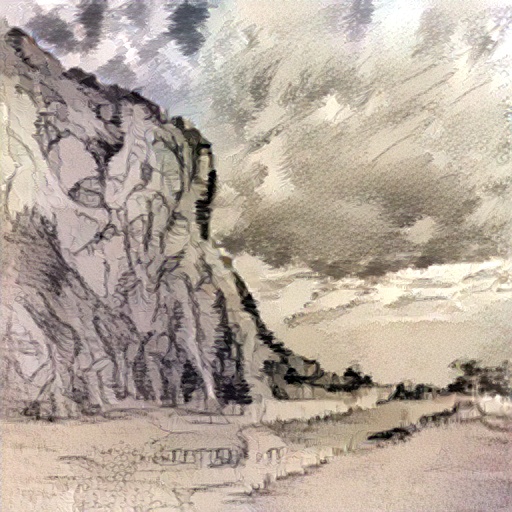} &
\figsD{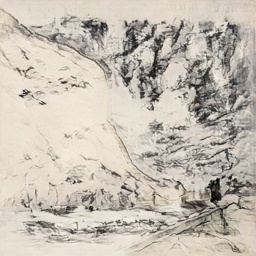} &
\figsD{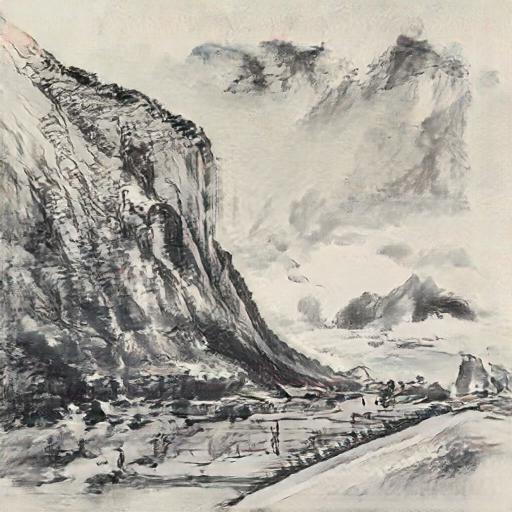} \\

\raisebox{1mm}{ \rotatebox{90}{~~~~~Monet} } &
\figsD{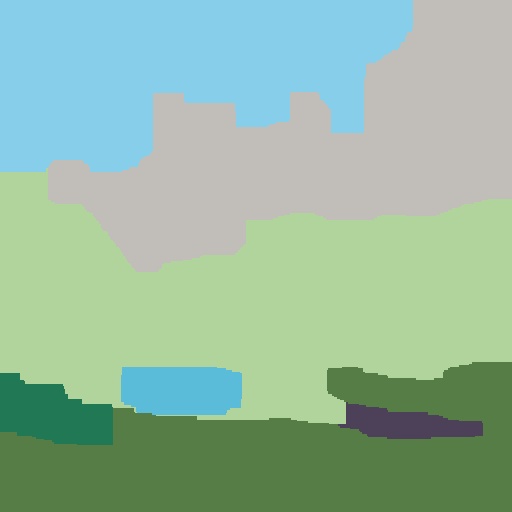} &
\figsD{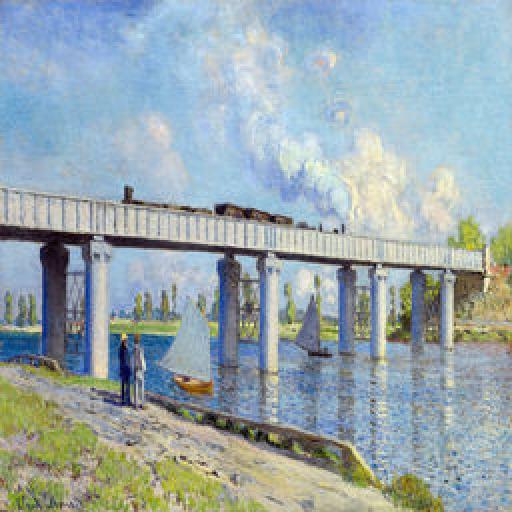} &
\figsD{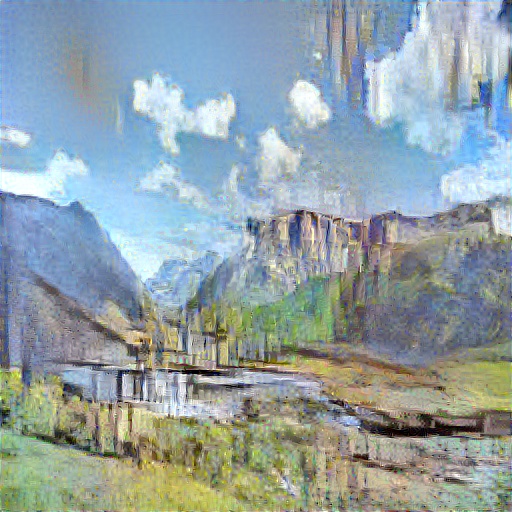} &
\figsD{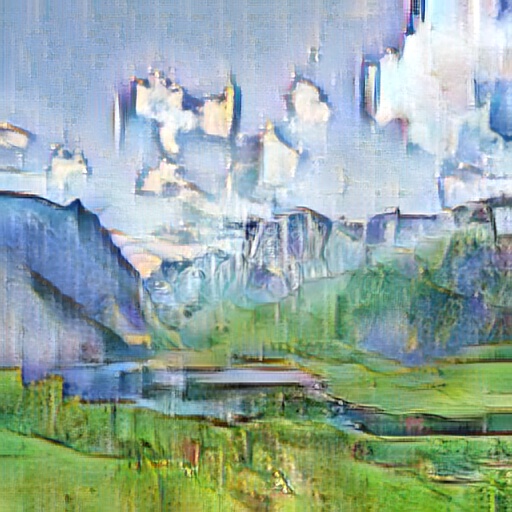} &
\figsD{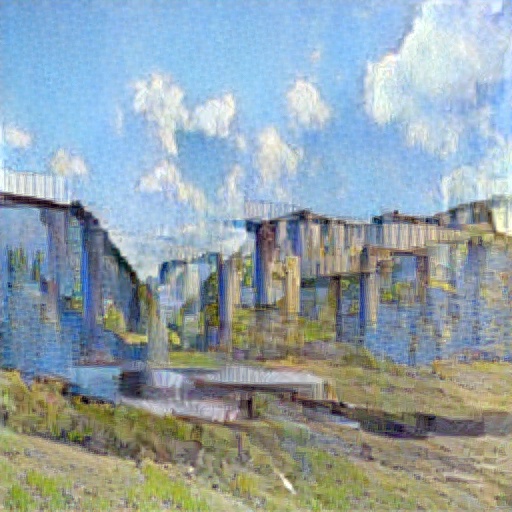} &
\figsD{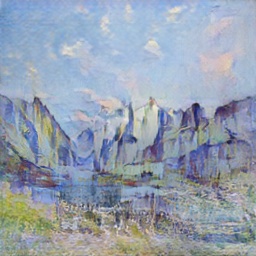} &
\figsD{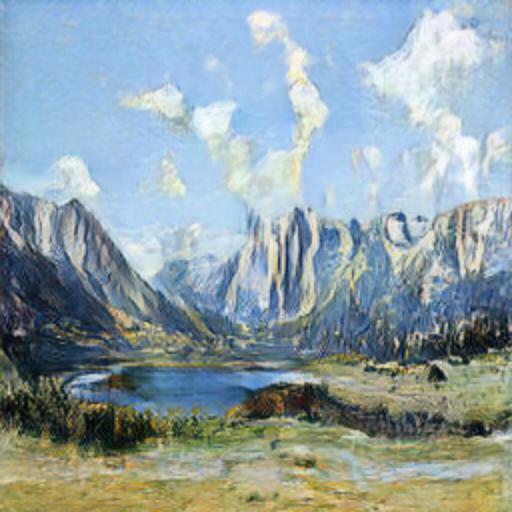}\\

\raisebox{1mm}{ \rotatebox{90}{~~Van Gogh} } &
\figsD{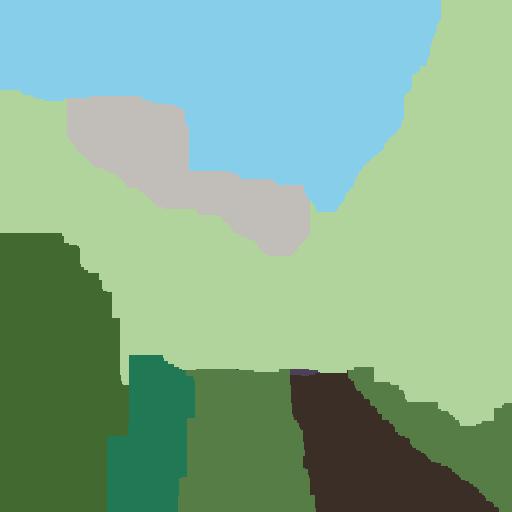} &
\figsD{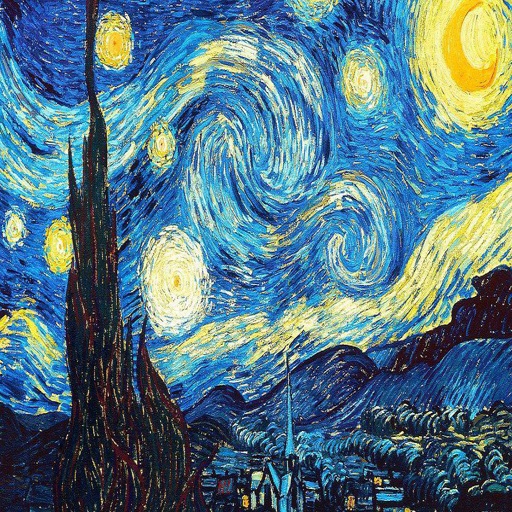} &
\figsD{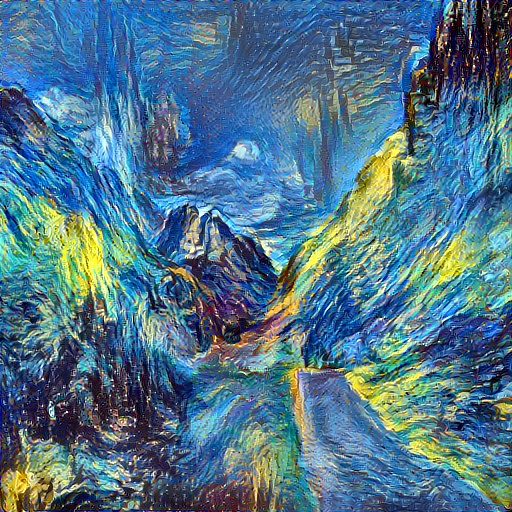} &
\figsD{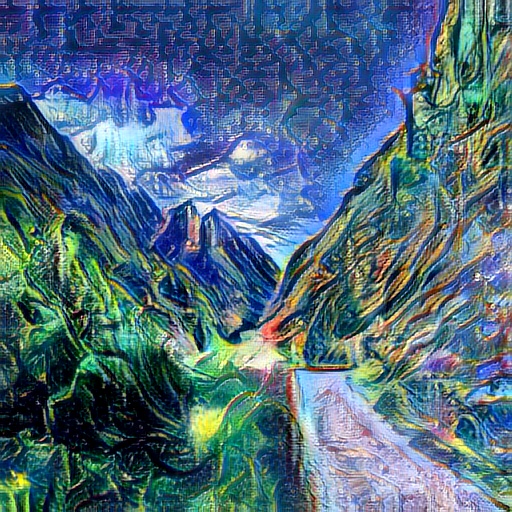} &
\figsD{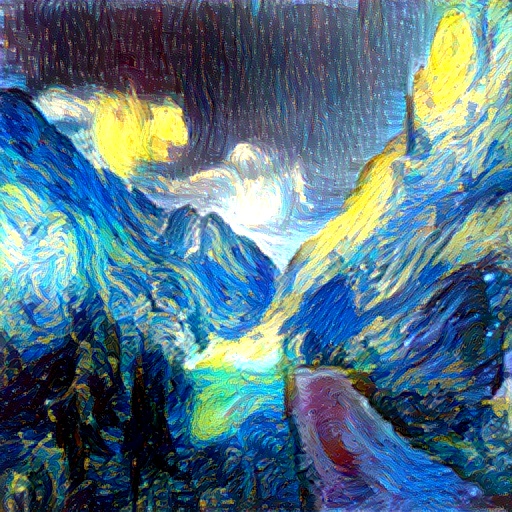} &
\figsD{26769471712_vangogh_SEAN_.jpg} &
\figsD{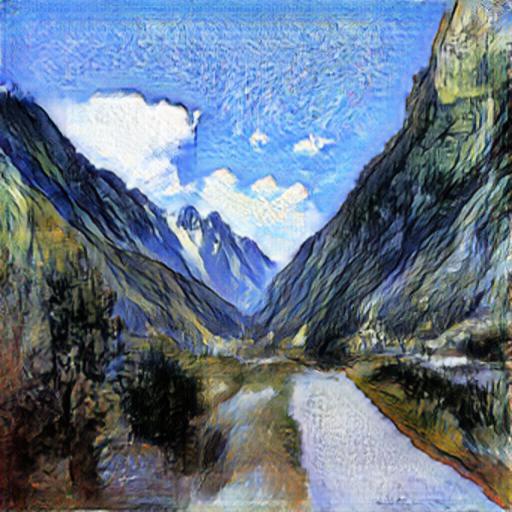}\\

\raisebox{1mm}{ \rotatebox{90}{~~Watercolor} } &
\figsD{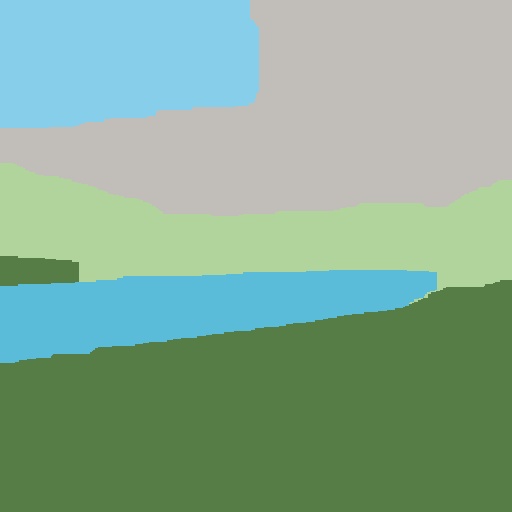} &
\figsD{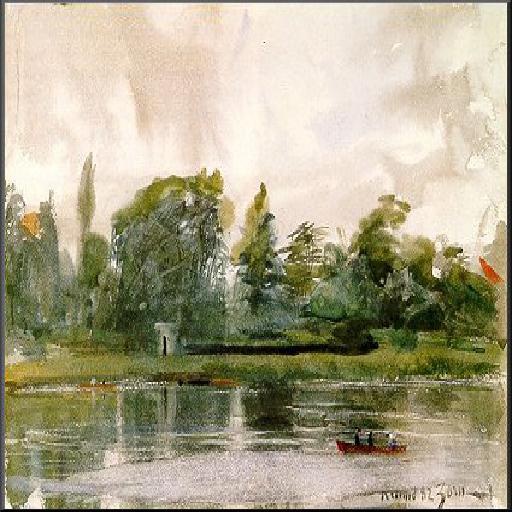} &
\figsD{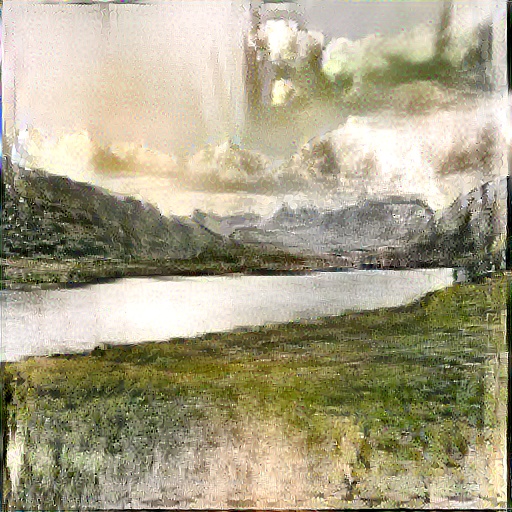} &
\figsD{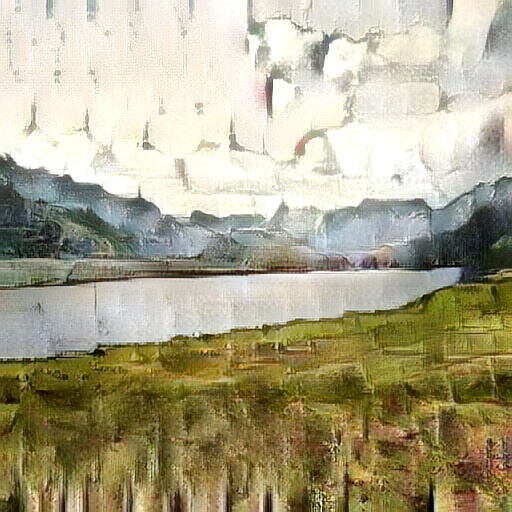} &
\figsD{8038931637_water_1_3.jpg} &
\figsD{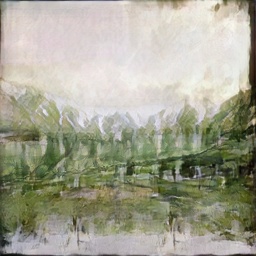} &
\figsD{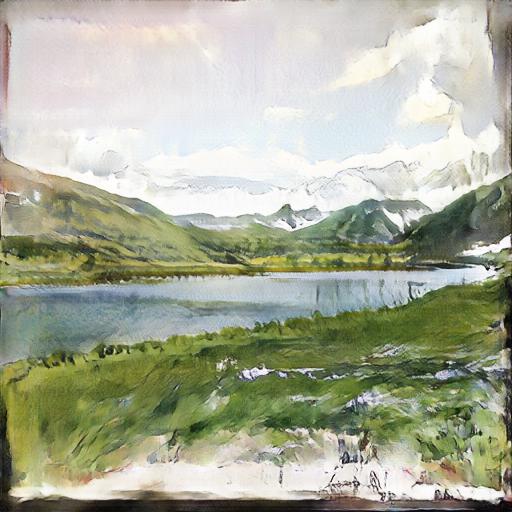} \\

\end{tabular}
\setlength{\tabcolsep}{6pt}
\caption{\textbf{Qualitative comparison of reference-based artwork generation
with neural style transfer models.} All reference images are licensed under the
public domain.}
\label{fig:final_results_with_ref}

\newcommand{\figsE}[1]{\includegraphics[width=0.145\linewidth]{results/additional_references/#1}}
\setlength{\tabcolsep}{1pt}
\begin{tabular}{ccccc}
 & Ink & Monet & Van Gogh & Watercolor \\

{~~~~Source Image} &
\figsE{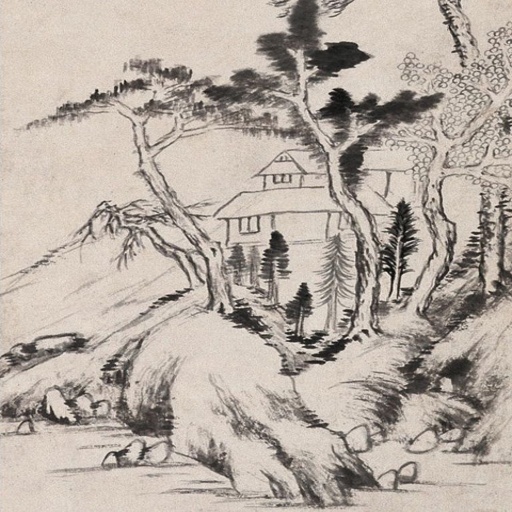} &
\figsE{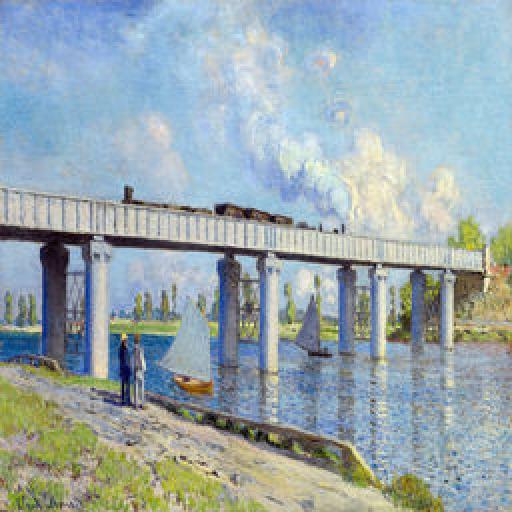} &
\figsE{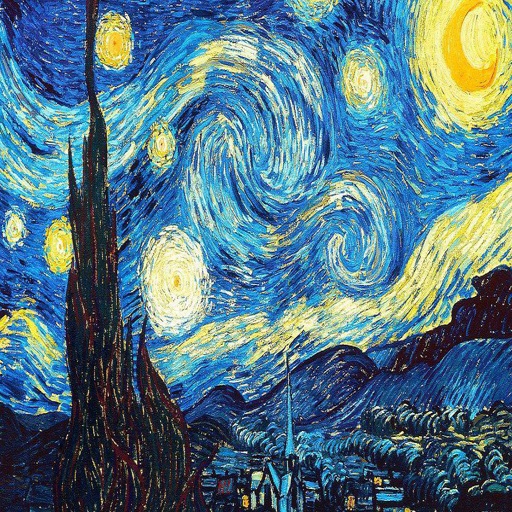} &
\figsE{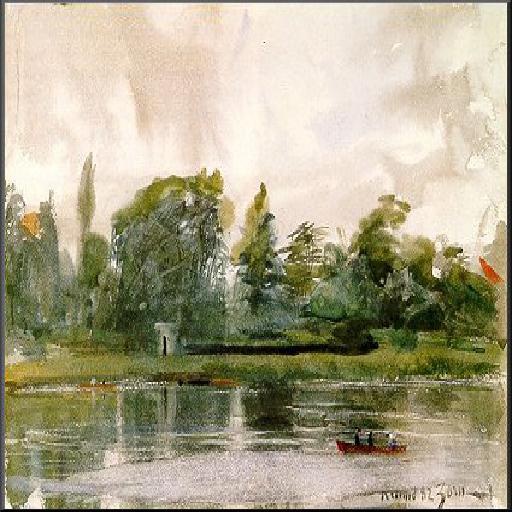}\\

\figsE{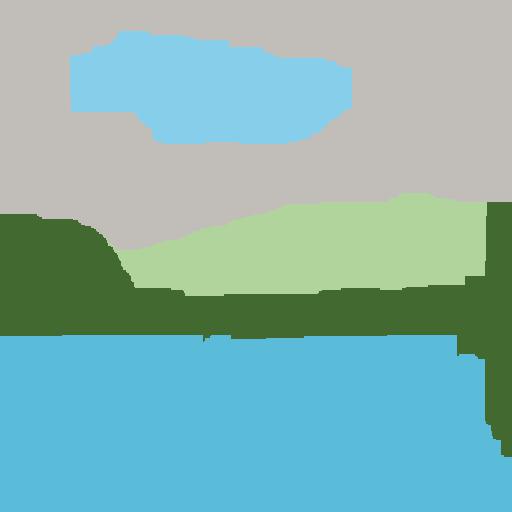} &
\figsE{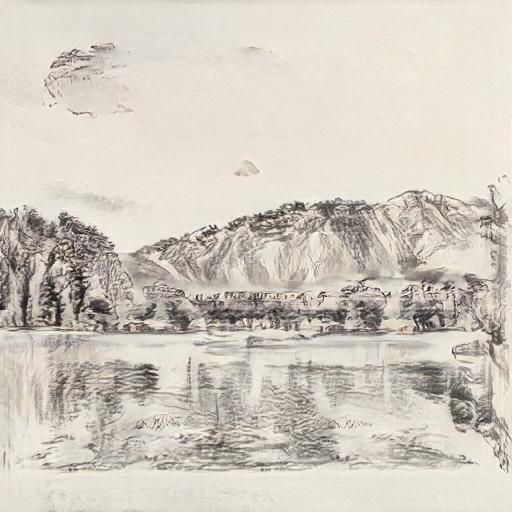} &
\figsE{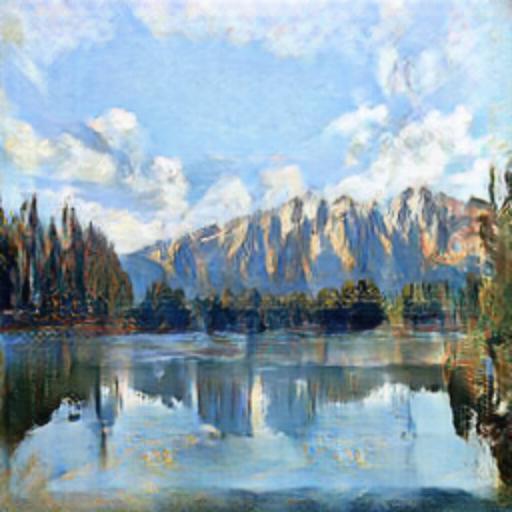} &
\figsE{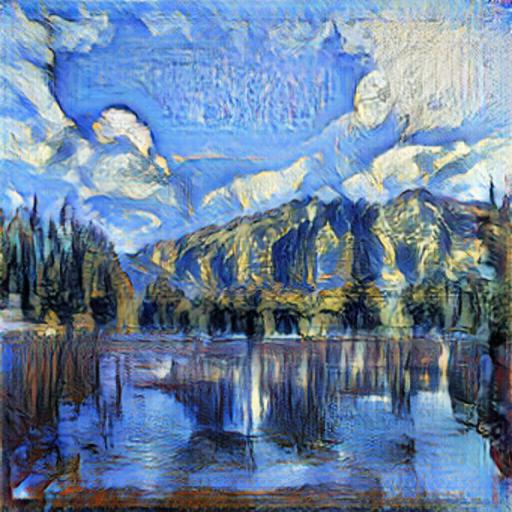} &
\figsE{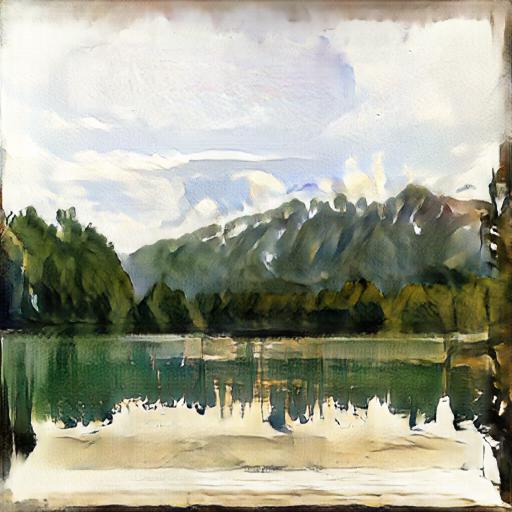}\\

\figsE{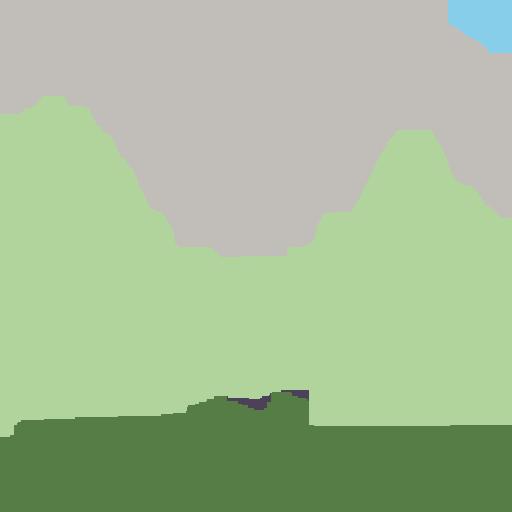} &
\figsE{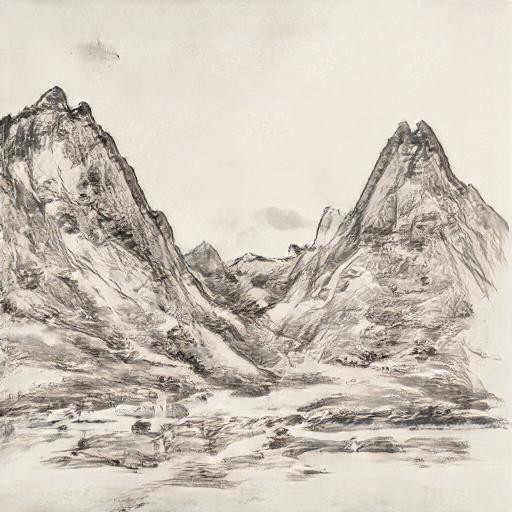} &
\figsE{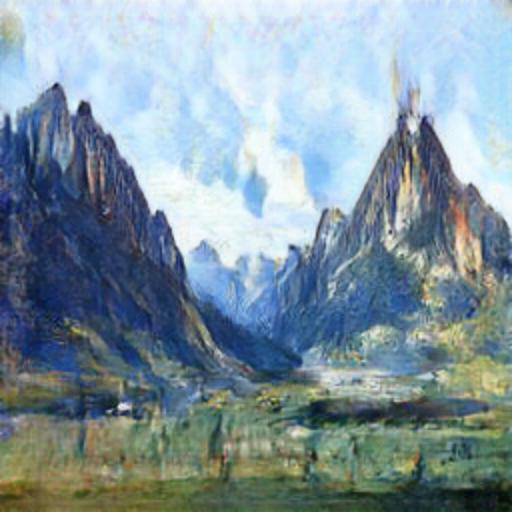} &
\figsE{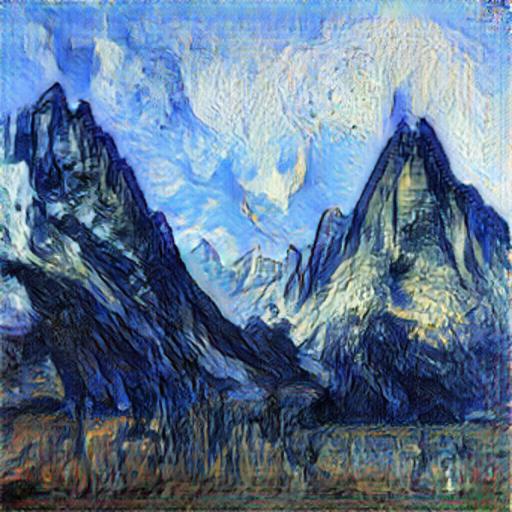} &
\figsE{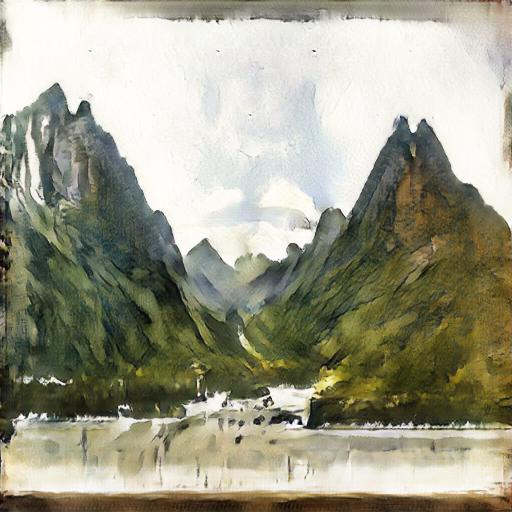}\\

\figsE{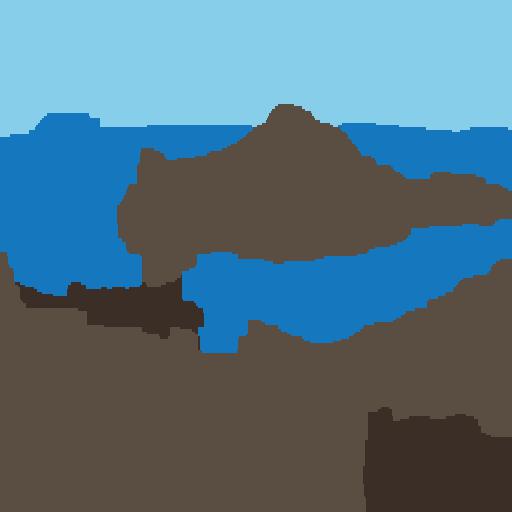} &
\figsE{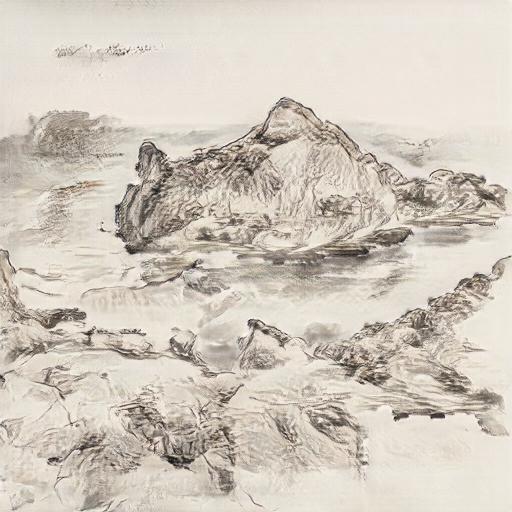} &
\figsE{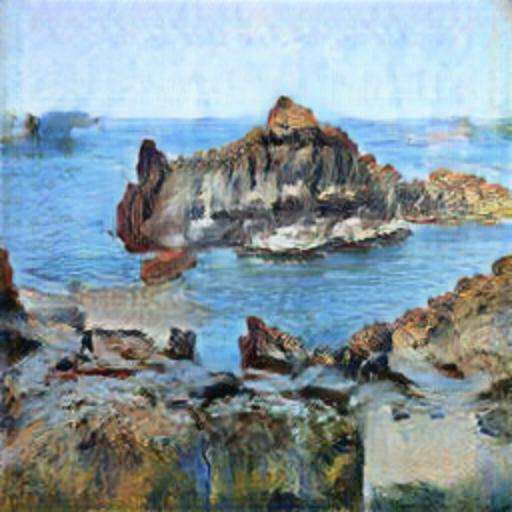} &
\figsE{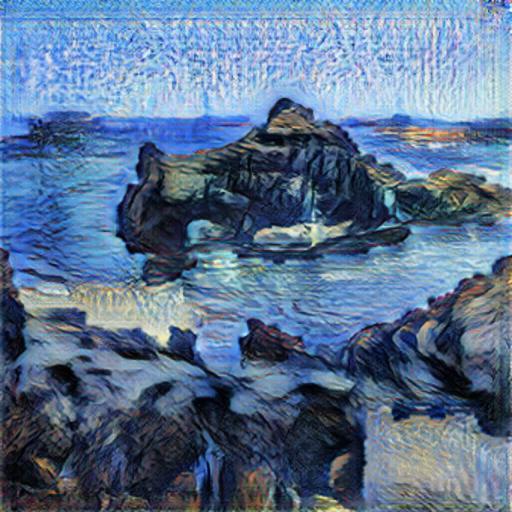} &
\figsE{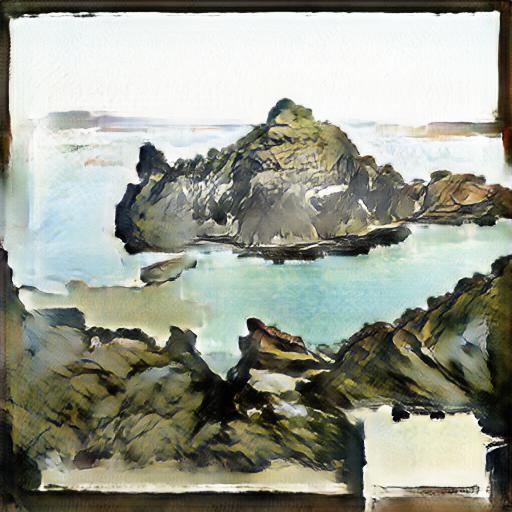}\\

\end{tabular}
\setlength{\tabcolsep}{6pt}
\end{center}
\vspace{-3mm}
\caption{\textbf{Reference-based generation of our proposed model in
multiple domains.} Our proposed method can apply styles similar to a
reference image while maintaining coherency with the input semantic
information.}
\label{fig:ours_reference}
\end{figure*}

\subsection{Quantitative Comparison}
We used the \textbf{Fréchet inception distance~(FID)}~\cite{heusel2017gans} as our
primary evaluation metric to capture the perceptual similarity of generated images
with real ones. After resizing all the images to the same size of $512\times 512$ pixels,
we calculated FID between the real and generated artwork for each
domain. As listed in Table~\ref{FID_results}, our
approach significantly outperforms existing approaches.
When input reference images are available, we compare our proposed
method with SEAN model, and three style transfer methods,
including NST~\cite{styletransfer}, AdaIN~\cite{AdaIN}, STROTSS~\cite{STROTSS}, as shown in Table~\ref{FID_results_with_ref}. Specifically, we used a simple
two-stage method to enable style transfer models for semantic image synthesis
by adding a pre-trained SPADE model~\cite{park2019gaugan} to generate synthesized images from
semantic layouts. We randomly selected 200 style images and 100 synthesized
images as content images for each domain and then calculated FID based on the 20,000
generated images in each domain.

\begin{table}[t]
\caption{\textbf{Perceptual user study results.}
The numbers indicate the percentage of users that prefer our method with respect to
existing approaches.}
\centering
\begin{tabular}{>{\centering\arraybackslash}p{1.2cm}	
>{\centering\arraybackslash}p{0.85cm}
>{\centering\arraybackslash}p{0.85cm}
>{\centering\arraybackslash}p{1.1cm}
>{\centering\arraybackslash}p{0.85cm}
>{\centering\arraybackslash}p{0.85cm}
}
 \toprule
   vs. & SMIS & OASIS & Co-Mod & SEAN & Real\\
  &  & & & \\
 \midrule
 CMSAS & \multirow{2}{*}{79.61\%} & \multirow{2}{*}{67.71\%} & \multirow{2}{*}{69.15\%} & \multirow{2}{*}{65.26\%} & \multirow{2}{*}{21.64\%} \\
 (Ours)\\
 \bottomrule
\end{tabular}
 \label{user_study}
\end{table}

\subsection{Perceptual User Study}
We evaluated our method through a perceptual user study with 15
participants. 
We randomly select 5,000 generated images per domain for each approach,
and all real artworks from our dataset.
In each round, two randomly selected images
from different approaches or real images were shown to the user.
We asked the participants to choose which image seemed better 
in terms of realism and style for 500 rounds per user.
As shown in Table~\ref{user_study}, our approach is preferred over
existing approaches. Furthermore, compared to real artwork, our
approach was considered better than 21.64\% of the time. This matches the
quantitative comparison of results.

To evaluate the layout preservation, we use the same 20,000 selected images
from each approach along with all the paired semantic layouts. We asked the same 15
participants to choose an absolute scale from 0 to 3 of how well they think
the generated artwork matches the label map.
In each round, a randomly selected image and its paired input layout were shown. The results in Table~\ref{user_study_label} 
indicate that our approach is preferred over the baseline models in terms
of layout preservation.

\begin{table}[t]
\caption{\textbf{Perceptual user study on layout preservation.} The numbers
indicate the average scores of how well the participants think the generated
artworks of each method preserve the semantic layout of the input. The scores
are on a scale of 0 to 3, with higher scores corresponding to better layout preservation. The best results are highlighted in bold.}
\begin{tabular}{rcccccc}
 \toprule
   & SMIS & OASIS & Co-ModGAN & SEAN & CMSAS\\
  &  & & &  &(Ours)\\
 \midrule
 Score & 2.32 & 2.61 & 2.51 & 2.55 & \textbf{2.68} \\
 \bottomrule
\end{tabular}
 \label{user_study_label}
\end{table}

\newcommand{\figsC}[1]{\includegraphics[width=0.107\linewidth]{results/SSTAN_ablation_study/#1}}
\begin{figure*}[t]
\begin{center}
\setlength{\tabcolsep}{1pt}
\begin{tabular}{ccccccc|cc}

Input & B & B+D & B+D+A & B+D+A+S1 & B+D+A+S3 & B+D+A+S6 & B+C & B+C+A \\

\figsC{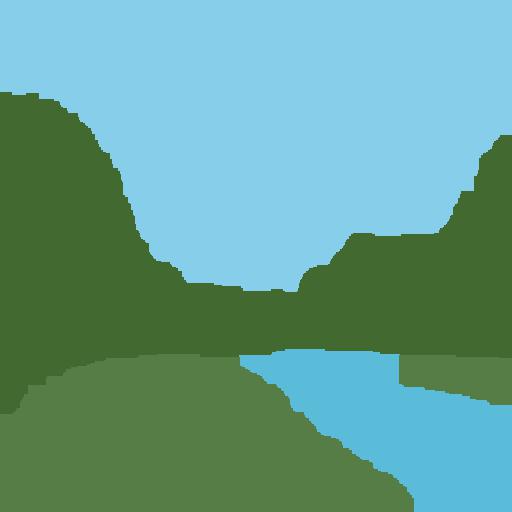} &
\figsC{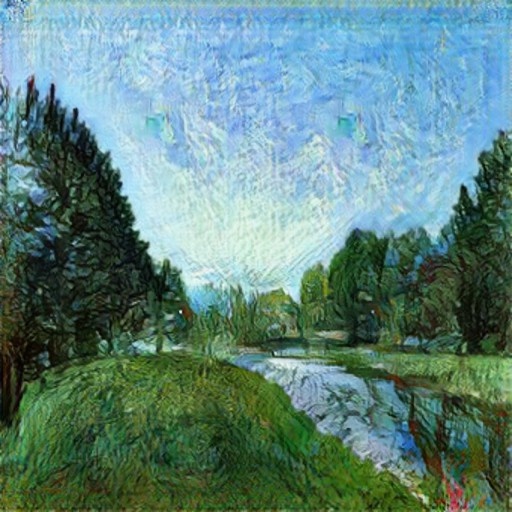} &
\figsC{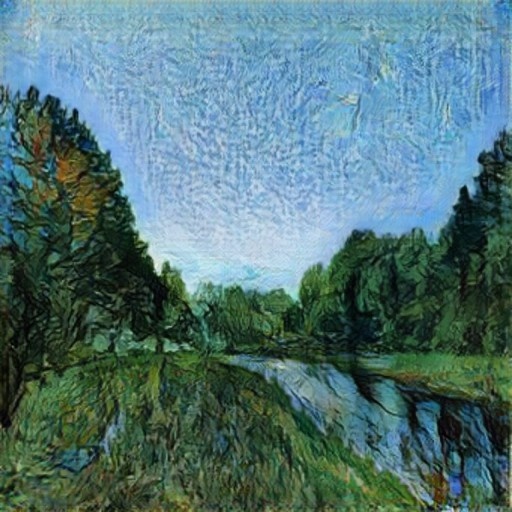} &
\figsC{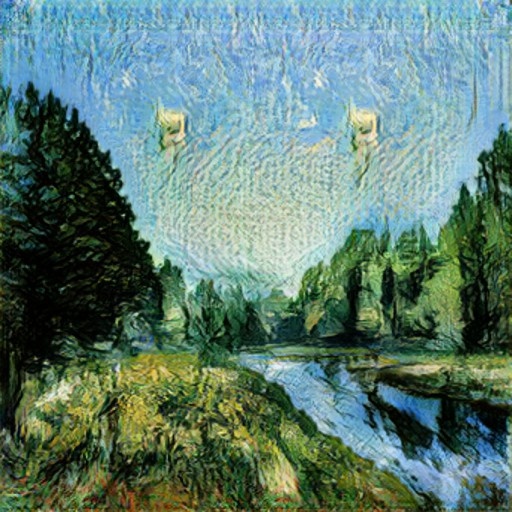} &
\figsC{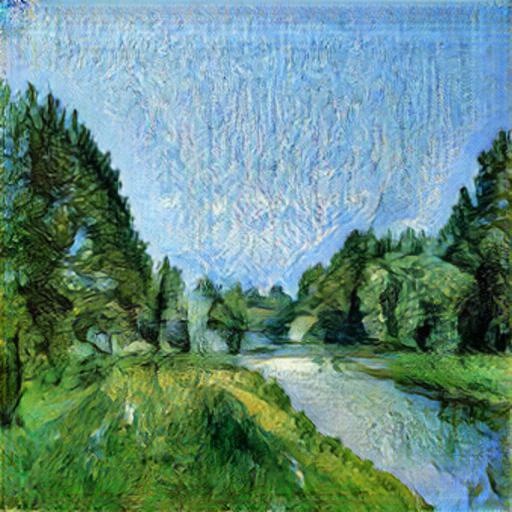} &
\figsC{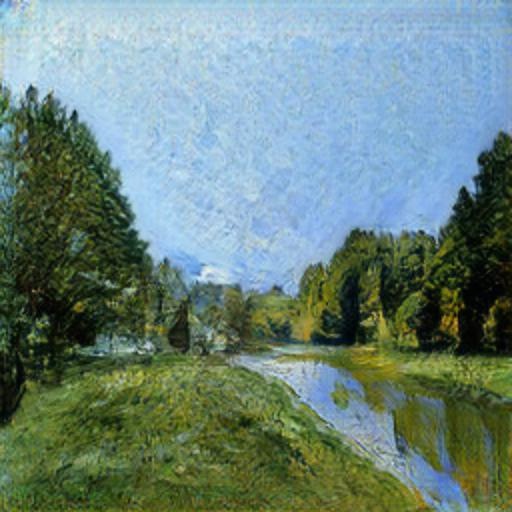} &
\figsC{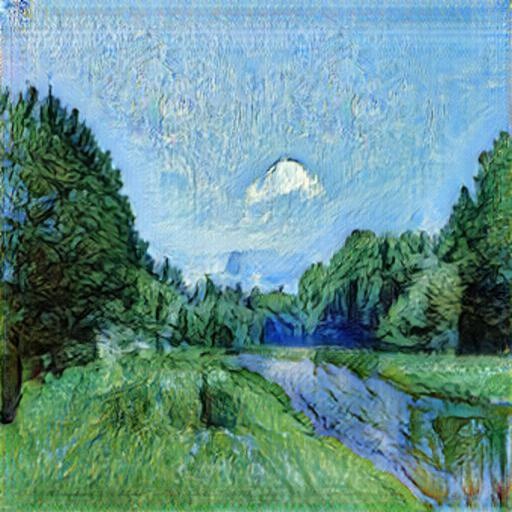} &
\figsC{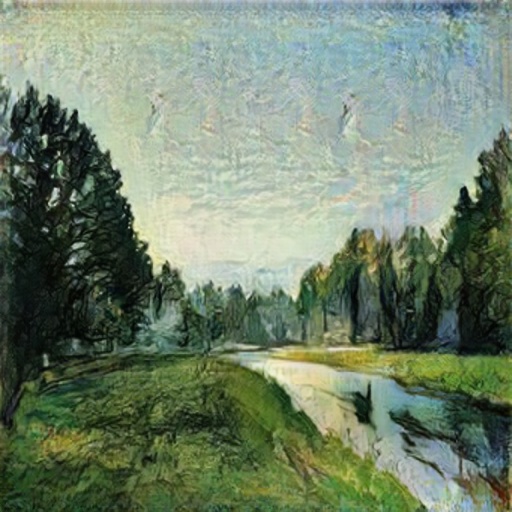} &
\figsC{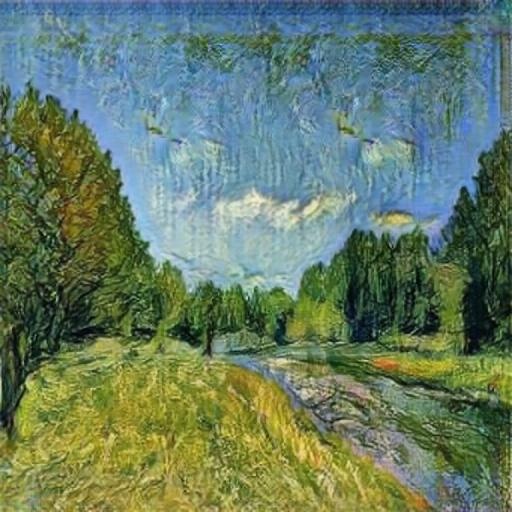} \\

\figsC{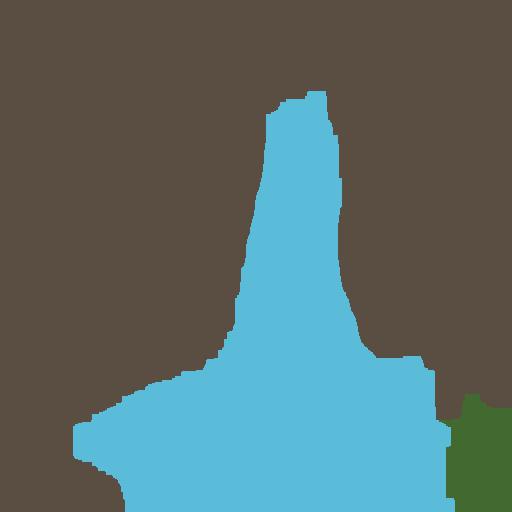} &
\figsC{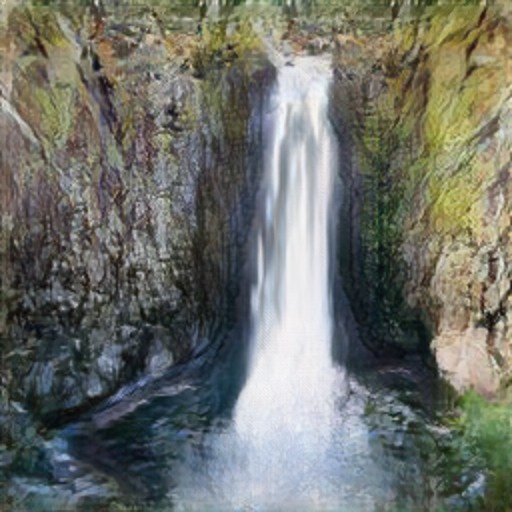} &
\figsC{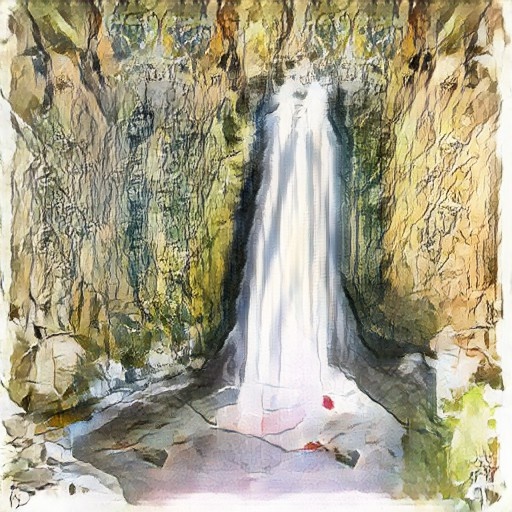} &
\figsC{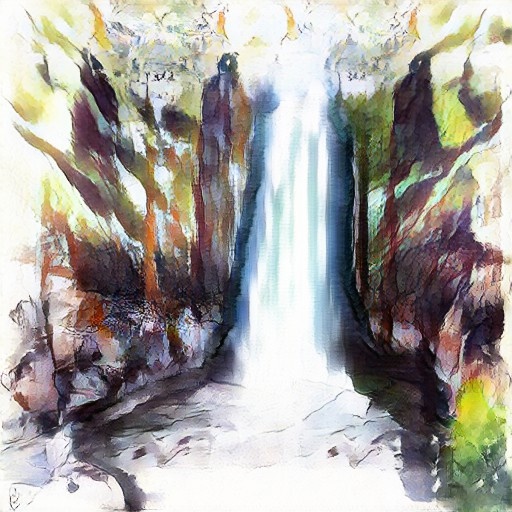} &
\figsC{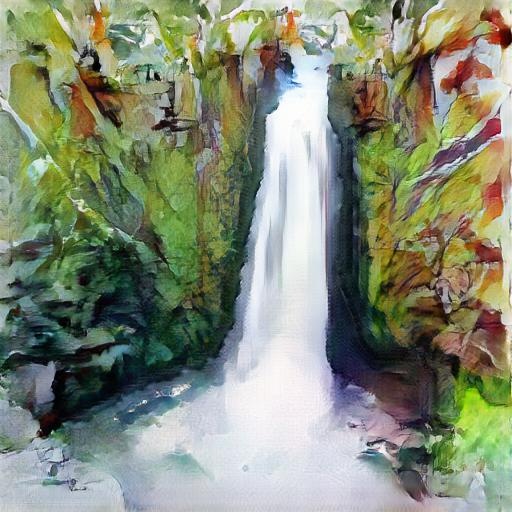} &
\figsC{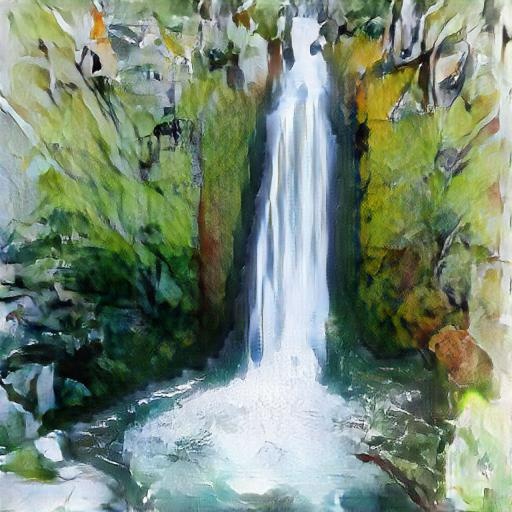} &
\figsC{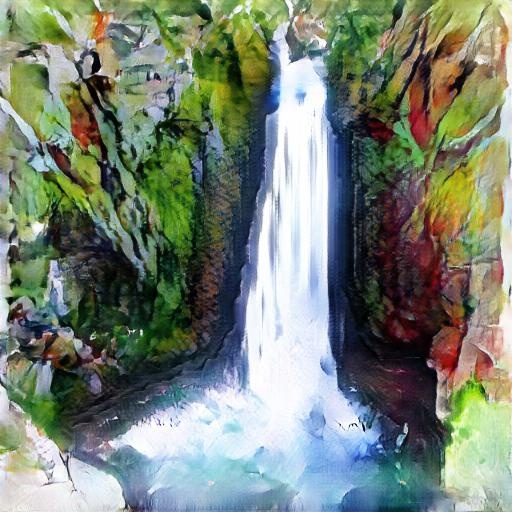} &
\figsC{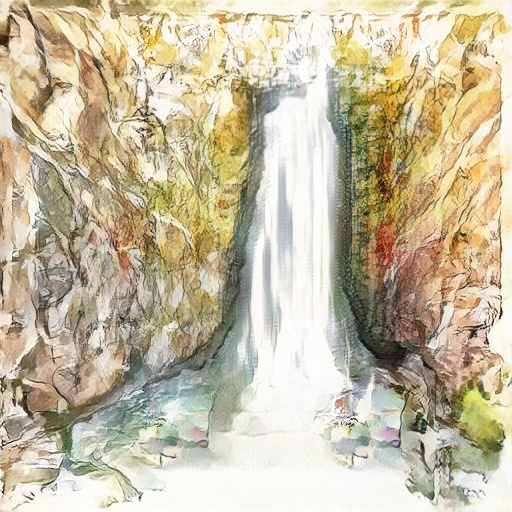} &
\figsC{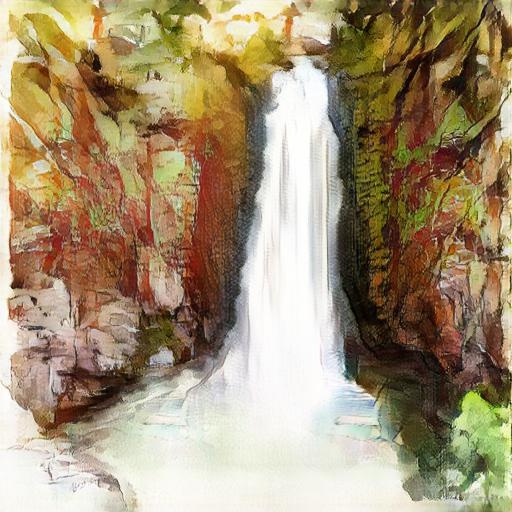} \\

\end{tabular}
\setlength{\tabcolsep}{6pt}
\end{center}
\caption{\textbf{Examples from ablation study.}
We show sets of results for different configurations given two different
semantic label maps and latent codes. We show results for the base model
(\emph{B}) with different components, including attention modules
(\emph{A}), separate domain encoders (\emph{D}), and SSTAN
for 1 layer (\emph{S1}), 3 layer (\emph{S3}), and 6 layer
(\emph{S6}) configurations, alongside a class-conditional approach
(\emph{C}). The top and bottom rows show the \emph{Van Gogh} and
\emph{Watercolor} domain, respectively.}
\label{fig:ablation}
\end{figure*}

\subsection{Visual Quality Comparison} 

Given that some of the baseline models cannot generate artwork in a specific domain,
we divide the visual quality comparison into two parts: 
joint multimodal generation in all domains and generation in a specific domain.

For the multimodal generation scenario, we provide a qualitative comparison of results in
Fig.~\ref{fig:multimodal} against the baseline approaches.
SMIS generates poorer results in terms of style,
whereas OASIS and Co-mod GAN models generate higher quality artwork
albeit with few details. The reflections on the water, texture, and style details on the mountains, as well as the
clouds of our proposed method, exhibit higher quality compared to baseline models.
We hypothesize that the performance increase comes from SSTAN enforcing the
influence of style information, leading to better style-aware generation.

We qualitatively compare domain-specific
generation against the SEAN model in Fig.~\ref{fig:final_results_1}. Compared to our approach, the results using SEAN tend to be fuzzier and have fewer details. Furthermore, the proposed CMSAS model allows for more control with the latent
codes of the style instead of a fixed pre-computed style code used in
the SEAN model.

We also provided a qualitative comparison when reference images were available. 
In CMSAS model, domain-specific encoders were used to convert the reference images into latent codes, 
which are then passed to a generator that synthesizes artwork with a similar appearance.
To compare with existing style transfer methods without semantic image synthesis,
we added a pre-trained SPADE model to generate synthesized images from the semantic layouts. 
The synthesized images are then passed on to style transfer models to apply the reference style.
As shown in Fig.~\ref{fig:final_results_with_ref}, 
the proposed method significantly outperformed the baseline models.
However, when the reference image is extremely stylistic, as in
the Starry Night example in Fig.~\ref{fig:final_results_with_ref}, unlike
baseline methods that try to extract style representations from the reference
image, the proposed method mimics a style similar to that of the learned
latent space, which may result in less stylism owing to the limitations of the
training dataset. The additional reference-based generation results of the proposed model are
shown in Fig.~\ref{fig:ours_reference}. Given the same fixed semantic layouts
as input, our proposed method can apply a similar style based on the
reference image.

\begin{table}[!t]
\caption{\textbf{Ablation study.} We evaluate the effect of different
components on the FID scores. Our base model (\emph{B}) is equivalent to a SPADE
model trained with a single image encoder. In
particular, we look at the effects of the attention module (\emph{A}), separate
domain encoders (\emph{D}), and SSTAN for 1
layer (\emph{S1}), 3 layer (\emph{S3}), and 6 layer (\emph{S6}) configurations.
We also compare with a class-conditional (\emph{C}) method in the last
two columns, in which we concatenate the latent code with a one-hot class label
that indicates a different domain. Thus, we generate artwork in different
domains via one-hot class labels in the last two methods, whereas other
approaches use latent space manipulation~\ref{DomainControl} to separate
domains for more controllability, such as cross-domain style
morphing or reference-based generation. The method~(B + D + A + S6)
corresponds to our full CMSAS model. The best results are highlighted in
bold.}

\centering
\begin{tabular}{lp{1cm}cp{1cm}cp{1cm}cp{1cm}cp{1cm}}
 \toprule
 Methods &  Ink-wash & Monet & Van Gogh & Watercolor\\
 \midrule
 B & 78.83 & 74.31 & 140.23 & 95.43\\
 B + D & 77.63 & 72.14 & 134.96 & 89.85\\
 B + D + A & 61.25 & 69.34 & 123.21 & 76.15\\
 B + D + A + S1 & 56.36 & 68.51 & 121.13 & 73.01 \\
 B + D + A + S3 & 51.17 & 67.32 & 119.05 & 72.24 \\
 B + D + A + S6 & \textbf{49.33} & \textbf{65.97} & \textbf{115.3} & \textbf{68.96} \\
 \midrule
 B + C & 71.74 & 72.97 & 134.14 & 84.45\\
 B + C + A & 60.97 & 68.71 & 124.11 & 75.52\\
 \bottomrule
\end{tabular}
\label{Ablation_Study}
\end{table}

\subsection{Ablation Study}
An ablation study was conducted to verify the effectiveness of each element.
The results presented in Table~\ref{Ablation_Study} and Fig.~\ref{fig:ablation} show that
performance is significantly improved by adding each element to the complete CMSAS
model.
In addition, we provide a comparison with the traditional
class-conditional approach that concatenates latent code with a one-hot
class labels indicating different domains. The results indicate that the
proposed method outperformed class-conditional approaches. In
addition, our proposed model can perform across-domain style morphing and
reference-based generation, leading to higher controllability.

\subsection{Limitations and Discussion}

\newcommand{\limifig}[1]{\includegraphics[width=0.48\linewidth]{figures/limitation/#1}}
\begin{figure}[t]
    \centering
    \setlength{\tabcolsep}{1pt}
    \begin{tabular}{cc}
    \limifig{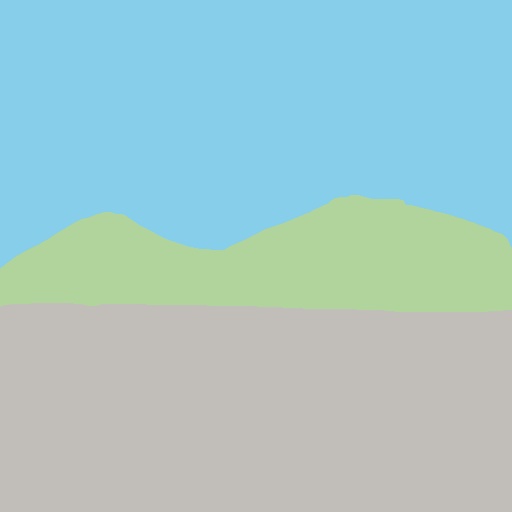} & \limifig{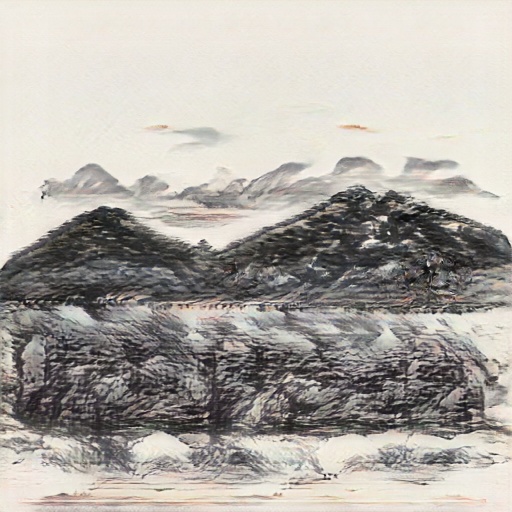} \\
    \limifig{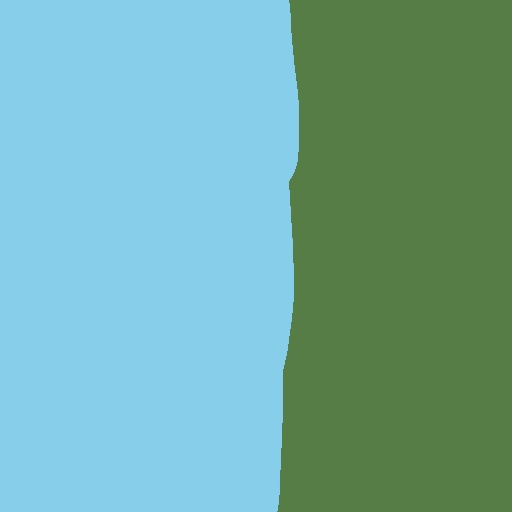} & \limifig{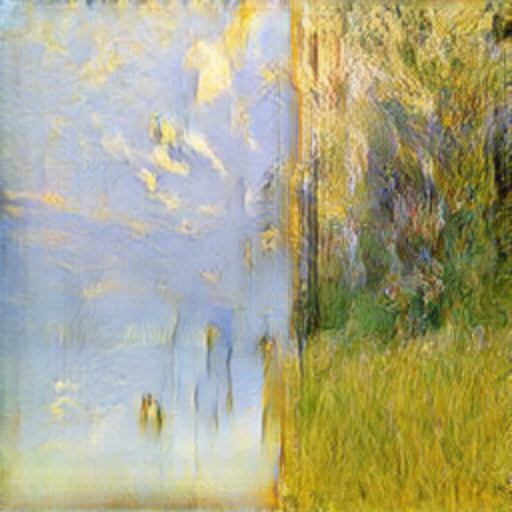} \\
    Input & Output \\
    \end{tabular}
    \setlength{\tabcolsep}{6pt}
    \caption{\textbf{Generation conditioned on unnatural inputs.} When the
    inputs are unnatural, such as clouds under mountains or 90°
   rotations, the model may struggle to generate results as the input
    differs significantly from the training images.}

    \label{limitation1}
\end{figure}

\begin{figure}[t]
    \centering
    \setlength{\tabcolsep}{1pt}
    \begin{tabular}{cc}
    \limifig{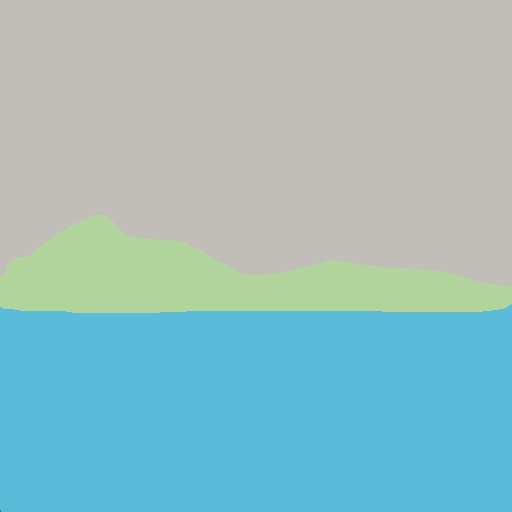} & \limifig{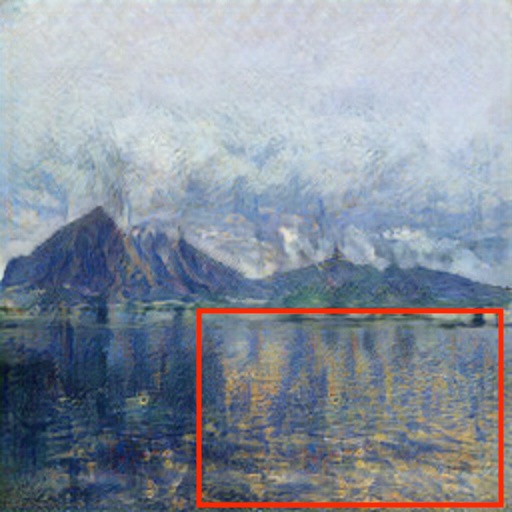} \\
    \limifig{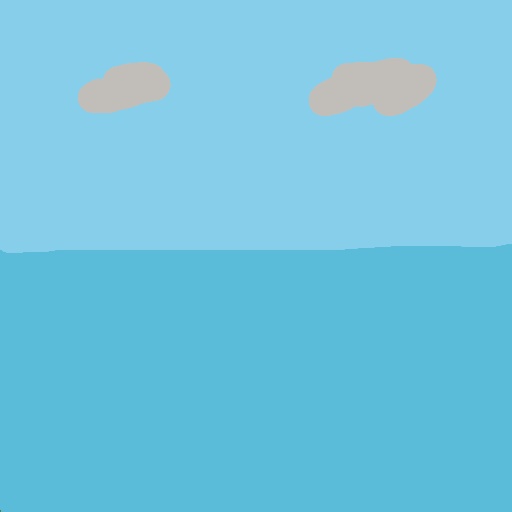} & \limifig{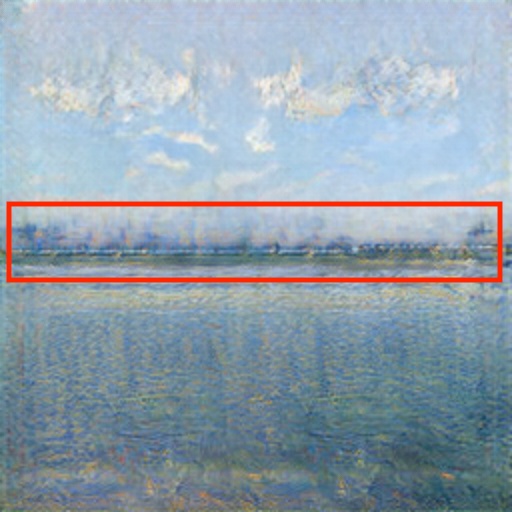} \\
    Input & Output \\
    \end{tabular}
    \setlength{\tabcolsep}{6pt}
    \caption{\textbf{Dataset bias.} When the input semantic maps do not share
    characteristics with the training data, the model may fail to accurately
    convey the intended scene. In the top row, although
    the user specified a fully cloudy sky, there is a reflection of the sun in the water.
In the bottom row, despite the absence of land in the input semantic map, the model attempts to generate a thin piece of land on the
    horizon.}

    \label{limitation2}
\end{figure}

Although our framework can generate high-quality artwork using
semantic label maps, it has several limitations owing to its data-driven approach.
In particular, its application is limited to known labels and new data must be
acquired to extend the model to new labels, such as animals. Furthermore, the
model learns a mapping from realistic semantic maps to artwork images and can
fail if the input semantic maps diverge significantly from the training data, as
shown in Fig.~\ref{limitation1}.
Moreover, the style or tone of the generated artwork was restricted to those
similar to real artwork used for the training dataset. In certain scenes, the
input semantic can also produce unexpected results owing to dataset bias.
Examples are presented in Figs.~\ref{limitation2}.

\section{Conclusion}
We present a novel paired dataset of semantic maps, landscape images, and
artwork from different domains, along with a high-performance method for
generating artwork from semantic maps. Our approach uses 
multi-domain artworks, exploiting the structure of the latent space to
precisely manipulate the resulting artwork. Furthermore,
because our approach is generated from semantic maps, it can be used in
interactive scenarios. We plan to release our dataset and hope that it will
encourage other researchers to further investigate artwork generation tasks.

\subsection*{Acknowledgements}
This study was supported by the Japan Science and Technology Agency Support for Pioneering Research Initiated by the Next Generation~(JST SPRING); Grant Number JPMJSP2124.

\subsection*{Declaration of competing interest}
The authors have no competing interests to declare.

\subsection*{Electronic Supplementary Material}
The supplementary material is available in the online version of this article.

\bibliographystyle{CVMbib}
\bibliography{refs}

\begin{thebibliography}{10}
\expandafter\ifx\csname urlstyle\endcsname\relax
  \providecommand{\doi}[1]{doi:\discretionary{}{}{}#1}\else
  \providecommand{\doi}{doi:\discretionary{}{}{}\begingroup
  \urlstyle{rm}\Url}\fi

\bibitem{hertzmann2018_art}
Hertzmann A. Can Computers Create Art? \emph{Arts}, 2018, 7(2): 18,
  \doi{10.3390/arts7020018}.

\bibitem{styletransfer}
{Gatys} LA, {Ecker} AS, {Bethge} M. Image Style Transfer Using Convolutional
  Neural Networks. In \emph{2016 IEEE Conference on Computer Vision and Pattern
  Recognition (CVPR)}, 2016, 2414--2423.

\bibitem{artgan}
Tan WR, Chan CS, Aguirre HE, Tanaka K. ArtGAN: Artwork synthesis with
  conditional categorical GANs. In \emph{2017 IEEE International Conference on
  Image Processing (ICIP)}, 2017, 3760--3764.

\bibitem{elgammal2017can}
Elgammal A, Liu B, Elhoseiny M, Mazzone M. Can: Creative adversarial networks,
  generating" art" by learning about styles and deviating from style norms.
  \emph{arXiv preprint arXiv:1706.07068}, 2017.

\bibitem{zhu2017unpaired}
Zhu JY, Park T, Isola P, Efros AA. Unpaired image-to-image translation using
  cycle-consistent adversarial networks. In \emph{Proceedings of the IEEE
  international conference on computer vision}, 2017, 2223--2232.

\bibitem{pix2pix}
Isola P, Zhu JY, Zhou T, Efros AA. Image-to-image translation with conditional
  adversarial networks. In \emph{Proceedings of the IEEE conference on computer
  vision and pattern recognition}, 2017, 1125--1134.

\bibitem{liu2020sketch}
Liu B, Song K, Zhu Y, Elgammal A. Sketch-to-art: Synthesizing stylized art
  images from sketches. In \emph{Proceedings of the Asian Conference on
  Computer Vision}, 2020, 207--222.

\bibitem{men2018common}
Men Y, Lian Z, Tang Y, Xiao J. A common framework for interactive texture
  transfer. In \emph{Proceedings of the IEEE Conference on Computer Vision and
  Pattern Recognition}, 2018, 6353--6362.

\bibitem{champandard2016semantic}
Champandard AJ. Semantic style transfer and turning two-bit doodles into fine
  artworks. \emph{arXiv preprint arXiv:1603.01768}, 2016.

\bibitem{park2019gaugan}
Park T, Liu MY, Wang TC, Zhu JY. Semantic image synthesis with
  spatially-adaptive normalization. In \emph{Proceedings of the IEEE Conference
  on Computer Vision and Pattern Recognition}, 2019, 2337--2346.

\bibitem{COMOD}
Zhao S, Cui J, Sheng Y, Dong Y, Liang X, Chang EI, Xu Y. Large scale image
  completion via co-modulated generative adversarial networks. \emph{arXiv
  preprint arXiv:2103.10428}, 2021.

\bibitem{OASIS}
Sushko V, Sch{\"o}nfeld E, Zhang D, Gall J, Schiele B, Khoreva A. You only need
  adversarial supervision for semantic image synthesis. \emph{arXiv preprint
  arXiv:2012.04781}, 2020.

\bibitem{zhu2020SMIS}
Zhu Z, Xu Z, You A, Bai X. Semantically Multi-modal Image Synthesis. In
  \emph{Proceedings of the IEEE/CVF Conference on Computer Vision and Pattern
  Recognition}, 2020, 5467--5476.

\bibitem{zhu2020sean}
Zhu P, Abdal R, Qin Y, Wonka P. SEAN: Image Synthesis with Semantic
  Region-Adaptive Normalization. In \emph{Proceedings of the IEEE/CVF
  Conference on Computer Vision and Pattern Recognition}, 2020, 5104--5113.

\bibitem{image_analogies}
Hertzmann A, Jacobs CE, Oliver N, Curless B, Salesin DH. Image analogies. In
  \emph{Proceedings of the 28th annual conference on Computer graphics and
  interactive techniques}, 2001, 327--340.

\bibitem{dehlinger2007fine}
Dehlinger H. On fine art and generative line drawings. \emph{Journal of
  Mathematics and the Arts}, 2007, 1(2): 97--111.

\bibitem{phon2012controlling}
Phon-Amnuaisuk S, Panjapornpon J. Controlling Generative Processes of
  Generative Art Somnuk Phon. \emph{Procedia Computer Science}, 2012, 13:
  43--52.

\bibitem{goodfellow2014}
Goodfellow IJ, Pouget-Abadie J, Mirza M, Xu B, Warde-Farley D, Ozair S,
  Courville A, Bengio Y. Generative Adversarial Networks, 2014.

\bibitem{artgan_improved}
{Tan} WR, {Chan} CS, {Aguirre} HE, {Tanaka} K. Improved ArtGAN for Conditional
  Synthesis of Natural Image and Artwork. \emph{IEEE Transactions on Image
  Processing}, 2019, 28(1): 394--409.

\bibitem{xue2021end}
Xue A. End-to-end chinese landscape painting creation using generative
  adversarial networks. In \emph{Proceedings of the IEEE/CVF Winter Conference
  on Applications of Computer Vision}, 2021, 3863--3871.

\bibitem{dobler2022art}
Dobler K, H{\"u}bscher F, Westphal J, Sierra-M{\'u}nera A, de~Melo G, Krestel
  R. Art Creation with Multi-Conditional StyleGANs. \emph{arXiv preprint
  arXiv:2202.11777}, 2022.

\bibitem{ramesh2022hierarchical}
Ramesh A, Dhariwal P, Nichol A, Chu C, Chen M. Hierarchical text-conditional
  image generation with clip latents. \emph{arXiv preprint arXiv:2204.06125},
  2022.

\bibitem{rombach2022high}
Rombach R, Blattmann A, Lorenz D, Esser P, Ommer B. High-resolution image
  synthesis with latent diffusion models. In \emph{Proceedings of the IEEE/CVF
  Conference on Computer Vision and Pattern Recognition}, 2022, 10684--10695.

\bibitem{CNN}
Krizhevsky A, Sutskever I, Hinton GE. Imagenet classification with deep
  convolutional neural networks. In \emph{Advances in neural information
  processing systems}, 2012, 1097--1105.

\bibitem{styimprove1}
Dumoulin V, Shlens J, Kudlur M. A learned representation for artistic style.
  \emph{arXiv preprint arXiv:1610.07629}, 2016.

\bibitem{styimprove2}
Li Y, Wang N, Liu J, Hou X. Demystifying neural style transfer. \emph{arXiv
  preprint arXiv:1701.01036}, 2017.

\bibitem{styimprove3}
Yin R. Content aware neural style transfer. \emph{arXiv preprint
  arXiv:1601.04568}, 2016.

\bibitem{styimprove5}
Gatys LA, Bethge M, Hertzmann A, Shechtman E. Preserving color in neural
  artistic style transfer. \emph{arXiv preprint arXiv:1606.05897}, 2016.

\bibitem{STROTSS}
Kolkin N, Salavon J, Shakhnarovich G. Style transfer by relaxed optimal
  transport and self-similarity. In \emph{Proceedings of the IEEE/CVF
  Conference on Computer Vision and Pattern Recognition}, 2019, 10051--10060.

\bibitem{EMD}
Kusner MJ, Sun Y, Kolkin NI, Weinberger KQ. From Word Embeddings to Document
  Distances. In \emph{Proceedings of the 32nd International Conference on
  International Conference on Machine Learning - Volume 37}, 2015, 957–966.

\bibitem{AdaIN}
Huang X, Belongie S. Arbitrary style transfer in real-time with adaptive
  instance normalization. In \emph{Proceedings of the IEEE International
  Conference on Computer Vision}, 2017, 1501--1510.

\bibitem{highresolutionpix2pix}
Wang TC, Liu MY, Zhu JY, Tao A, Kautz J, Catanzaro B. High-resolution image
  synthesis and semantic manipulation with conditional gans. In
  \emph{Proceedings of the IEEE conference on computer vision and pattern
  recognition}, 2018, 8798--8807.

\bibitem{yi2017dualgan}
Yi Z, Zhang H, Tan P, Gong M. Dualgan: Unsupervised dual learning for
  image-to-image translation. In \emph{Proceedings of the IEEE international
  conference on computer vision}, 2017, 2849--2857.

\bibitem{hoffman2018cycada}
Hoffman J, Tzeng E, Park T, Zhu JY, Isola P, Saenko K, Efros A, Darrell T.
  Cycada: Cycle-consistent adversarial domain adaptation. In
  \emph{International conference on machine learning}, 2018, 1989--1998.

\bibitem{zhu2017toward}
Zhu JY, Zhang R, Pathak D, Darrell T, Efros AA, Wang O, Shechtman E. Toward
  multimodal image-to-image translation. In \emph{Advances in neural
  information processing systems}, 2017, 465--476.

\bibitem{huang2018multimodal}
Huang X, Liu MY, Belongie S, Kautz J. Multimodal unsupervised image-to-image
  translation. In \emph{Proceedings of the European Conference on Computer
  Vision (ECCV)}, 2018, 172--189.

\bibitem{lee2018diverse}
Lee HY, Tseng HY, Huang JB, Singh M, Yang MH. Diverse image-to-image
  translation via disentangled representations. In \emph{Proceedings of the
  European conference on computer vision (ECCV)}, 2018, 35--51.

\bibitem{lee2020drit++}
Lee HY, Tseng HY, Mao Q, Huang JB, Lu YD, Singh M, Yang MH. Drit++: Diverse
  image-to-image translation via disentangled representations.
  \emph{International Journal of Computer Vision}, 2020: 1--16.

\bibitem{liu2019few}
Liu MY, Huang X, Mallya A, Karras T, Aila T, Lehtinen J, Kautz J. Few-shot
  unsupervised image-to-image translation. In \emph{Proceedings of the IEEE
  International Conference on Computer Vision}, 2019, 10551--10560.

\bibitem{chen2018gated}
Chen X, Xu C, Yang X, Song L, Tao D. Gated-gan: Adversarial gated networks for
  multi-collection style transfer. \emph{IEEE Transactions on Image
  Processing}, 2018, 28(2): 546--560.

\bibitem{chang2020domain}
Chang HY, Wang Z, Chuang YY. Domain-specific mappings for generative
  adversarial style transfer. In \emph{Computer Vision--ECCV 2020: 16th
  European Conference, Glasgow, UK, August 23--28, 2020, Proceedings, Part VIII
  16}, 2020, 573--589.

\bibitem{SwappingAutoencoder}
Park T, Zhu JY, Wang O, Lu J, Shechtman E, Efros A, Zhang R. Swapping
  Autoencoder for Deep Image Manipulation. In H~Larochelle, M~Ranzato,
  R~Hadsell, M~Balcan, H~Lin, editors, \emph{Advances in Neural Information
  Processing Systems}, volume~33, 2020, 7198--7211.

\bibitem{ChipGAN}
He B, Gao F, Ma D, Shi B, Duan LY. ChipGAN: A generative adversarial network
  for Chinese ink wash painting style transfer. In \emph{Proceedings of the
  26th ACM international conference on Multimedia}, 2018, 1172--1180.

\bibitem{deeplabv2}
Chen LC, Papandreou G, Kokkinos I, Murphy K, Yuille AL. Deeplab: Semantic image
  segmentation with deep convolutional nets, atrous convolution, and fully
  connected crfs. \emph{IEEE transactions on pattern analysis and machine
  intelligence}, 2017, 40(4): 834--848.

\bibitem{v2v}
Wang TC, Liu MY, Zhu JY, Liu G, Tao A, Kautz J, Catanzaro B. Video-to-video
  synthesis. \emph{arXiv preprint arXiv:1808.06601}, 2018.

\bibitem{choi2018stargan}
Choi Y, Choi M, Kim M, Ha JW, Kim S, Choo J. Stargan: Unified generative
  adversarial networks for multi-domain image-to-image translation. In
  \emph{Proceedings of the IEEE conference on computer vision and pattern
  recognition}, 2018, 8789--8797.

\bibitem{qi2018semiparametric}
Qi X, Chen Q, Jia J, Koltun V. Semi-parametric image synthesis. In
  \emph{Proceedings of the IEEE Conference on Computer Vision and Pattern
  Recognition}, 2018, 8808--8816.

\bibitem{wang2019example}
Wang M, Yang GY, Li R, Liang RZ, Zhang SH, Hall PM, Hu SM. Example-guided
  style-consistent image synthesis from semantic labeling. In \emph{Proceedings
  of the IEEE/CVF conference on computer vision and pattern recognition}, 2019,
  1495--1504.

\bibitem{zhang2019shadowgan}
Zhang S, Liang R, Wang M. Shadowgan: Shadow synthesis for virtual objects with
  conditional adversarial networks. \emph{Computational Visual Media}, 2019, 5:
  105--115.

\bibitem{zhou2021jittor}
Zhou WY, Yang GW, Hu SM. Jittor-GAN: A fast-training generative adversarial
  network model zoo based on Jittor. \emph{Computational Visual Media}, 2021,
  7: 153--157.

\bibitem{wang2023towards}
Wang C, Tang F, Zhang Y, Wu T, Dong W. Towards harmonized regional style
  transfer and manipulation for facial images. \emph{Computational Visual
  Media}, 2023, 9(2): 351--366.

\bibitem{styleGANv2}
Karras T, Laine S, Aittala M, Hellsten J, Lehtinen J, Aila T. Analyzing and
  improving the image quality of stylegan. In \emph{Proceedings of the IEEE/CVF
  Conference on Computer Vision and Pattern Recognition}, 2020, 8110--8119.

\bibitem{SegmentationArt}
Cohen N, Newman Y, Shamir A. {Semantic Segmentation in Art Paintings}.
  \emph{Computer Graphics Forum}, 2022, \doi{10.1111/cgf.14473}.

\bibitem{coco}
Lin TY, Maire M, Belongie S, Hays J, Perona P, Ramanan D, Doll{\'a}r P, Zitnick
  CL. Microsoft coco: Common objects in context. In \emph{European conference
  on computer vision}, 2014, 740--755.

\bibitem{zhou2019semantic}
Zhou B, Zhao H, Puig X, Xiao T, Fidler S, Barriuso A, Torralba A. Semantic
  understanding of scenes through the ade20k dataset. \emph{International
  Journal of Computer Vision}, 2019, 127(3): 302--321.

\bibitem{LabelSmoothingviaGraphCuts}
{Boykov} Y, {Veksler} O, {Zabih} R. Fast approximate energy minimization via
  graph cuts. \emph{IEEE Transactions on Pattern Analysis and Machine
  Intelligence}, 2001, 23(11): 1222--1239.

\bibitem{VGG}
Simonyan K, Zisserman A. Very deep convolutional networks for large-scale image
  recognition. \emph{arXiv preprint arXiv:1409.1556}, 2014.

\bibitem{HED}
Xie S, Tu Z. Holistically-nested edge detection. In \emph{Proceedings of the
  IEEE international conference on computer vision}, 2015, 1395--1403.

\bibitem{gu2020giqa}
Gu S, Bao J, Chen D, Wen F. Giqa: Generated image quality assessment. In
  \emph{Computer Vision--ECCV 2020: 16th European Conference, Glasgow, UK,
  August 23--28, 2020, Proceedings, Part XI 16}, 2020, 369--385.

\bibitem{fu2019dual}
Fu J, Liu J, Tian H, Li Y, Bao Y, Fang Z, Lu H. Dual attention network for
  scene segmentation. In \emph{Proceedings of the IEEE/CVF Conference on
  Computer Vision and Pattern Recognition}, 2019, 3146--3154.

\bibitem{interfaceGAN}
Shen Y, Gu J, Tang X, Zhou B. Interpreting the latent space of gans for
  semantic face editing. In \emph{Proceedings of the IEEE/CVF Conference on
  Computer Vision and Pattern Recognition}, 2020, 9243--9252.

\bibitem{UMAP}
McInnes L, Healy J, Melville J. Umap: Uniform manifold approximation and
  projection for dimension reduction. \emph{arXiv preprint arXiv:1802.03426},
  2018.

\bibitem{johnson2016perceptual}
Johnson J, Alahi A, Fei-Fei L. Perceptual losses for real-time style transfer
  and super-resolution. In \emph{Computer Vision--ECCV 2016: 14th European
  Conference, Amsterdam, The Netherlands, October 11-14, 2016, Proceedings,
  Part II 14}, 2016, 694--711.

\bibitem{kldloss}
Kingma DP, Welling M. Auto-encoding variational bayes. \emph{arXiv preprint
  arXiv:1312.6114}, 2013.

\bibitem{heusel2017gans}
Heusel M, Ramsauer H, Unterthiner T, Nessler B, Hochreiter S. Gans trained by a
  two time-scale update rule converge to a local nash equilibrium. In
  \emph{Advances in neural information processing systems}, 2017, 6626--6637.

\end{thebibliography}

\subsection*{Author biography}
\begin{biography}[figures/huang_yuantian]{Yuantian Huang} is a Ph.D. student at
Degree Programs in Systems and Information Engineering,
University of Tsukuba. 
His research interests include image editing and generation, computer graphics, and computer vision.
\end{biography}

\begin{biography}[figures/iizuka_satoshi]{Satoshi Iizuka} is an associate professor at 
University of Tsukuba's Faculty of Engineering, Information and Systems. 
He received his Ph.D. in engineering from the University of Tsukuba. 
His research interests are in computer graphics and vision, 
including image processing and editing based on machine learning.
\end{biography}
 
\begin{biography}[figures/edgar_simoserra]{Edgar Simo-Serra} is an associate professor
at Waseda University. He received his Ph.D. from BarcelonaTech (UPC). His
general research interests are in the intersection of computer vision, computer
graphics, and machine learning with applications to large-scale real-world problems.
\end{biography}
\begin{biography}[figures/kazuhiro_fukui]{Kazuhiro Fukui}
joined the Toshiba Corporate R\&D Center and served as a
Senior Research Scientist at Multimedia Laboratory. He received his
Ph.D. from Tokyo Institute of Technology in 2003. He is
currently a professor in the Department of Computer Science, Graduate
School of Systems and Information Engineering at the University of
Tsukuba. His research interests include the theory of machine
learning, computer vision, pattern recognition, and their applications.
He is a member of IEEE and SIAM.
\end{biography}

\end{document}